%% file: main.tex
\documentclass{article}
\PassOptionsToPackage{numbers,sort&compress}{natbib}
\usepackage[preprint]{neurips_2026}

\usepackage[utf8]{inputenc} % allow utf-8 input
\usepackage[T1]{fontenc}    % use 8-bit T1 fonts
\usepackage{hyperref}       % hyperlinks
\usepackage{url}            % simple URL typesetting
\usepackage{booktabs}       % professional-quality tables
\usepackage{amsfonts}       % blackboard math symbols
\usepackage{nicefrac}       % compact symbols for 1/2, etc.
\usepackage{microtype}      % microtypography
\usepackage{xcolor}         % colors
\usepackage{algorithm}
\usepackage{algpseudocode}
\usepackage{subcaption}
\usepackage{multirow}
\usepackage{graphicx}
\usepackage{amsmath}

\title{Adaptive Second-Order Solvers for Fast Stochastic Diffusion Sampling}

\workshoptitle{BeNTo: Beyond Next‑Token Prediction}

\author{%
  Ella Kemperman \\
  Radboud University\\
  \texttt{ella.kemperman@ru.nl} \\
  \And
  Luca Ambrogioni\\
  Radboud University\\
  \texttt{luca.ambrogioni@donders.ru.nl} \\
}

\begin{document}

\maketitle

\begin{abstract}
Diffusion models rely on numerical solvers requiring time-discretization, which has a large influence on the tradeoff between sampling cost and quality. However, the computational difficulty of the reverse process varies along the sampling trajectory and across data distributions, making the choice of discretization important. We adapt proportional-integral (PI) step-size control to diffusion, using our diffusion noise-normalised error estimator. Unlike existing adaptive methods in diffusion that respond only to the current error, the PI solver also incorporates the previous error, yielding smoother step adaptation. We further show that these per-sample trajectories exhibit shared structure and can be aggregated into a fixed schedule that retains much of the benefit of adaptive sampling. We evaluate both approaches on natural-image and language datasets, in terms of quality, measured by FID at a matched number of neural network evaluations (NFE), comparing them with widely used stochastic solvers and schedules. For images, our average discretization outperforms the commonly used EDM schedule in terms of sample quality when used with the stochastic Heun sampler, and with the EDM-churn sampler at low NFE. Additionally, our PI adaptive solver obtains better FID than most stochastic and adaptive baselines, although it does not beat the EDM-churn sampler at low NFE. Moreover, we find our solver outperforms both the EDM and the entropy schedule on language diffusion at low-to-medium NFE in terms of perplexity, with the drawback of lower token entropy. Lastly, we find that the benefit of per-sample adaptivity is problem-dependent. It is highly beneficial in 1D toy examples, while only marginal for image and language data, where the average schedule sometimes even outperforms the per-sample adaptive solver. Code is available at \href{https://github.com/ellakemperman/adaptive-second-order-diffusion-solvers}{https://github.com/ellakemperman/adaptive-second-order-diffusion-solvers}. 
\end{abstract}

\input{sections/body/1-introduction}

\input{sections/body/3-related_works}

\input{sections/body/2-background}

\input{sections/body/4-proportional_integral_solver}
\input{sections/body/5-results}
\input{sections/body/6-discussion}
\input{sections/body/7-AI_statement}
\input{sections/body/9-reproducibility_statement}

\bibliographystyle{unsrtnat}
\bibliography{bibliography}

\appendix
\input{sections/appendices/edm_forward_sde}

\input{sections/appendices/solver_details}
\input{sections/appendices/hyperparemeter_selection}
\input{sections/appendices/implementation_details}
\input{sections/appendices/samples}

\end{document}

%% file: sections/body/1-introduction.tex
\section{Introduction}
\label{chap:intro}
\begin{figure}
    \begin{subfigure}[b]{0.45\textwidth}
        \centering
        \includegraphics[width=0.99\linewidth]{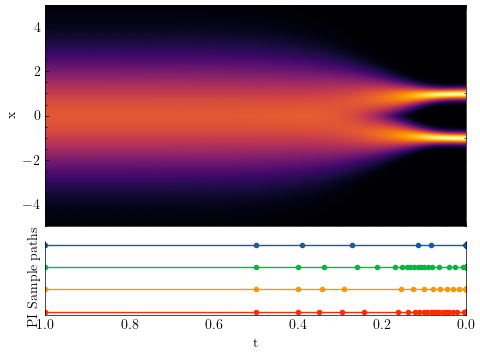}
        \caption{}
        \label{fig:bifurcation}
    \end{subfigure}
    \hfill
    \begin{subfigure}[b]{0.45\textwidth}
        \centering
        \includegraphics[width=0.97\linewidth]{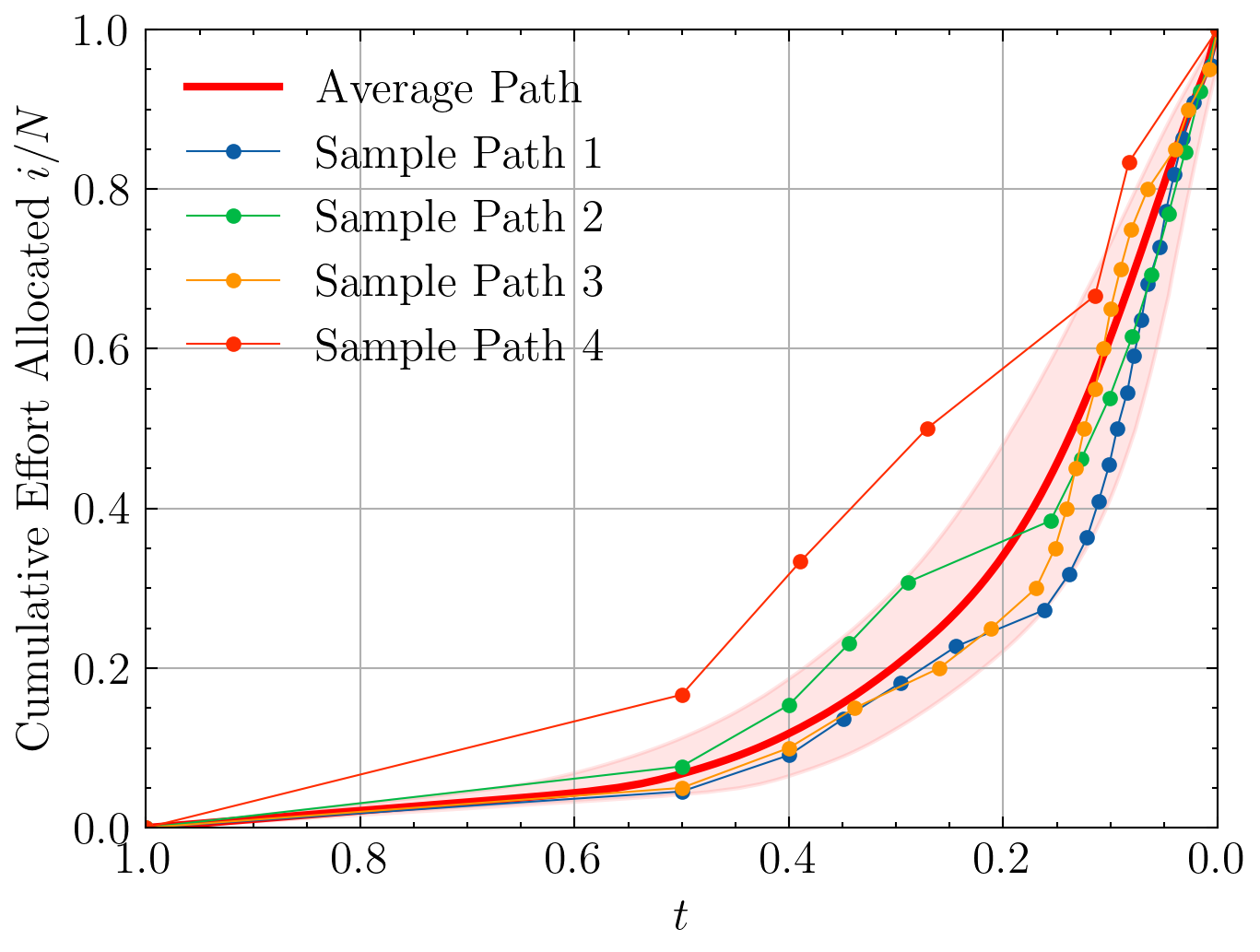}
        \caption{}
        \label{fig:intro_traj}
    \end{subfigure}
    \caption{The difficulty of generative diffusion differs across time, requiring different step-sizes at different time-points, which our PI solver determines (a). These per-sample paths share structure and can then be aggregated into an average path usable for sampling (b).}
    \label{fig:intro_fig}
\end{figure}

Diffusion models generate high-quality images  \citep{song_score-based_2021, karras_elucidating_2022, song_denoising_2022, rombach_high-resolution_2022}, videos \citep{singer_make--video_2022}, and audio \citep{kong_diffwave_2021}. However, diffusion sampling is slow, as it is a process requiring a large number of function evaluations (NFE) of a neural network \citep{song_score-based_2021, karras_elucidating_2022, jolicoeur-martineau_gotta_2021}. \citet{song_score-based_2021} show that the process of diffusion can be treated as solving either an ordinary or stochastic differential equation (ODE/SDE). Recent work \citep{li_critical_2024} identified critical windows within the integration interval during which many of the generative decisions are made, a phenomenon arising from symmetry-breaking bifurcations in the reverse denoising process \citep{bytezcom_spontaneous_2023, ambrogioni_statistical_2025}. This suggests that numerical effort should not be distributed uniformly across the diffusion interval, but rather, more effort should be allocated to the parts of the interval where dynamics are difficult (\autoref{fig:bifurcation}), which could improve the NFE-quality tradeoff.

Many widely used schedules are handcrafted and calibrated to specific modalities \citep{song_score-based_2021, karras_elucidating_2022, song_denoising_2022, ho_denoising_2020}. Recent work by \citet{stancevic_entropic_2025} discussed constructing these schedules in a data-driven manner by enforcing a linear progression in entropy. However, the authors find that their schedules are effective only for first-order samplers, with performance degrading when used with second-order methods. This limits the applicability of such schedules, as second-order samplers have been shown to improve the NFE--quality trade-off in diffusion models \citep{karras_elucidating_2022}.

Thus, we require a method of generating schedules in a data-driven manner that also functions for second-order samplers. Numerical analysis literature offers an alternative that satisfies both conditions: adaptive solvers \citep{lamba_mean-square_2006, lehn_adaptive_2002, kelly_adaptive_2022, ilie_adaptive_2015, mauthner_step_1998, noauthor_solving_1993}, which adapt step-size based on the local difficulty, allowing smaller step-sizes along more difficult regions. Additionally, these are generally agnostic to solver-order, allowing them to generate second-order schedules. This allows the schedule to differ on a per-sample basis, and removes the need for handcrafting. Application of these adaptive solvers to diffusion remains limited, though not unexplored \citep{karras_elucidating_2022, jolicoeur-martineau_gotta_2021}. \citet{jolicoeur-martineau_gotta_2021} devise an adaptive SDE solver specifically designed for diffusion, adapting step-size based on the current local-error. 

We instead adapt stochastic proportional-integral (PI) control for step-size adaptation \citep{ilie_adaptive_2015}, which incorporates the previous local error alongside the current one, giving smoother step-size adaptation than integral-only control \citep{gustafsson_api_1988}. We adapt its error estimator to normalise by diffusion noise, substantially improving its performance in diffusion sampling. We compare our method to existing diffusion sampler-schedule combinations in terms of NFE-quality tradeoff. Additionally, we show that the per-sample trajectories have shared structure, which can be exploited by aggregating the trajectories into a single static schedule, usable as any other handcrafted schedule (an example of this is shown in \autoref{fig:intro_traj}). We evaluate its performance and compare it with the per-sample adaptive solver to determine if per-sample adaptivity is beneficial.

%% file: sections/body/3-related_works.tex
\section{Related Works}
\label{chap:related_works}
\paragraph{Diffusion Samplers}  Many improvements to diffusion sampling without retraining have been made since diffusion models were introduced \citep{sohl-dickstein_deep_2015, ho_denoising_2020}. \citet{song_score-based_2021} introduce the Predictor-Corrector sampler, alternating predictor and corrector steps. \citet{zhao_unipc_2023} devise a unified predictor-corrector framework called UniPC. A different line of work looks into higher-order solvers. \citet{karras_elucidating_2022} propose a second-order solver using Heun's method, finding that it sits optimally at the NFE-quality tradeoff, and introduce a
stochastic sampler that temporarily increases noise at each step before denoising (churn). \citet{lu_dpm-solver_2022, lu_dpm-solver_2025} develop an efficient higher-order solver, using the insight that the drift part of the DEs can be analytically solved. To summarize, extensive work has gone into improving diffusion samplers and improving the sampling technique can substantially improve sample quality at low NFE. Unlike these works, this paper focuses on time-step selection using adaptive step-size control for second-order samplers, rather than the specific numerical integration method used.

\paragraph{Diffusion Discretization}  Determining the optimal discretizations for the probability flow ODE and reverse SDE has been an active topic of research after \citet{song_score-based_2021} introduced the SDE framework and two handcrafted discretizations. \citet{karras_elucidating_2022} devise a new way of parametrizing the probability flow ODE and handcraft a new discretization (EDM). Besides handcrafting, there are methods that obtain discretizations in a principled way. \citet{sabour_align_2024} introduce Align Your Steps, a technique that is grounded in stochastic calculus, obtaining optimized discretizations based on the solver used and the dataset. \citet{chen_adaptive_nodate} similarly optimize the time discretization, using gradient descent. \citet{stancevic_entropic_2025} discretize time in a way that is linearly spaced in terms of entropy, which can be reliably estimated from the score matching loss. However, the performance of their schedule drops for second-order samplers. Moreover, there are techniques that aim to optimize time-discretizations in an empirical way. \citet{xia_towards_2025} train a simple model that predicts the next time step based on a sample $\mathbf{x}_t$ and a time $t$. \citet{huang_art_2026} take a reinforcement learning approach, treating time-step scheduling as a control problem. They use this approach to determine a data-driven time discretization, which they show can be interpolated and extrapolated across different NFE. However, they only show their technique to work for first-order solvers. In contrast to these approaches, this work does not directly optimize the discretization based on a predefined objective. Instead, discretizations are determined per sample and per step, using a second-order adaptive solver with PI control, and these are aggregated into one static discretization.

\paragraph{Adaptive SDE Samplers}  A substantial amount of work in the field of numerical analysis has gone into improving SDE sampling with adaptive solvers, as many different fields benefit from this. \citet{mauthner_step_1998} develop a technique that uses an embedded Stochastic Runge-Kutta method to determine the local error, using that error to control the step-size. \citet{ilie_adaptive_2015} expand on this approach, introducing an adaptive solver using PI control specifically designed for multidimensional SDEs, taking inspiration from \citet{gustafsson_api_1988} and adding a component incorporating the previous error. This approach forms the basis of the adaptive solver we apply to generative diffusion. The closest work in diffusion literature is by \citet{jolicoeur-martineau_gotta_2021}. They devise an adaptive stochastic solver and apply it to diffusion, using the difference between the first- and second-order estimates as an error metric. Their work differs from this work in the step-size controller used, only using a component proportional to the current error, not considering previous error.

%% file: sections/body/2-background.tex
\section{Background}
\label{chap:background}

%\subsection{Score-Based Generative Diffusion}
%\label{sec:diffusion}
A diffusion process progressively adds noise to a sample, transforming an initial distribution into another distribution over time. More formally, let a diffusion process run on the time interval $[t_0, T]$. Let $\mathbf{x} \in \mathbb{R}^d$ be a sample from the data distribution $p_t(\mathbf{x})$ at time $t$. At $t=t_0$, $p_0(\mathbf{x})$ represents the target distribution, e.g. the distribution of all images in a dataset. At $t=T$, $p_T(\mathbf{x})$ represents a simple distribution from which can be sampled, which is generally chosen to be a Gaussian. Diffusion modeling uses two separate but related diffusion processes. The forward diffusion process transforms data from $p_0(\mathbf{x})$ to $p_T(\mathbf{x})$ and is required for training. The reverse diffusion process does the opposite and is used for inference. These two diffusion processes can be modeled by SDEs \citep{song_score-based_2021}. 

The forward diffusion process progressively adds noise to a sample $\mathbf{x}$ at $t=t_0$, until the sample is fully corrupted at $t=T$ \citep{song_score-based_2021}. This process is governed by the following SDE in Itô form:
$
    \mathrm{d}\mathbf{x} =  \mathbf{f}(\mathbf{x}, t)\mathrm{d}t + g(t)\mathrm{d}\mathbf{w}
$.
With $\mathbf{f}(\mathbf{x}, t):\mathbb{R}^d\times \mathbb{R}\rightarrow \mathbb{R}^d$ the drift function, $g(t): \mathbb{R}\rightarrow \mathbb{R}$ the diffusion function, and $\mathbf{w}(t)$ the standard Wiener process. To perform inference, the reverse diffusion process is used. This process progressively removes noise from a sample, eventually obtaining a sample from the target distribution. The reverse diffusion process is governed by the reverse time SDE \citep{song_score-based_2021, anderson_reverse-time_1982}:
\begin{equation}
    \label{eq:rdp}
    \mathrm{d}\mathbf{x} =[\mathbf{f}(\mathbf{x}, t) - g(t)^2\nabla_\mathbf{x} \log p_t(\mathbf{x})]\mathrm{d}t + g(t)\mathrm{d}\mathbf{\tilde{w}}
\end{equation}
Here, $\mathbf{\tilde{w}}(t)$ is the reverse Wiener process and $\nabla_\mathbf{x} \log p_t(\mathbf{x})$ is known as the score. The score can be interpreted as a vector field that points towards the modes of the distribution. The score is generally given by a trained neural network $s_{\theta}(\mathbf{x}, t)$ trained via denoising score matching \citep{song_score-based_2021}.

While samples from the data distribution can be obtained with the reverse SDE (\ref{eq:rdp}), this process can be slow \citep{karras_elucidating_2022, song_denoising_2022}. However, the process can also be modeled by an ODE, known as the probability flow ODE \citep{song_score-based_2021}:
\begin{equation}
    \label{eq:ode}
        \mathrm{d}\mathbf{x} =[\mathbf{f}(\mathbf{x}, t) - \frac{1}{2}g(t)^2\nabla_\mathbf{x} \log p_t(\mathbf{x})]\mathrm{d}t 
\end{equation}
The trajectories of the ODE have the same marginal probability densities as the reverse SDE.

\paragraph{Numerical Integration} As the reverse SDE (\ref{eq:rdp}) generally does not have analytical solutions, they have to be numerically approximated by discretizing continuous time into time points $t_i|_{i=0}^N$, and computing a time step $h_i = t_{i+1} - t_{i}$. This time step is used to numerically integrate the differential equation, computing $\mathbf{\hat{x}}_{i+1} = \mathbf{\hat{x}}_{i} + \Delta\mathbf{x}_i$, where $\Delta \mathbf{x}_i(\mathbf{x}_{i}, h_i)$ is based on the to-be-solved SDE and the numerical integration scheme. Several integration schemes exist, such as Euler-Maruyama \citep{kloeden_numerical_1992} and stochastic Heun's method \citep{rumelin_numerical_1982}. Additionally, different ways of determining the discretization exist. Traditionally, this is picked by hand.

%Instead, adaptive methods determine the discretization dynamically while the SDE is solved. They are based on the principle that not all regions of a differential equation have the same difficulty \citep{noauthor_solving_1993}, assuming more difficult regions take more steps to accurately approximate than less difficult regions. A region can be seen as more difficult if the same step size introduces more discretization error in this region than in another, less difficult region. To vary the step amount in specific regions, adaptive solvers adapt the step-size $h_i$ at each point $i$ based on some computed quantity, e.g. the local error of the step taken. This computed quantity is then given to a step-size controller, which uses this quantity to adapt the step size. This can create different step trajectories for each sample. Generally, adaptive solvers still use an integration scheme like non-adaptive solvers. The exact method used depends on the adaptive solver.

% While handcrafted discretisations can create similar trajectories, the individual adaptivity is lost with a handcrafted schedule. Whether this per-sample adaptivity is beneficial, or if an average trajectory constructed from the individual trajectories suffices, is investigated in this work.

%% file: sections/body/4-proportional_integral_solver.tex
\section{PI Controlled Adaptive Solver}
\label{chap:adaptive-solver}
Our PI diffusion solver uses the PI control method introduced in \citet{ilie_adaptive_2015}. The solver consists of an error estimator, a step controller, optional added steps on the probability flow ODE, and we discuss how the solver handles batches. The algorithm is written for an $\mathbf{x}$ that is not part of a batch, but was extended to work with batches for the experiments (details in \autoref{app:solver_details}). For brevity, we omit the ODE steps taken from this algorithm. The full algorithm can be found in Algorithm \ref{alg:pi-solver}.
%Note that the solver itself uses stochastic Heun's method \citep{rumelin_numerical_1982} for integration.

\paragraph{Error Estimator} We adapt step-size based on the local error, which is given by the difference between the exact and estimated solution. However, for the reverse SDE (\ref{eq:rdp}), no exact solution is known in most diffusion contexts. Therefore, we estimate the error instead. \citet{ilie_adaptive_2015} use the extrapolated error to estimate the local error, given by the difference between the second-order estimate $x''$ and the first-order estimate $x'$. Our solver uses stochastic Heun's method \citep{rumelin_numerical_1982} to determine $x''$. We adapt the error estimator by \citet{ilie_adaptive_2015}, adding a tolerance term relative to the diffusion noise, yielding better performance for variance-exploding diffusion. For more information on these changes, see \autoref{app:solver_details}.
\begin{equation}
    \label{eq:error_1}
    \epsilon(\mathbf{x}_i, h_i) = \sqrt{\frac{1}{d}\sum_{k=1}^d \left(\frac{\mathbf{x}_{i+1, k}' - \mathbf{x}_{i+1, k}''}{\tau_{abs} + \tau_{rel} \sigma(t)}\right)^2}
\end{equation}
Where $\epsilon(\mathbf{x}_i, h_i)\in\mathbb{R}_{>0}$ is the local extrapolated error, $\tau_{abs} \in \mathbb{R}_{>0}$ a user-defined tolerance parameter which scales the error by a constant, $\tau_{rel} \in \mathbb{R}_{>0}$ the relative tolerance, which scales the error based on the data. Specifically, $\tau_{rel}$ scales by the current noise level $\sigma(t)$. 

Note that \citet{ilie_adaptive_2015} also use a rejection mechanism. However, we found that rejected steps generally led to a worse NFE-quality tradeoff, as compute was wasted on a rejected step. Therefore, for the final experiments, step-rejection was disabled.

\paragraph{Step Controller} The PI solver adapts the step size based on the error. For this, the PI controller by \citet{ilie_adaptive_2015} is used. The PI controller not only uses the current error, but also uses the error of the previous step. They define the PI controller as:
\begin{equation}
    \label{eq:pi-controller}
    h_{i+1} = h_i\min \left(\alpha_{max}, \max\left[\alpha_{min},\left(\frac{\alpha}{\epsilon(\mathbf{x}_i, h_i)}\right)^{k_i + k_p}\left(\frac{\epsilon(\mathbf{x}_{i-1}, h_{i-1})}{\alpha}\right)^{k_p}\right]\right)
\end{equation}
With $ 0< \alpha \leq 1$ a safety factor, generally set to 0.9. $\alpha_{min}$ and $\alpha_{max}$ are also safety factors, which set a minimum and maximum value the step-size can increase in a single iteration. $k_i$ is the integral gain, and $k_p$ is the proportional gain. \citet{ilie_adaptive_2015} suggested two settings for $k_i$ and $k_p$, namely $k_i=0.3, k_p=0.1$ and $k_i=0.101$ and $k_p=0.009$. After some testing, it was found that the setting $k_i=0.3$ and $k_p = 0.1$ performed well (see \autoref{app:hyperparameter-selection} for more details). 

This equation implicitly introduces two more hyperparameters. The first of these is the starting step size $h_0$. This hyperparameter is dependent on all other hyperparameters, and the values used can be found in \autoref{app:hyperparameter-selection}. The second of these being the initial previous error $\epsilon(\mathbf{x}_0, h_0)$. We set $\epsilon(\mathbf{x}_0, h_0) = \alpha$, as that cancels out the influence of the second term in the first iteration.

%\subsubsection{Batch Averaging}
%From early experimentation, it was found that the independent paths of each element in the batch had very low variance. Furthermore, the error that is separately calculated for each element in the batch is quite sensitive to noise. Therefore, an additional mode of operation was added to the solver. 

%This mode of operation normalises the error over each element in the batch. The batch-averaged error is defined as follows:
%\begin{equation}
%    \label{eq:batch-norm}
%    \epsilon_{batch}(\mathbf{x}_{i}, h_i) = \frac{1}{N}\sum_{j=1}^N\epsilon(\mathbf{x}_{i, j}, h_i)
%\end{equation}
%Where $N$ is the batch size and $\mathbf{x}_{i,j}$ is the $j$'th element of the batch at the $i$'th step. Note that as the same error is used for each element in the batch, it is implied that each element in the batch will have the same next step $h_{i+1}$, implying that the images, using batch averaging, will all have the same step $h_i$ at each $i$.

\paragraph{ODE Steps} Early experiments showed that at low NFE, the solver produces grainy images. This was likely caused by noise being introduced by stochasticity that cannot be fully removed at low step counts, which \citet{karras_elucidating_2022} also observe with their stochasticity injection method. They run the last few steps without stochasticity to avoid this problem. Therefore, we sample the last few steps at low noise levels using the probability flow ODE (\ref{eq:ode}) with a Heun sampler and a predetermined schedule.

\begin{algorithm}
\caption{The PI solver altered to function well on the reverse diffusion process.}
\label{alg:pi-solver}
\begin{algorithmic}
\Procedure{PISolver}{$\tau_{abs}$, $\tau_{rel}$, $T$, $t_{ode}$, $t_0$, $h_{0}$, $\sigma(t)$, $\mathbf{f}(\mathbf{x}, t)$, $g(t)$, $s_{\theta}(\mathbf{x}, t)$,$k_i = 0.3$, $k_p=0.1$, $\alpha=0.9$, $\alpha_{min}=0.2$, $\alpha_{max}=5$}
    \State \textbf{sample} $x \sim \mathcal{N}(0, \sigma(T)^2\mathbf{I})$
    \State $h \leftarrow -|h_{0}|$ \Comment{Ensure steps are taken in the reverse direction}
    \State $t \leftarrow T$, $\epsilon_0 \leftarrow \alpha$, $i \leftarrow 1$
    \While{$t > t_{ode}$}
        \State \textbf{sample} $\Delta w \sim \mathcal{N}(0, |h|\mathbf{I})$
        \State $\Delta \mathbf{x}' \leftarrow [\mathbf{f}(\mathbf{x}, t) - g(t)^2s_{\theta}(\mathbf{x}, t)]h + g(t)\Delta w$ \Comment{Euler-Maruyama step}
        \State $\mathbf{x}' \leftarrow \mathbf{x} + \Delta \mathbf{x}$
        \State $\Delta \mathbf{x}'' \leftarrow [\mathbf{f}(\mathbf{x}', t + h) - g(t + h)^2s_{\theta}(\mathbf{x}', t + h)]h + g(t + h)\Delta w$ \Comment{Second-order step}
        \State $\mathbf{x}'' \leftarrow \mathbf{x} + \frac{1}{2}(\Delta \mathbf{x}' + \Delta\mathbf{x}'')$ \Comment{Heun}
        \State $\epsilon_i \leftarrow \sqrt{\frac{1}{d}\sum_{k\leftarrow1}^d \left(\frac{\mathbf{x}_{i+1, k}' - \mathbf{x}_{i+1, k}''}{\tau_{abs} + \tau_{rel}\sigma(t)}\right)^2}$ \Comment{Extrapolated error (\ref{eq:error_1})}
        \\
        \State $\mathbf{x} \leftarrow \mathbf{x}''$ \Comment{Use second order estimate to update $\mathbf{x}$}
        \State $t \leftarrow t + h$
        \State $h \leftarrow h\min \left(\alpha_{max}, \max\left[\alpha_{min},\left(\frac{\alpha}{\epsilon_i}\right)^{k_i + k_p}\left(\frac{\epsilon_{i-1}}{\alpha}\right)^{k_p}\right]\right)$ \Comment{Update step (\ref{eq:pi-controller})}
        \State $h \leftarrow \max(h, t_{ode} - t)$ \Comment{Ensure $t$ is never less than $t_{ode}$}
        \State $i \leftarrow i + 1$
    \EndWhile
    % \ForAll{$h \in H_{ode}$} \Comment{Heun ODE steps}
    %     \State $\Delta \mathbf{x}' \leftarrow \mathbf{f}_{ode}(\mathbf{x}, t)h$
    %     \State $\mathbf{x}' \leftarrow \mathbf{x} + \Delta \mathbf{x}$
    %     \State $\Delta \mathbf{x}'' \leftarrow \mathbf{f})_{ode}(\mathbf{x}', t + h)h$
    %     \State $\mathbf{x} \leftarrow \mathbf{x}''$
    %     \State $t \leftarrow t + h$
    % \EndFor
    \State \Return $\mathbf{x}$
\EndProcedure
\end{algorithmic}
\end{algorithm}

\section{Obtaining a Static Schedule}
\label{section:static_schedule}
In this section, we detail how we can combine individual length-varying trajectories into a single static trajectory. As individual discretizations differ in length, aggregating over the individual discretizations cannot be done directly. Linear interpolation was applied to solve this problem, as it can be used to take samples from each discretization at exactly the same fraction of the total steps taken. Formally, linear interpolation was applied to the time points $t_{i,j}|_{i=0}^{N_j}$ for each discretization $j$, using $\frac{i}{N_j}$ to have a common scale from 0 to 1 for each discretization, which indicates the total fraction of steps of the discretization have been taken at that point. This creates new, interpolated paths $\hat{t}_{i,j}|_{i=0}^{\hat{N}}$, where $\hat{N}$ is the number of steps in each interpolated path. These new paths can be averaged for each different total fraction along the discretization over each discretization, obtaining the average discretization:
\begin{equation}
    \label{eq:average_discretization}
    \bar{t}_{i} = \frac{1}{M}\sum_{j=1}^M \hat{t}_{i,j} \quad \text{for } 0\leq i < \hat{N}
\end{equation}
With $M$ the total number of paths and $\bar{t}_i$ the average time at the $i$'th point in the discretization. The obtained discretization represents the average location of each evaluation point of the PI solver, conceptually representing the amount of effort the PI solver allocates to each time point in the integration interval. The larger the amount of time between each point on the discretization, the less effort the PI solver allocates to that interval.

%Note that each PI discretization has a part after $t_{ode}$. When obtaining the static schedule, only the part of the discretization on the interval $[T, t_{ode}]$ is considered. To obtain the full discretization, $H_{ode}$ is concatenated to the average schedule.

%% file: sections/body/5-results.tex
\section{Experiments}
\label{chap:results}
\subsection{1D Multimodal Gaussian Diffusion}
\label{sec:1d_experiments}
We evaluated our solver on a one-dimensional multi-modal Gaussian, comparing it to stochastic Heun's method \citep{rumelin_numerical_1982} and Euler-Maruyama \citep{kloeden_numerical_1992}, both using equidistant steps. Additionally, we ran stochastic Heun's method with a static schedule generated by the PI solver using the method described in \autoref{section:static_schedule}. The experiment was performed on the variance-preserving SDE \citep{song_score-based_2021}, using an analytical expression of the score function. We measure the Wasserstein distance at varying NFE. For the PI solver, $\tau_{rel}=0$ was used, and $\tau_{abs}$ was used to control NFE. For details on the experimental implementation, see \autoref{app:implementation_details}, and for the hyperparameters, see \autoref{app:hyperparameter-selection}. 

\begin{figure}
    \begin{subfigure}[b]{0.45\textwidth}
        \centering
        \includegraphics[width=0.99\linewidth]{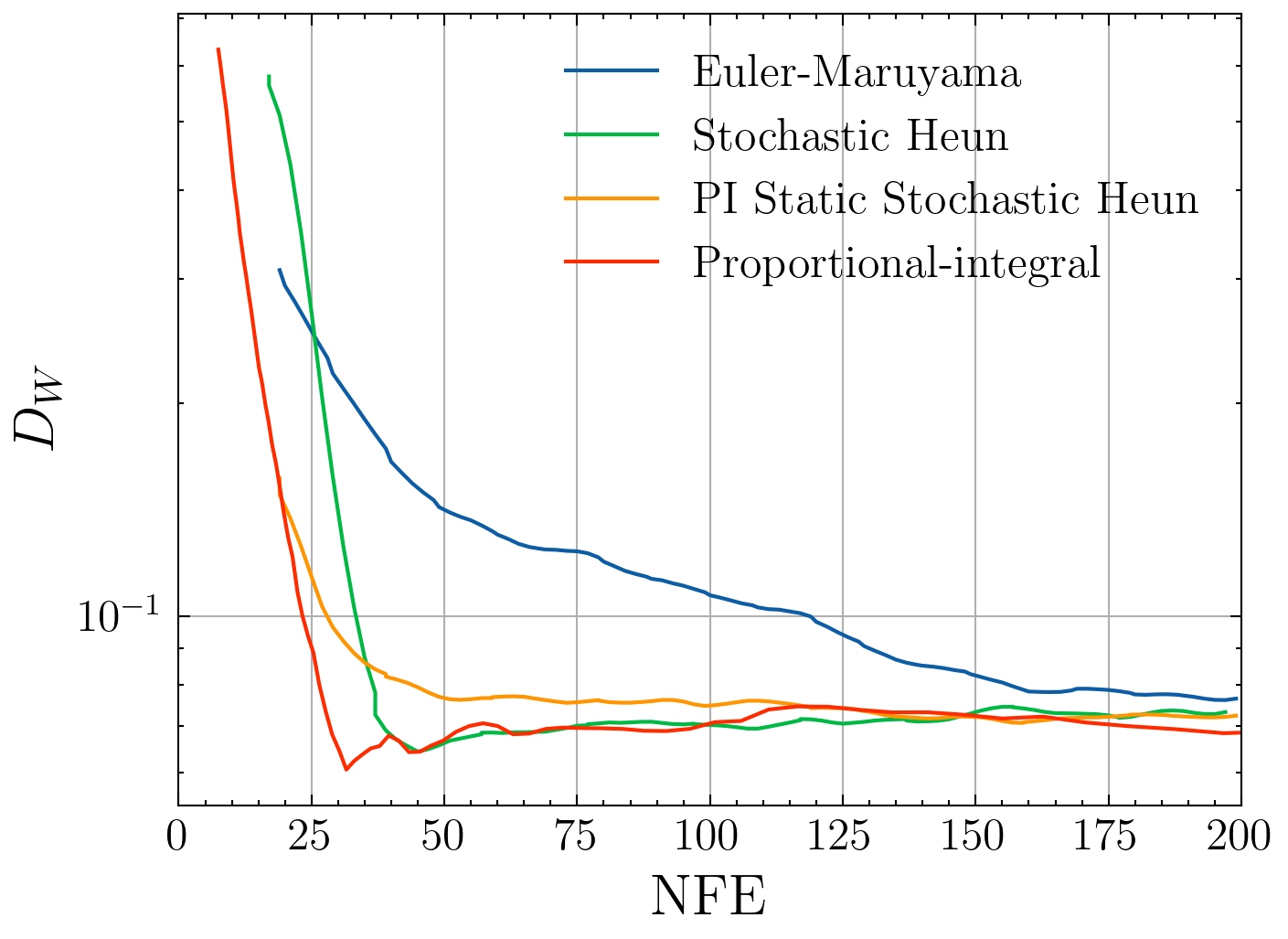}
        \caption{}
        \label{fig:complex_gaussian}
    \end{subfigure}
    \hfill
    \begin{subfigure}[b]{0.45\textwidth}
        \centering
        \includegraphics[width=0.97\linewidth]{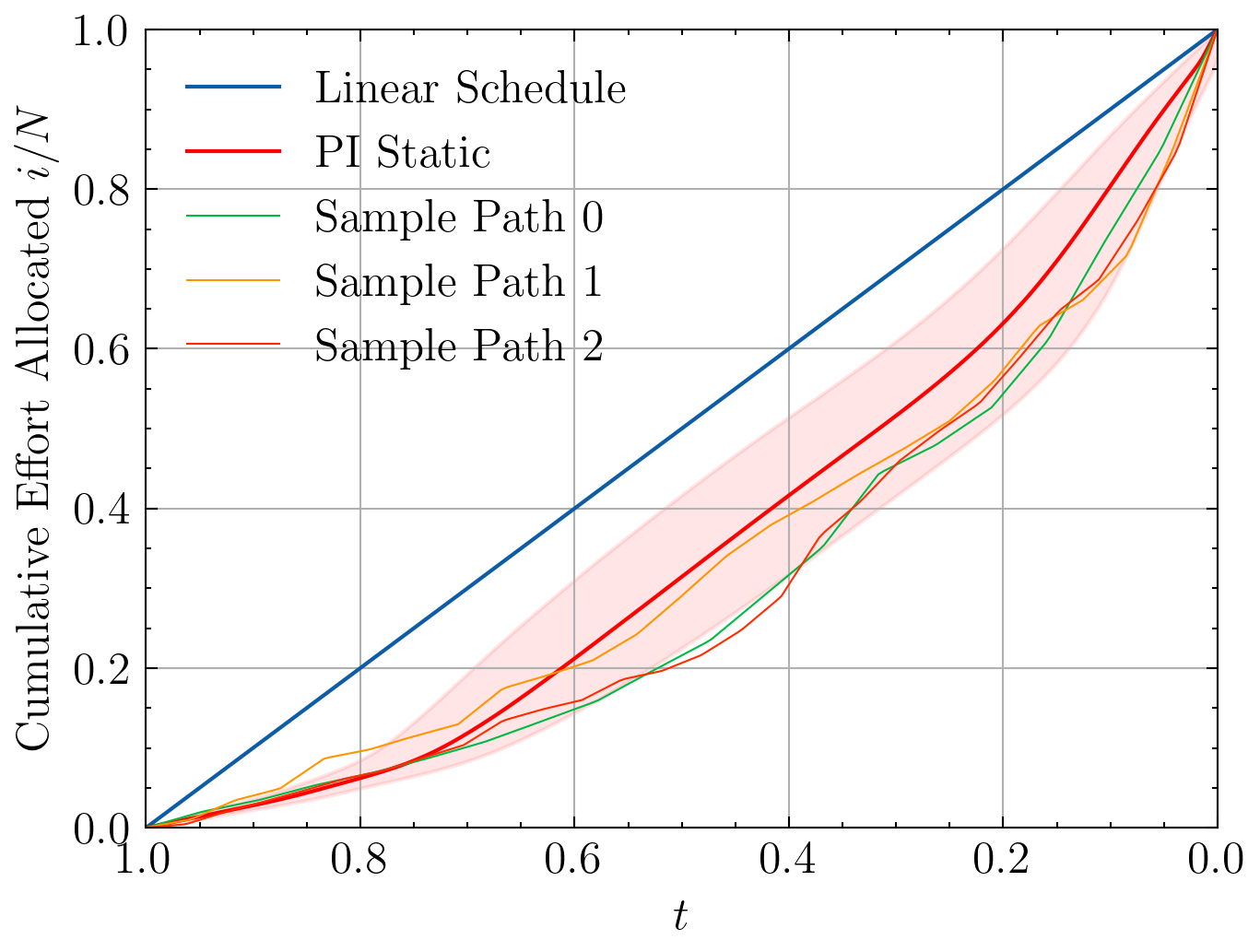}
        \caption{}
        \label{fig:complex_gaussian_trajectory}
    \end{subfigure}
    \caption{The Wasserstein distance against the NFE for different SDE solvers applied to diffusion on a mixture of Gaussians (a). The cumulative computational effort allocated ($\frac{i}{N}$) plotted against diffusion time $t$ at $\tau_{abs}\approx0.02656$ applied to diffusion on a Gaussian mixture (b).}
    \label{fig:1d_nfe_error_tradeoff}
\end{figure}

\autoref{fig:complex_gaussian} shows that the PI solver outperforms the other solvers by a large margin at low NFE. When NFE increases, the benefit over stochastic Heun's method with a linear schedule decreases. This can be explained by both samplers reaching the minimum Wasserstein distance, with PI converging more quickly. Stochastic Heun's with a PI static time schedule performs well at very low NFE ranges, but performance quickly degrades. \autoref{fig:complex_gaussian_trajectory} shows how our solver divides computational effort over the time interval. Note that there is substantial variance in the individual allocation. This likely explains the poor performance of the stochastic Heun with PI static configuration, and suggests that per-sample adaptivity is beneficial for this problem.

\subsection{FFHQ and ImageNet}
\label{sec:imagenet_experiments}
We evaluated the PI solver and the schedules it generates on FFHQ \citep{karras-2019} and ImageNet 64x64 \citep{russakovsky_imagenet_2015}. We compare it to the Euler-Maruyama \citep{kloeden_numerical_1992}, stochastic Heun \citep{rumelin_numerical_1982}, EDM-churn \citep{karras_elucidating_2022}, and the Gotta Go Fast (named after their paper) \citep{jolicoeur-martineau_gotta_2021} solvers. Notably, for each of these solvers, we run the steps from $t_{ode}$ onwards on the probability flow ODE. The SDE parametrization detailed in \autoref{app:edm_ode} was used. We used pretrained models from EDM and EDM2 \citep{karras_elucidating_2022, karras_analyzing_2024}, using the unconditional VE model for FFHQ and the ImageNet-64 EDM2-XL for ImageNet-64. The FID of the generated samples is evaluated at 49, 75, and 99 NFE, taking three separate samples of 50000 images to compute the mean and standard deviation. Note that for the adaptive samplers, $\tau_{rel}$ was altered to bring the average NFE to a value that rounds to the target NFE. For details on the process of obtaining hyperparameters, refer to \autoref{app:hyperparameter-selection}, and for the experimental setup, refer to \autoref{app:implementation_details}. Due to computational constraints, we prioritized other configurations over second-order sampler runs for the entropy schedule on ImageNet, as \citet{stancevic_entropic_2025} mention the entropic schedule has deteriorated performance for second-order samplers.

\begin{figure}
    \begin{subfigure}[c]{0.40\textwidth}
        \centering
        \includegraphics[width=0.99\linewidth]{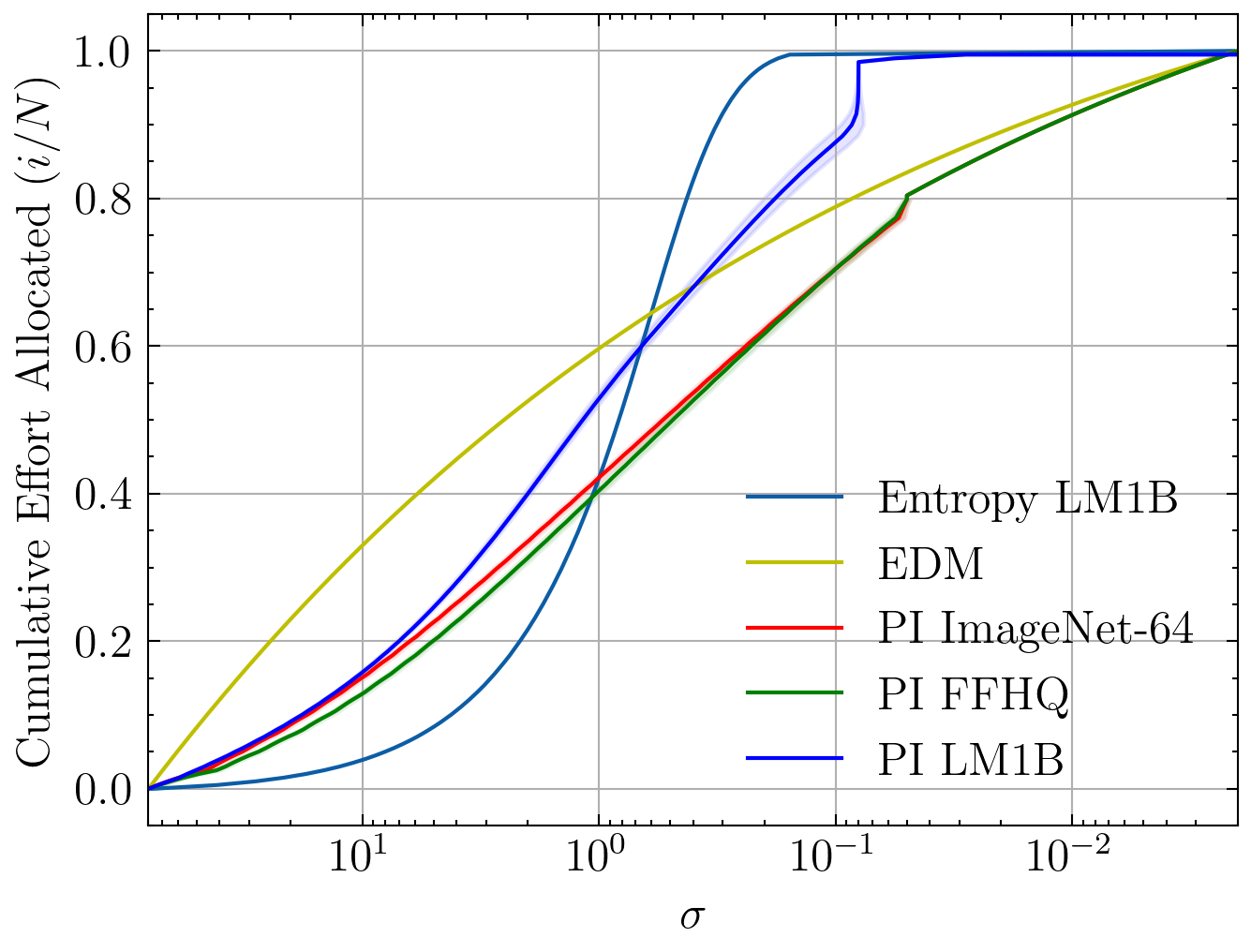}
        \caption{}
        \label{fig:image_pi_discretisation}
    \end{subfigure}
    \hfill
    \begin{subfigure}[c]{0.59\textwidth}
        \centering
        \includegraphics[width=0.99\linewidth]{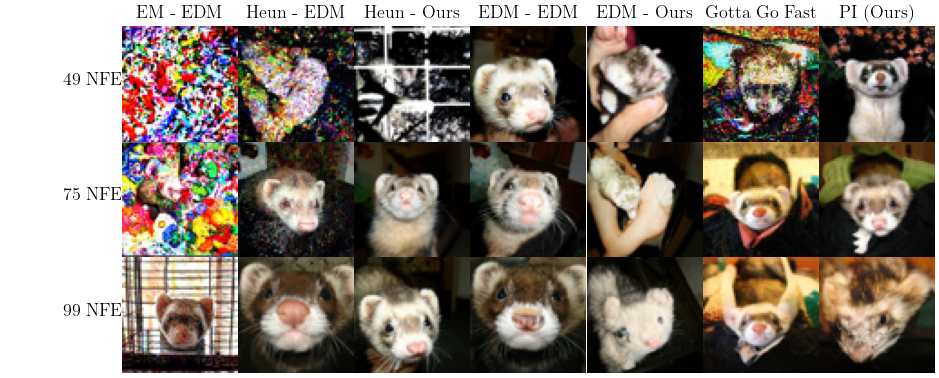}
        \caption{}
    \label{fig:imagenet_comparison}
    \end{subfigure}
    \caption{The cumulative effort allocation ($\frac{i}{N}$), plotted against diffusion noise $\sigma$, arrived at by the PI solver at 75 NFE for ImageNet-64 and FFHQ, and 127 NFE for LM1B, compared to the EDM schedule and LM1B entropic schedule (a). Comparison of ImageNet-64 samples generated using different solver-schedule combinations at 49, 75, and 99 NFE, with the same random seed (b).}
    \label{fig:image_comparison_and_discretisation}
\end{figure}

The results of this experiment can be found in \autoref{tab:sde_results}, with some samples found in \autoref{fig:imagenet_comparison}, and additional samples found in \autoref{app:samples}. For FFHQ, the PI static schedule clearly outperforms the EDM schedule across all NFE. This difference is less clear for ImageNet-64, although at 49 NFE, our schedule does perform better. Notably, our solver outperforms Gotta Go Fast by 2-5 times at 49 and 75 NFE. Additionally, for all the solvers except for EDM-churn, and in some cases stochastic Heun with our PI static schedule, our solver outperforms them in all cases for ImageNet-64, and some cases for FFHQ. Lastly, there is generally only a slight difference between the PI solver and stochastic Heun with a PI Static schedule, implying there is little per-sample benefit.

The gap between the EDM-churn solver and the pure SDE solver configurations (Euler-Maruyama, stochastic Heun, PI, Gotta Go Fast), can be explained by the sampling mechanisms. The EDM-churn solver samples from the probability flow ODE (\ref{eq:ode}), adding noise before the denoising. This noise has different magnitude and is added every second-order step, so once per two NFE. In contrast, the SDE configurations sample from \eqref{eq:rdp}, where noise is added every step, after denoising. This introduces more total noise, causing the solvers to require more steps to fully converge to high-quality images. The stochastic Heun results confirm this, as both use a Heun's method for sampling, only differing in the way noise is added. 

The average cumulative computational effort allocation of the PI static configuration for both datasets can be found in \autoref{fig:image_pi_discretisation} (refer to \autoref{subsec:bitstream} for comparison of the image schedules with language schedules). Firstly, it is clear that most individual discretizations lie very close to the average, as is indicated by the almost unnoticeably small error bars. Furthermore, the average discretizations deviate greatly from the EDM. Primarily, the PI discretizations take larger steps at high noise levels, while taking smaller steps than EDM at low noise levels. The low per-sample variance likely explains why there is only a slight difference between stochastic Heun with a PI static schedule and the PI solver, and only in ImageNet-64.

We performed a grid search over $\tau_{rel}$ and $\tau_{abs}$ on FFHQ, to determine the relation between those values, NFE, and image quality (see \autoref{app:hyperparameter-selection} for details). We measured image quality with the MIND metric \citep{berthet_mind_2026}, as it is more reliable at low sample size. The grid search indicates that both sampling speed and quality are largely independent of $\tau_{abs}$ when $\tau_{abs}$ is small, supporting the claim that $\tau_{rel}$ is the primary method of selecting a point on the NFE-FID tradeoff. Additionally, MIND seems to be fairly consistent in the $\tau_{rel}<1$ regime. This is likely caused by MIND not being sensitive enough at low sample sizes.

\begin{table}
\centering
\caption{The mean FID ($\downarrow$) and standard deviation of a number of solver-schedule combinations at different NFE for ImageNet-64 and FFHQ.}
\label{tab:sde_results}
\scriptsize
\setlength{\tabcolsep}{3pt}
\begin{tabular}{llllllll}\toprule
\multicolumn{1}{c}{\multirow{2}{*}{Solver}} & \multicolumn{1}{c}{\multirow{2}{*}{Schedule}} & \multicolumn{3}{c}{FFHQ}                      & \multicolumn{3}{c}{ImageNet 64x64}            \\\cmidrule(lr){3-5}\cmidrule(l){6-8}
\multicolumn{1}{c}{}                        & \multicolumn{1}{c}{}                          & 49 NFE        & 75 NFE        & 99 NFE        & 49 NFE        & 75 NFE        & 99 NFE        \\\midrule
\multirow{2}{*}{Euler-Marayuma}   & EDM                                           & $318.96\!\pm\!0.07$        & $171.55\!\pm\!0.27$        & $91.83\!\pm\!0.17$         & $187.66\!\pm\!0.23$        & $101.48\!\pm\!0.40$        & $58.23\!\pm\!0.15$         \\
                                  & Entropy                                       & $226.70\!\pm\!0.20$        & $81.33\!\pm\!0.20$         & $43.98\!\pm\!0.20$         & $90.54\!\pm\!0.31$         & $36.55\!\pm\!0.06$         & $18.65\!\pm\!0.03$         \\\midrule
\multirow{3}{*}{Stochastic Heun}  & EDM                                           & $274.33\!\pm\!0.23$        & $23.55\!\pm\!0.14$         & $3.58\!\pm\!0.05$          & $167.75\!\pm\!0.23$        & $20.98\!\pm\!0.26$         & $2.71\!\pm\!0.02$          \\
                                  & Entropy                                       & $348.81\!\pm\!0.15$        & $3.26\!\pm\!0.02$          & $3.54\!\pm\!0.05$          & N/A                        & N/A                        & N/A                        \\
                                  & PI Static (ours)                              & $137.69\!\pm\!0.21$        & $3.26\!\pm\!0.06$          & $3.22\!\pm\!0.07$          & $66.93\!\pm\!1.10$         & $\mathbf{1.48}\!\pm\!0.02$ & $\mathbf{1.59}\!\pm\!0.02$ \\\midrule
\multirow{3}{*}{EDM-churn}        & EDM                                           & $3.26\!\pm\!0.02$          & $2.97\!\pm\!0.03$          & $2.85\!\pm\!0.03$          & $1.58\!\pm\!0.04$          & $1.58\!\pm\!0.02$          & $1.62\!\pm\!0.04$          \\
                                  & Entropy                                       & $2.76\!\pm\!0.02$          & $2.79\!\pm\!0.06$          & $2.90\!\pm\!0.02$          & N/A                        & N/A                        & N/A                        \\
                                  & PI Static (ours)                              & $\mathbf{2.67}\!\pm\!0.02$ & $\mathbf{2.66}\!\pm\!0.02$ & $\mathbf{2.81}\!\pm\!0.05$ & $\mathbf{1.49}\!\pm\!0.03$ & $1.56\!\pm\!0.02$          & $1.65\!\pm\!0.04$          \\\midrule
\multicolumn{1}{l}{Gotta Go Fast} & Adaptive                                      & $261.84\!\pm\!0.10$        & $7.73\!\pm\!0.03$          & $3.10\!\pm\!0.03$          & $181.46\!\pm\!5.85$        & $7.05\!\pm\!0.08$          & $2.50\!\pm\!0.11$          \\\midrule
\multicolumn{1}{l}{PI (ours)}     & Adaptive                                      & $83.95\!\pm\!0.27$         & $3.32\!\pm\!0.06$          & $3.14\!\pm\!0.05$          & $37.27\!\pm\!0.27$         & $\mathbf{1.48}\!\pm\!0.00$ & $1.61\!\pm\!0.01$          \\\bottomrule
\end{tabular}
\end{table}
\begin{figure}
    \begin{subfigure}[b]{0.45\textwidth}
        \centering
        \includegraphics[width=0.99\linewidth]{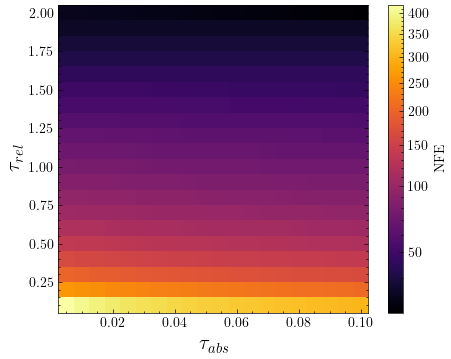}
        \caption{}
        \label{fig:nfe_grid_main}
    \end{subfigure}
    \hfill
    \begin{subfigure}[b]{0.45\textwidth}
        \centering
        \includegraphics[width=0.99\linewidth]{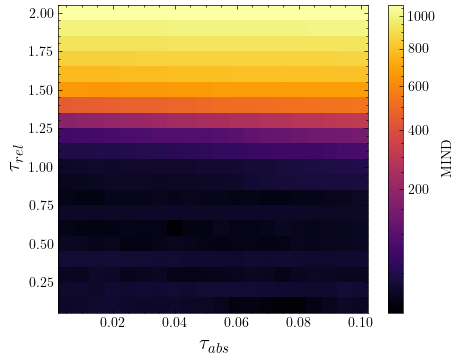}
        \caption{}
        \label{fig:mind_grid_main}
    \end{subfigure}
    \caption{The NFE (a) and the image quality measured by MIND (b) as a function of $\tau_{rel}$ and $\tau_{abs}$ measured by generating 64 ImageNet-64 images at each combination of $\tau_{rel}, \tau_{abs}$.}
    \label{fig:tolerance_grid_main}
\end{figure}

\subsection{Bitstream Language Diffusion}
\label{subsec:bitstream}
To determine how our solver transfers between modalities, we applied the PI solver to language modeling using the Continuous Bitstream Diffusion framework \citep{batzolis_cobit_2026}, which models each token as a series of analogue bits, on which continuous diffusion is applied. We compare it to \citet{batzolis_cobit_2026}'s stochastic DDIM configurations and EDM with churn \citep{karras_elucidating_2022}, both using the EDM, entropic, and PI static schedules. We use the pretrained models and evaluation pipeline provided by \citet{batzolis_cobit_2026}. We measure the generative perplexity and unigram token entropy at varying NFE on the LM1B dataset \citep{chelba_one_2014}, generating 1024 strings of 128 tokens. For details on the hyperparameters, refer to \autoref{app:hyperparameter-selection}, and for experimental details, refer to \autoref{app:implementation_details}. The results of this experiment can be found in \autoref{tab:lm1b_results}. Notably, our solver outperforms the EDM-churn solver in terms of perplexity at all NFE. Additionally, at 63 and 127 NFE, our solver outperforms all configurations in terms of perplexity. This improvement in perplexity does come at the cost of a lower unigram entropy, implying our samples have slightly less variance. Our PI static schedule performs better than the entropic schedule for the second-order EDM sampler, but performs worse than the entropic schedule for the first-order DDIM sampler. Additionally, it always outperforms the handcrafted EDM schedule. Interestingly, per-sample adaptivity seems to be relevant for this dataset.

\autoref{fig:image_pi_discretisation} shows the PI static schedule generated on LM1B. Note that it is substantially different from both the EDM schedule, and the schedules for image generation. It does more closely resemble the entropy schedule specifically computed for this dataset, and allocates the most effort along the center of the SDE trajectory.

\begin{table}
\caption{The perplexity ($\downarrow$) and unigram entropy ($\uparrow$) of solver-schedule combination at different NFE for LM1B.}
\label{tab:lm1b_results}
\centering
\scriptsize
\setlength{\tabcolsep}{3pt}
\begin{tabular}{clllllll}\toprule
\multirow{2}{*}{Sampler}   & \multicolumn{1}{c}{\multirow{2}{*}{Schedule}} & \multicolumn{2}{c}{63 NFE}      & \multicolumn{2}{c}{127 NFE}    & \multicolumn{2}{c}{255 NFE}    \\\cmidrule(lr){3-4}\cmidrule(lr){5-6}\cmidrule(l){7-8}
                           & \multicolumn{1}{c}{}                          & Perplexity      & Entropy       & Perplexity     & Entropy       & Perplexity     & Entropy       \\\midrule
\multicolumn{1}{l}{\multirow{3}{*}{DDIM}} & Entropic                                      & $128.26\!\pm\!1.62$          & $4.35\!\pm\!0.00$          & $87.03\!\pm\!0.28$          & $4.34\!\pm\!0.01$          & $\mathbf{60.02}\!\pm\!0.83$ & $4.31\!\pm\!0.00$          \\
\multicolumn{1}{l}{}                      & EDM                                           & $233.26\!\pm\!3.13$          & $\mathbf{4.36}\!\pm\!0.01$ & $159.14\!\pm\!3.36$         & $\mathbf{4.35}\!\pm\!0.00$ & $109.00\!\pm\!0.31$         & $\mathbf{4.34}\!\pm\!0.01$ \\
\multicolumn{1}{l}{}                      & PI Static (ours)                              & $165.56\!\pm\!0.64$          & $4.35\!\pm\!0.00$          & $121.67\!\pm\!0.68$         & $4.34\!\pm\!0.01$          & $83.46\!\pm\!0.77$          & $4.33\!\pm\!0.01$          \\\midrule
\multirow{3}{*}{EDM-churn}                & Entropic                                      & $152.85\!\pm\!1.37$          & $4.33\!\pm\!0.00$          & $115.95\!\pm\!0.97$         & $4.33\!\pm\!0.00$          & $84.84\!\pm\!0.66$          & $4.31\!\pm\!0.01$          \\
                                          & EDM                                           & $214.76\!\pm\!0.29$          & $4.33\!\pm\!0.01$          & $156.30\!\pm\!2.05$         & $4.33\!\pm\!0.01$          & $116.49\!\pm\!1.59$         & $4.32\!\pm\!0.01$          \\
                                          & PI Static (ours)                              & $121.87\!\pm\!0.85$          & $4.33\!\pm\!0.01$          & $95.18\!\pm\!0.75$          & $4.32\!\pm\!0.00$          & $69.87\!\pm\!0.18$          & $4.30\!\pm\!0.01$          \\\midrule
\multicolumn{1}{l}{PI}                    & Adaptive                                      & $\mathbf{107.04}\!\pm\!1.11$ & $4.26\!\pm\!0.01$          & $\mathbf{83.75}\!\pm\!0.18$ & $4.27\!\pm\!0.00$          & $62.33\!\pm\!0.35$          & $4.27\!\pm\!0.00$          \\\bottomrule
\end{tabular}
\end{table}

\subsection{Ablations}
\paragraph{PI Solver Ablations} We ran ablations to determine the impact of the changes made to the PI solver compared to the original design by \citep{ilie_adaptive_2015} and the Gotta Go Fast solver \citep{jolicoeur-martineau_gotta_2021}. These were run on FFHQ. The results of this can be found in \autoref{tab:pi_ablations}. Notably, configuration A, used in the other experiments, outperforms all the other configurations. Additionally, both of the alternate tolerance options (configurations C and D) perform substantially worse than our final solver. Lastly, there is some performance gain of PI control over integral-only control, although not as large as for the other alterations. Details and hyperparameters used to obtain these ablations can be found in \autoref{app:implementation_details}. 

\paragraph{Schedule Across Modalities} To show that schedules are dataset-specific, we compare the schedules generated on FFHQ with those on LM1B. We do this comparison on both datasets. The results of this can be found in \autoref{tab:modality_ablations}. We observe that the native schedules always obtain better performance than the foreign ones. This benefit is highlighted at the low NFE range, with the native FFHQ schedule obtaining a 3$\times$ improvement over the LM1B schedule. For implementation details, refer to \autoref{app:implementation_details}.

\begin{table}[]
\centering
\caption{PI solver ablations on FFHQ, measuring FID ($\downarrow$) of the PI solver at different NFE with varying components changed or removed.}
\label{tab:pi_ablations}
\scriptsize
\setlength{\tabcolsep}{3pt}
\begin{tabular}{lllll}\toprule
Configuration   & Components                            & 49 NFE         & 75 NFE        & 99 NFE        \\\midrule
Config A (base) & PI Control + Mixed Noise Tolerance    & \textbf{84.09} & \textbf{3.36} & \textbf{3.11} \\
Config B        & I Control + Mixed Noise Tolerance     & 125.81         & 4.04          & 3.53          \\
Config C        & PI Control + Mixed Absolute Tolerance & 149.23         & 66.7          & 49.77         \\
Config D        & PI Control + Non-Mixed Tolerance      & 357.64         & 319.14        & 298.59        \\\bottomrule
\end{tabular}
\end{table}

\begin{table}[!htb]
      \caption{A PI schedule generated by the PI solver on FFHQ (images) compared to one generated on LM1B (language), evaluated in terms of FID ($\downarrow$) at different NFE on FFHQ (a) and in terms of generative perplexity at different NFE on LM1B (b).}
      \label{tab:modality_ablations}
    \begin{minipage}{.5\linewidth}
      \subcaption{}
      \centering
\scriptsize
\setlength{\tabcolsep}{2pt}
\begin{tabular}{lllll}\toprule
\multicolumn{1}{c}{\multirow{2}{*}{Solver}} & \multicolumn{1}{c}{\multirow{2}{*}{Schedule}} & \multicolumn{3}{c}{FFHQ}    \\\cmidrule(lr){3-5}
\multicolumn{1}{c}{}                        & \multicolumn{1}{c}{}                          & 49 NFE        & 75 NFE        & 99 NFE        \\\midrule
\multirow{2}{*}{Stochastic Heun}                             & PI FFHQ (Native)                              & 137.69        & 3.26          & 3.22          \\
                                            & PI LM1B (Foreign)                             & 388.45        & 337.05        & 3.43          \\\midrule
\multirow{2}{*}{EDM-churn}                  & PI FFHQ (Native)                              & \textbf{2.67} & \textbf{2.66} & \textbf{2.81} \\
                                            & PI LM1B (Foreign)                             & 2.83          & 2.83          & 2.98          \\\bottomrule
\end{tabular}
    \end{minipage}%
    \begin{minipage}{.5\linewidth}
      \centering
      \subcaption{}
\scriptsize
\setlength{\tabcolsep}{2pt}
\begin{tabular}{lllll}\toprule
\multicolumn{1}{c}{\multirow{2}{*}{Solver}} & \multicolumn{1}{c}{\multirow{2}{*}{Schedule}} & \multicolumn{3}{c}{LM1B}        \\\cmidrule(lr){3-5}
\multicolumn{1}{c}{}                        & \multicolumn{1}{c}{}                          & 63 NFE          & 127 NFE        & 255 NFE        \\\midrule
\multirow{2}{*}{Stochastic DDIM}            & PI LM1B (Native)                              & 165.56          & 121.67         & 83.46          \\
                                            & PI FFHQ (Foreign)                             & 184.29          & 125.92         & 87.73          \\\midrule
\multirow{2}{*}{EDM-churn}                  & PI LM1B (Native)                              & \textbf{121.87} & \textbf{95.18} & \textbf{69.87} \\
                                            & PI FFHQ (Foreign)                             & 127.79          & 99.35          & 74.35          \\\bottomrule
\end{tabular}
    \end{minipage} 
\end{table}

%\begin{figure}
%    \centering
%    \includegraphics[width=0.8\linewidth]{figures/image_samples.png}
%    \caption{Comparison of ImageNet-64 samples generated using different solver-schedule combinations at 49, 75, and 99 NFE, with the same random seed}
%    \label{fig:imagenet_comparison}
%\end{figure}

%% file: sections/body/6-discussion.tex
\section{Conclusion}
\label{chap:discussion}
We adapted and applied a proportional-integral (PI) solver from the numerical analysis literature \citep{ilie_adaptive_2015} to generative diffusion, introduced an error estimator based on the diffusion noise level, and developed a technique to aggregate the PI solver's individual adaptive step trajectories into a static discretization. Our solver can be used to generate second-order compatible schedules in a data-driven manner, and these improve on the existing handcrafted EDM schedule in several of our evaluated settings.

\paragraph{Limitations} While we did evaluate the influence of $\tau_{rel}$ and $\tau_{abs}$ on the NFE-quality tradeoff, no full search over the hyperparameter space was conducted. Moreover, all image and language experiments were conducted using VE models. The 1D experiments were done with a VP SDE and show a significant benefit of adaptivity. It is unclear if this is a property of the VP SDE or of the specific dataset. Additionally, these experiments were only conducted for stochastic sampling. Preliminary testing indicates the PI solver can also be applied to deterministic sampling, but it is left to future work to investigate if it outperforms existing sampler-schedule combinations. This could also allow the solver to discover the final ODE steps by itself, instead of requiring an existing static schedule for the last steps.

%% file: sections/body/7-AI_statement.tex
\section{AI Use Statement}
In this work, we have used generative AI to provide feedback on the research methodology, asking it to specifically highlight potentially important additional ablations. Moreover, we used it to provide feedback on the manuscript in terms of writing style and consistency. Lastly, we used AI to help with identifying the cause of bugs in the code. Generative AI was not used for any other purposes in this paper. All of the feedback provided by LLMs was first manually checked for correctness and usefulness before applying the feedback. No AI-generated code ended up in the final codebase, so AI-generated code did not have to be checked for correctness. We take responsibility for the final content of this work, including text, claims or artifacts produced with the aid of generative AI.

%% file: sections/body/9-reproducibility_statement.tex
\section{Reproducibility Statement}
To make the paper reproducible, we provide a link to the github repository used to obtain the results of this paper in the abstract. Additionally, \autoref{chap:adaptive-solver} details the PI solver we adapted, with \autoref{app:solver_details} providing additional detail. Furthermore, \autoref{app:hyperparameter-selection} provides detail on how the hyperparameters were obtained from scratch, with \autoref{app:implementation_details} providing the detail required to reconstruct the results of this paper without requiring access to the codebase. 

%% file: sections/appendices/edm_forward_sde.tex
\section{Reparametrizing the EDM Probability Flow ODE}
\label{app:edm_ode}
To be able to use the EDM schedule using the formulation of the reverse SDE (\ref{eq:rdp}), the parametrization \citet{karras_elucidating_2022} provide needs to be rewritten to yield a drift and diffusion compatible with the forward SDE.
\begin{equation}
  \mathrm{d}\mathbf{x} =  \mathbf{f}(\mathbf{x}, t)\mathrm{d}t + g(t)\mathrm{d}\mathbf{w}  
\end{equation}
With $\mathbf{f}(\mathbf{x}, t):\mathbb{R}^d\times \mathbb{R}\rightarrow \mathbb{R}^d$ the drift function, $g(t): \mathbb{R}\rightarrow \mathbb{R}$ the diffusion function, and $\mathbf{w}(t)$ the standard Wiener process.

\citet{karras_elucidating_2022} define the probability flow ODE:
\begin{equation}
\mathrm{d}\mathbf{x} = \left[\frac{\dot{s}(t)}{s(t)}\mathbf{x} - s(t)^2\dot{\sigma}(t)\sigma(t)\nabla_\mathbf{x} \log p_t\left(\frac{\mathbf{x}}{s(t)}\right)\right]\mathrm{d}t    
\end{equation}
With $s(t)\in \mathbb{R} \rightarrow \mathbb{R}$ a scaling function, $\sigma(t)\in \mathbb{R} \rightarrow \mathbb{R}$ a function indicating the noise at a time $t$. This is an alternate formulation of \eqref{eq:ode}, assuming $s(t)=1$:
$$\mathrm{d}\mathbf{x} =[\mathbf{f}(\mathbf{x}, t) - \frac{1}{2}g(t)^2\nabla_\mathbf{x} \log p_t(\mathbf{x})]\mathrm{d}t$$
With $\nabla_\mathbf{x} \log p_t(\mathbf{x})$ the score. To obtain expressions for the drift and diffusion to plug in the forward SDE, one can set:
\begin{equation}
    \mathbf{f}(\mathbf{x}, t) = \frac{\dot{s}(t)}{s(t)}\mathbf{x} \quad \text{and} \quad \frac{1}{2}g(t)^2 = s(t)^2\dot{\sigma}(t)\sigma(t)
\end{equation}
Solving for $g(t)$ yields:
\begin{equation}
    g(t) = s(t)\sqrt{2\dot{\sigma}(t)\sigma(t)}
\end{equation}
\citet{karras_elucidating_2022} define for the EDM framework the parameters $s(t) = 1$ and $\sigma(t) = t$. The time derivatives of these are
$$\dot{s}(t) = 0 \quad \text{and} \quad \dot{\sigma}(t)=1$$ 
These can be used to obtain the drift and diffusion of the EDM parametrization:
\begin{equation}
    \mathbf{f}(\mathbf{x}, t) = 0 \quad \text{and} \quad g(t) = \sqrt{2t}
\end{equation}
Which can then be filled into the forward SDE to obtain the EDM SDE:
\begin{equation}
    \label{eq:edmsde}
    \mathrm{d}\mathbf{x} = \sqrt{2t}\mathrm{d}\mathbf{w}
\end{equation}
This SDE can then be used together with the variance-exploding denoising networks provided by EDM and EDM2 \citep{karras_elucidating_2022, karras_analyzing_2024} by using the derived values for the drift and diffusion.

%% file: sections/appendices/solver_details.tex
\section{Solver Details}
\label{app:solver_details}

\subsection{Error Estimator}
The error estimator given by \eqref{eq:error_1} differs from the one defined by \citet{ilie_adaptive_2015}. Here, we detail how we arrived at our error estimator. \citet{ilie_adaptive_2015} define the error as:
\begin{equation}
    \label{eq:error_ilie}
    \epsilon(\mathbf{x}_i, h_i) = \sqrt{\frac{1}{d}\sum_{k=1}^d \left(\frac{\mathbf{x}_{i+1, k}' - \mathbf{x}_{i+1, k}''}{\tau_{abs}}\right)^2}
\end{equation}
Where $\epsilon(\mathbf{x}_i, h_i)\in\mathbb{R}_{>0}$ the local extrapolated error, and $\tau_{abs} \in \mathbb{R}_{>0}$ a user-defined tolerance parameter which scales the error. 

We implemented this error estimator, which performed well for the 1D experiments, which use the Variance Preserving (VP) SDE. However, when we applied this to the image datasets, for which the models were trained using the Variance Exploding (VE) SDE instead, this error estimator struggled to achieve low NFE, and low NFE results were of low quality (see \autoref{tab:pi_ablations}).

To solve this problem, a relative tolerance parameter was introduced. This is a common parameter in ODE literature \citep{noauthor_solving_1993}, that takes into account the scale of the differential equation. This is specifically relevant when the scale of the problem varies largely, which is exactly the case in the VE formulation of diffusion. Generally, the scale of the problem is accounted for by multiplying the relative tolerance by the absolute value of each dimension. The mixed absolute tolerance is then defined as:
\begin{equation}
    \label{eq:abs_tolerance}
    \mathbf{\tau}(\mathbf{x}') = \tau_{abs} + \tau_{rel} |\mathbf{x}'|
\end{equation}
Where the absolute value of each element in $\mathbf{x}$ is calculated elementwise, and $\tau_{abs}\in \mathbb{R}_{>0}$ is the absolute tolerance, which is equivalent to the tolerance in \eqref{eq:error_1}. $\tau_{rel}$ is the relative tolerance.

In the diffusion process, the noise at each time point is a known quantity, which also indicates the scale of each dimension. Therefore, it was hypothesised that instead of normalizing each dimension by its absolute value, we could normalise by the noise level. This yields the equation for the mixed noise tolerance:
\begin{equation}
    \label{eq:mixed_noise_tol}
    \tau(t) = \tau_{abs} + \tau_{rel} \sigma(t)
\end{equation}
Where $\sigma(t)\in \mathbb{R}_{>0}$ indicates the noise at time $t$. Ablations showed that the mixed noise tolerance generally required less NFE for similar sample quality compared to the absolute noise tolerance and the non-mixed tolerance. Therefore, the mixed noise tolerance is used in all experiments.

\subsection{Step Rejection}
\citet{ilie_adaptive_2015} also define a step-rejection mechanism in their paper. They define that the error should satisfy the following condition:
\begin{equation}
    \label{eq:reject-condition}
    \epsilon(\mathbf{x}_i, h_i) \leq 1
\end{equation}
If this condition is not satisfied, step $h_i$ is rejected, and a new step $h_{i+1}$ is tried. As the tolerance directly scales the error, setting a smaller tolerance effectively sets the error threshold higher, making steps more likely to be rejected, while a larger tolerance has the opposite effect; it makes the solver more permissive and less likely to reject steps.

However, we found that rejected steps generally led to wasted NFE. The best results were obtained when the rejection rate was $0$, and disabling the rejection mechanism led to the same or better quality results. Furthermore, for the 1D experiments, disabling step rejection improved the results. Therefore, we decided to disable step rejection for the final experiments. 

\subsection{Batches}
To efficiently make use of the GPU, most diffusion implementations apply the reverse diffusion process in batches. As the solver is intended to have different step-sizes for each element of the batch, certain elements of the batch can finish sampling ($t=t_0$) before others. This is handled by each element in the batch having its own independent time. To ensure that these elements are not unnecessarily passed into the score network, masking was used to remove finished elements from the batch, which was inspired by the approach of \citet{jolicoeur-martineau_gotta_2021}.

%% file: sections/appendices/hyperparemeter_selection.tex
\section{Hyperparameter Selection}
\label{app:hyperparameter-selection}

\begin{figure}[H]
    \begin{subfigure}[b]{0.49\textwidth}
        \centering
        \includegraphics[width=0.99\linewidth]{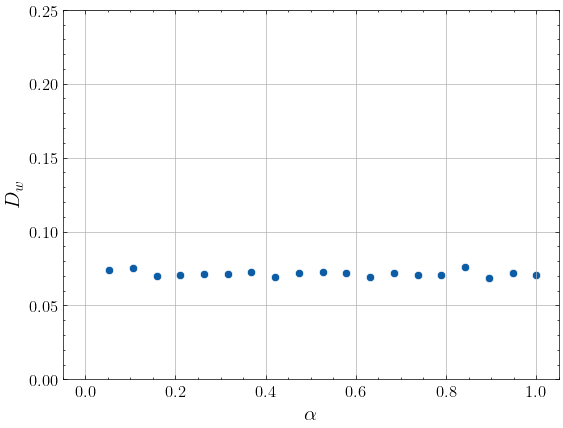}
        \caption{Varying $\alpha$}
    \end{subfigure}
    \hfill
    \begin{subfigure}[b]{0.49\textwidth}
        \centering
        \includegraphics[width=0.99\linewidth]{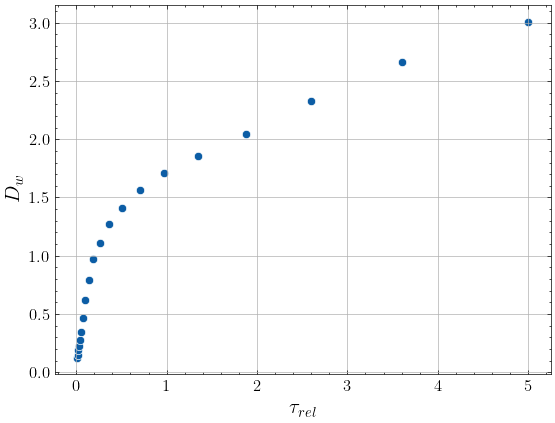}
        \caption{Varying $\tau_{rel}$}
    \end{subfigure}
    \begin{subfigure}[b]{0.49\textwidth}
        \centering
        \includegraphics[width=0.99\linewidth]{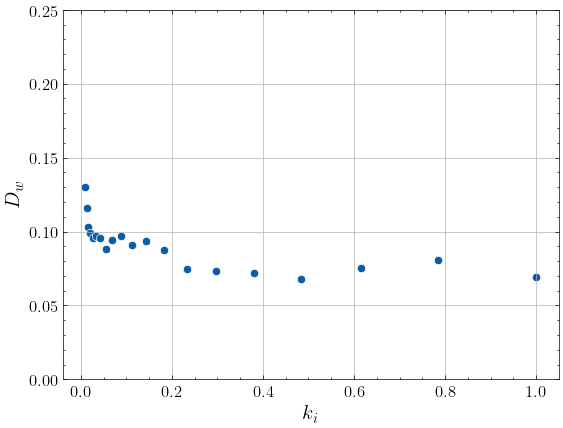}
        \caption{Varying $k_i$}
    \end{subfigure}
    \hfill
    \begin{subfigure}[b]{0.49\textwidth}
        \centering
        \includegraphics[width=0.99\linewidth]{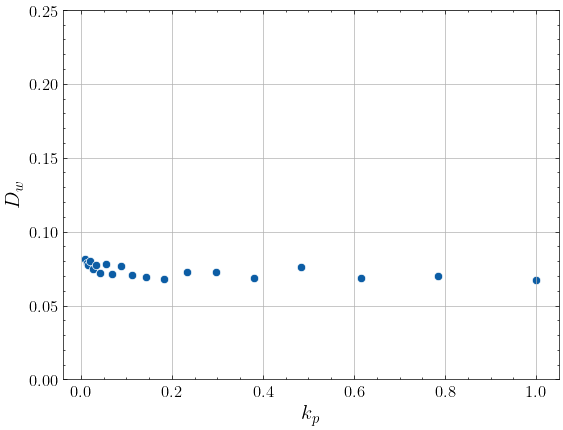}
        \caption{Varying $k_p$}
    \end{subfigure}
    \caption{The 1-dimensional sensitivity of the $\alpha$ (a), $\tau_{rel}$ (b), $k_i$ (c), and $k_p$ (d) hyperparameters of the PI solver with regards to the minimum Wasserstein distance of the Wasserstein distance - NFE tradeoff, evaluated on a 1D multimodal Gaussian with varying means, variances, and weights.}
    \label{fig:hyperparameters_1d}
\end{figure}

\subsection{1D Validation}
\label{appsec:1d_hyperparams}
In this subsection, we discuss how we determined the hyperparameters we started from in subsequent experiments, and assess influence of hyperparameters on the NFE-quality tradeoff. To assess this influence, each of the hyperparameters $\alpha$, $\tau_{rel}$, $h_0$, $k_p$, and $k_i$ were varied, while keeping the other hyperparameters constant. The setup described in \autoref{app:implementation_details} was used for this. Note that no grid search over these hyperparameters was performed, instead we performed a 1D sensitivity analysis. With the variation of $\tau_{abs}$ in addition to the other hyperparameters, doing a full evaluation over these hyperparameters would have a time complexity of $\mathcal{O}(n^6)$. With a decent resolution for each hyperparameter, this grid search was assessed to be infeasible. However, the one-dimensional sensitivity of each hyperparameter still gives information about the stability of the PI solver. The minimum Wasserstein Distance as a function of either $\alpha$, $\tau_{rel}$, $k_p$, and $k_i$, on the 1D multimodal Gaussian with varying means, variances, and weights can be seen in \autoref{fig:hyperparameters_1d}. For details on the default settings for all hyperparameters that were not varied, and for details on the Gaussian, refer to \autoref{app:implementation_details}. Note that $h_0$ is absent here, as its minimum error plot did not reflect the NFE - Wasserstein distance tradeoff as well as the others. This will be further discussed later on in this subsection.

It can be seen in \autoref{fig:hyperparameters_1d} that the solver is insensitive to $\alpha$. The Wasserstein distance barely varies between different $\alpha$. The main thing that $\alpha$ influences is that NFE becomes more sensitive to $\tau_{abs}$ at lower $\alpha$. However, the Wasserstein Distance - NFE tradeoff remains the same, implying $\alpha$ does not influence the tradeoff. There is some influence of $\tau_{rel}$. Mainly, larger values of $\tau_{rel}$ increase the Wasserstein distance. This is likely caused by the scale of the variance preserving SDE not varying enough to get good use of scaling the error by the noise level, and therefore, $\tau_{rel}$ primarily acts as a direct addition to $\tau_{abs}$, increasing the lowest Wasserstein Distance purely by having the tolerance be consistently larger. Therefore, for this 1D experiment, $\tau_{rel}$ is best kept at 0. The controller seems fairly insensitive to $k_i$. Mainly, very low $k_i$ ($k_i<0.05$) seems to be detrimental for performance. Larger values do not have much influence. Finally, the controller seems to be insensitive to $k_p$. Therefore, the literature values of $k_i=0.3$, $k_p=0.1$, $\alpha=0.9$ seem to be decent settings for the PI solver on 1D Gaussians. Furthermore, $\tau_{rel}$ should be kept at $0$ for this experiment.

\begin{figure}
    \centering
    \includegraphics[width=0.75\linewidth]{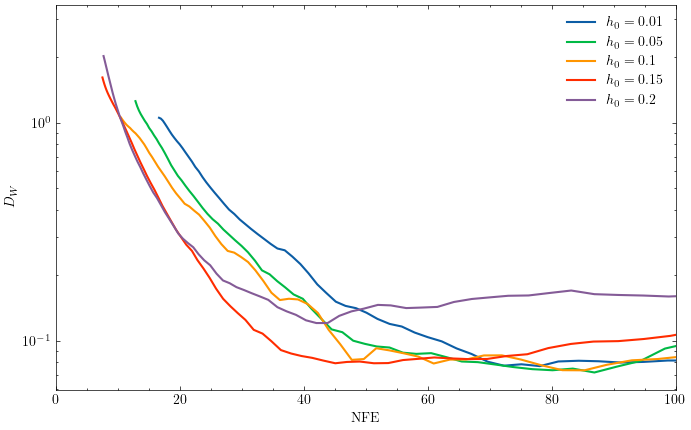}
    \caption{Influence of varying $h_0$ on the NFE - Wasserstein distance trade-off of the PI solver on a 1D multimodal Gaussian with varying means, variances, and weights.}
    \label{fig:h_start_varying}
\end{figure}

The influence of $h_0$ on the NFE - Wasserstein distance tradeoff cannot be easily reflected in plots just showing the minimum Wasserstein distance in the entire tradeoff. Mainly, choosing $h_0$ correctly has a large influence on how many steps the solver needs to converge to its optimal Wasserstein distance. To show this influence, consider \autoref{fig:h_start_varying}. In the figure, it can be seen that at low $h_0$, the Wasserstein distance reduces more slowly as a function of NFE, compared to at higher $h_0$. However, if $h_0$ is too large, which is the case at $h_0=0.2$, the minimum Wasserstein distance the solver reaches is larger than it can be if $h_0$ is chosen more optimally. Additionally, it can be seen that with $h_0=0.15$ the Wasserstein distance reduces quickly with NFE, while performance stabilizes at the same Wasserstein distance as the smaller $h_0$'s. Therefore, $h_0=0.15$ was chosen for the 1D experiments. Lastly, note that the influence of $h_0$ on the tradeoff is significant. It is therefore an important parameter to determine a good value for.

\subsection{Image Experiments}
 \label{appsec:image_hyperparms}
\subsubsection{PI Solver}
After concluding testing of the PI solver on 1D multimodal Gaussians, it was applied to ImageNet-64. Initially, all hyperparameters were transferred from the 1D experiment to the ImageNet-64 experiment. However, it was quickly apparent that $\tau_{abs}$, $\tau_{rel}$, and $h_0$ needed significant tweaking. The solver did work with the initial hyperparameters. It just took 750 NFE to generate an image. The main goal of hyperparameter optimization was thus to bring NFE down to a number more reasonable for modern diffusion, between 50 and 100. 

\begin{figure}
    \begin{subfigure}[b]{0.49\textwidth}
        \centering
        \includegraphics[width=0.99\linewidth]{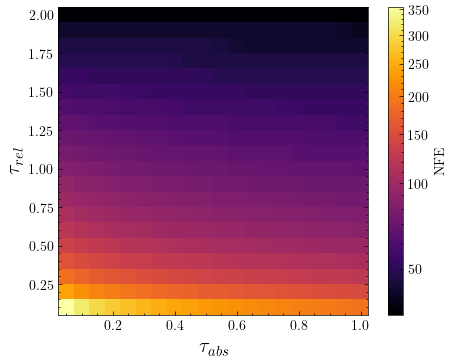}
        \caption{NFE as a function of $\tau_{abs}, \tau_{rel}$.}
        \label{fig:nfe_grid_initial}
    \end{subfigure}
    \hfill
    \begin{subfigure}[b]{0.49\textwidth}
        \centering
        \includegraphics[width=0.99\linewidth]{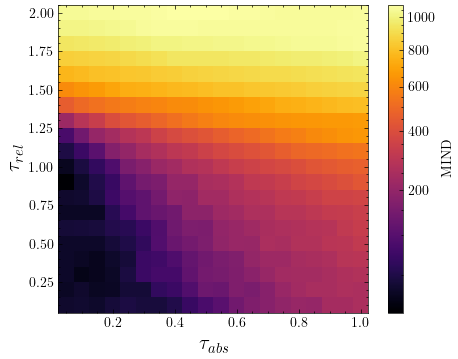}
        \caption{MIND as a function of $\tau_{abs}, \tau_{rel}$.}
        \label{fig:mind_grid_initial}
    \end{subfigure}
    \caption{The results of the initial grid-search over the absolute tolerance $\tau_{abs}$ and relative tolerance $\tau_{rel}$ of the PI solver, measuring NFE and the MIND error metric.}
    \label{fig:tolerance_grid_initial}
\end{figure}

An initial grid search over $\tau_{abs}$ and $\tau_{rel}$ was performed, keeping $h_0=0.15$, $\alpha=0.9$, $k_i=0.3$, and $k_p=0.1$. At each point in the grid, 64 samples were generated using the experiment setup described in \autoref{app:implementation_details}. For these 64 samples, the average NFE was recorded, and MIND \citep{berthet_mind_2026} was measured as an error metric, as it is more accurate than FID at low sample size. This was done on a $20\times20$ grid. Step rejection was also enabled, via the method described in \autoref{app:solver_details}. The initial grid was chosen with $\tau_{abs}\in[0.05, 1]$ and $\tau_{rel}\in[0.1, 2]$, with points linearly spaced. This grid was chosen as it spans a decently large space of the hyperparameters, and each of the four corners of the grid were stable and produced images of decent quality, with each corner of the grid converging with NFE less than 1000. The results of the grid search can be found in \autoref{fig:tolerance_grid_initial}. 

In \autoref{fig:tolerance_grid_initial}, it can be seen that when $\tau_{abs}+\tau_{rel}$ are sufficiently large together, image quality quickly degrades, although NFE reduces as well when this becomes larger. Additionally, as $\tau_{abs}$ increases, NFE goes down slightly, but this comes at the cost of an increase in MIND. Only at the low range of $\tau_{abs}$ does MIND become low. MIND does not go much lower than 50, but this is likely caused by the sample size being too small to yield MIND values that accurately represent the sample quality when the sample quality is sufficient. As NFE seemed to increase less while allowing larger $\tau_{rel}$ and therefore less NFE at lower $\tau_{abs}$, a second grid search was performed with $\tau_{abs}\in[0.005, 0.1]$ and $\tau_{rel}\in[0.1, 2]$, of which the results are presented in \autoref{fig:tolerance_grid_main}. In this figure, $\tau_{abs}$ seems to have very little impact on the NFE-MIND tradeoff, although lower seems slightly better. Therefore, we chose $\tau_{abs}=0.005$.

With $\tau_{abs}=0.005$, the main knob for changing NFE is the combination of $h_0$ and $\tau_{rel}$. $h_0$ is tuned in tandem with $\tau_{rel}$. The value used for $h_0$ was empirically determined after selecting the other hyperparameters, picking the largest value of $h_0$ that led to results of high quality. Choosing $h_0$ to high generally led to a sharp decline in quality. Tuning $\tau_{rel}$ was done using manual binary search, with the goal of setting the solver at a specific NFE, so it can be easily compared with the other solvers. An initial value of $\tau_{rel}$ was picked based on the tolerance grid plot \autoref{fig:tolerance_grid_main}. From there, the relative tolerance combined with $h_0$ was tuned until the desired NFE was reached by increasing $\tau_{rel}$ if NFE was too large, while decreasing if NFE was too low. An average NFE that rounds to the desired NFE when rounded to the nearest integer was taken to be sufficient.

Lastly, $t_{ode}$ and $H_{ode}$ need to be determined. These values are based on similar ones used by EDM \citep{karras_elucidating_2022}. In their stochastic sampler, they found that running purely on the probability flow ODE from $t=0.05$ was optimal for ImageNet-64. Therefore, $t_{ode}$ was set to $0.05$. ODE steps were taken using the EDM schedule \eqref{eq:edm-schedule}, as that is generally regarded as a good schedule for image data. The number of ODE steps taken was based on the percentage of steps the EDM schedule would take from $t=0.05$ until convergence. This was found to be approximately $20\%$ of total steps. Therefore, the NFE used by the Heun sampler using the EDM discretization was set to $20\%$ of the total NFE. As Heun steps take 2 NFE each, this means that 49 NFE gets 5 ODE steps, 75 NFE 7.5 (which is rounded down to 7), and 99 NFE 10. Please note that these NFE were considered in the total NFE measured for the PI solver.

\subsubsection{Gotta Go Fast Solver}
The Gotta Go Fast solver by \citet{jolicoeur-martineau_gotta_2021} did not work out of the box on ImageNet-64 using the EDM SDE \eqref{eq:edmsde} and checkpoint. Therefore, the hyperparameters of this solver needed to be tuned as well. This search started with using the default parameters used in the paper, $r=0.9$, $\alpha=0.9$ (this parameter is named $\theta$ in their paper, but corresponds to $\alpha$ of the PI solver), $\tau_{abs}=0.0078$ (named $\epsilon_{abs}$ in the paper, but corresponds to $\tau_{abs}$, $\tau_{rel}=0.5$ (named $\epsilon_{rel}$ in the paper, but corresponds to $\tau_{rel}$), and $h_0=0.01$. These parameters showed significant instability, causing large oscillations in step size and local error.

The oscillatory behavior seemed to largely be caused by too much sensitivity to the error for step size control. As step size is determined by $h = \min(t, \alpha h \epsilon^{-r})$, with $\epsilon$ the local error, $r$ was tweaked to a much smaller value ($0.1$). This instantly improved stability, and increasing $r$, even slightly, reintroduced some instability. However, the controller was still slightly unstable. By tweaking the other hyperparameters systematically to find which had the greatest impact on instability, it was found that the controller was quite sensitive to $\alpha$ as well. Lowering it slightly to $0.7$ improved stability. $\tau_{abs}$ was kept unaltered, as it seemed to be working mostly fine. The same methods as used to determine $\tau_{rel}$ and $h_0$ as used for the PI solver were applied to Gotta Go Fast solver. Note that the $\tau_{rel}$ required on this dataset to get low NFE was significantly larger than that of the original paper. 

Some tweaks were made to the solver to improve stability and quality of the solver samples. The solver was adapted to also take ODE steps at the end of the trajectory, as that puts the solver on more even ground with the PI solver.

\subsection{Bitstream Diffusion}
The method we used to determine the hyperparameters for the PI solver was similar to the one we used for the image experiments. We took the hyperparameters from the image experiments as a baseline and went from there. When tuning the hyperparameters, we noticed the solver sampled at very low NFE for the $\tau_{rel}$ values used in the image experiments. After more tuning, we found there to be a sort of tipping point. At low values of $\tau_{rel}$, the solver would quickly start needing an infinite number of steps to converge, and depending on the value of $\tau_{rel}$, this point occurred at a certain $t$. In the end, $\tau_{rel}$ could still be tuned accurately enough to have the NFE be rounded to the target NFE.

%% file: sections/appendices/implementation_details.tex
\section{Experiment Details}
\label{app:implementation_details}
The solvers were implemented in a new codebase, taking inspiration from the implementations by \citet{song_score-based_2021} and \citet{karras_elucidating_2022}. Addionally, the code by \citet{batzolis_cobit_2026} was integrated into the codebase, and extended to allow use of the PI solver. All experiments were performed on a cluster with 10 Quadro RTX 6000 GPUs using PyTorch.

\subsection{1D Multimodal Gaussian Diffusion}
\label{appsec:1d_data}
\subsubsection{Experiment Setup}
The 1D experiment was executed on a mixture of Gaussians with varying weights, means, and variance, given by:
\begin{equation}
    P(x) \sim \frac{1}{\sum_{j=1}\pi_j}\sum_{i=1}^N \pi_i \mathcal{N}(\mu_i, \sigma_i^2)
\end{equation}
With $\pi_i$ being the weight of the $i$'th Gaussian, $\mu_i$ being the mean of the $i$'th Gaussian, and $\sigma_i$ being the standard deviation of the $i$'th Gaussian. The specific parameters used for each Gaussian can be found in \autoref{tab:varying_gaussian}. Note that $N$ was set to $10$. The choices made were semi-random. The standard deviations were chosen such that both shallow and sharp peaks were found in the distribution. The means were chosen such that some parts of the distribution blend slightly, while some peaks have more space in between them. Finally, it was ensured that $N$ modes were found in the distribution, thus, no modes were fully blended.

An analytical expression for the score was used. A small value of $10^{-13}$ was added to the score, as otherwise, score values could get too small in areas outside of the distribution, causing them to become NaN values.

The experiment was run using the variance-preserving SDE, which \citet{song_score-based_2021} define as:
\begin{equation}
    \label{eq:vpsde}
    \mathrm{d}\mathbf{x} =  -\frac{1}{2}\beta(t)\mathbf{x}\mathrm{d}t + \sqrt{\beta(t)}\mathrm{d}\mathbf{w}
\end{equation}
Where we set $\beta(t)$ to the following linear function:
\begin{equation}
    \beta(t) = \beta_{min}+t(\beta_{max}-\beta_{min})
\end{equation}
With $\beta_{min}=0.1$ and $\beta_{max}=20$, as is the default setting for the VP SDE \citep{song_score-based_2021}. 

The Euler-Maruyama and Heun solvers used fully linear discretizations. The PI static Heun solver uses a discretization generated by a PI solver. The same PI discretization is used for each NFE, chosen as the discretization at the PI adaptive solver's lowest point, set by changing $\tau_{abs}$. $\tau_{abs}$ was set to $0.02656$. Synthetic data were created by taking 100000 samples from the multimodal Gaussian. Then, we sampled from the distribution obtained from \eqref{eq:vpsde} at $t=1$, yielding a sample distribution very close to $\mathcal{N}(0, 1)$. These were then passed into the solvers for each different number of steps (for non-adaptive solvers) or tolerance (for adaptive solvers). After obtaining the estimate from a solver, the Wasserstein distance was computed with respect to the samples before noise was added. This was recorded for each solver at each different number of steps or tolerance.

\begin{table}[]
\centering
\caption{Parametrization of the varying Gaussian.}
\label{tab:varying_gaussian}
\begin{tabular}{lllllllllll}\toprule
         & 1   & 2   & 3   & 4  & 5   & 6   & 7   & 8  & 9   & 10  \\\midrule
$\pi$    & 1   & 3   & 1   & 3  & 1   & 2   & 1   & 3  & 2   & 1   \\
$\mu$    & -12 & -10 & -8  & -5 & 0   & 4   & 9   & 13 & 15  & 17  \\
$\sigma$ & 0.5 & 1.5 & 0.2 & 2  & 0.3 & 1.5 & 0.1 & 1  & 0.2 & 0.1 \\\bottomrule
\end{tabular}
\end{table}

\subsubsection{Hyperparameters}
The Euler-Maruyama and Heun configurations were evaluated from 10 to 200 NFE with a resolution of 100 points. The PI solver was evaluated from $\tau_{abs}=0.001$ to $\tau_{abs}=1$. Note that the tolerance was exponentially spaced to give approximately equal resolution to values below $\tau_{abs}=1$ and above $\tau_{abs}=1$, as otherwise, the experiment would be quite skewed toward higher tolerance. The full hyperparameters used for the PI solver can be found in \autoref{tab:1d_hyperparams}.

\begin{table}[]
\centering
\caption{Hyperparameters used for the PI solver for 1D Gaussians using the VP SDE.}
\label{tab:1d_hyperparams}
\begin{tabular}{ll}\toprule
Parameter      & Value \\\midrule
$\tau_r$       & 0     \\
$\alpha$       & 0.9   \\
$\alpha_{min}$ & 0.2   \\
$\alpha_{max}$ & 5     \\
$h_0$    & 0.15  \\
$t_{ode}$      & 0     \\
$N$            & 0     \\
$k_i$          & 0.3   \\
$k_p$          & 0.1   \\\bottomrule
\end{tabular}
\end{table}

\subsection{Image Experiments}
\label{appsec:image_data}
\subsubsection{Image Generation}
The image generation process is heavily based on the implementation of EDM2 \citep{karras_analyzing_2024, karras2024guidingdiffusionmodelbad}, but adapted to work with various solvers. The checkpoint \href{https://nvlabs-fi-cdn.nvidia.com/edm2/posthoc-reconstructions/edm2-img64-xl-0671088-0.040.pkl}{edm2-img64-xl-0671088-0.040.pkl} was used for the ImageNet-64 experiments, while the \href{https://nvlabs-fi-cdn.nvidia.com/edm/pretrained/edm-ffhq-64x64-uncond-ve.pkl}{edm-ffhq-64x64-uncond-ve.pkl} was used for FFHQ. These checkpoints are denoising networks that predict the denoised image at $t=0$. This is not equivalent to a score network, which is required for \eqref{eq:rdp} to be applied. Therefore, the denoiser output $D(\mathbf{x}, t)$ is transformed to the score using \citep{karras_elucidating_2022}:
\begin{equation}
    \label{eq:denoise-to-score}
    \nabla_\mathbf{x} \log p_t(\mathbf{x}) = \frac{D(\mathbf{x}, t) - \mathbf{x}}{t^2}
\end{equation}
This can then be directly input into the reverse time SDE.

We used a batch size of 512. For each solver-NFE combination, the same random seed was used. Every new sample increments this random seed by one, changing the starting noise of that image and changing the noise added when solving the SDE. Conditional generation was done, and each sample used a label randomly generated with that sample's seed. This means that each solver generates images with the exact same starting noise, label, and adds the same random noise. 

For the Euler-Maruyama, stochastic Heun (EDM discretization and PI Static discretization), Gotta Go Fast, and PI solver, the drift and diffusion of \eqref{eq:edmsde} were used. The EDM solver does not directly sample from \eqref{eq:rdp}. Instead, it uses random churn added at each step. This was not expressed in the form of an SDE. The sampler implemented in the EDM2 repository \citep{karras_analyzing_2024, karras2024guidingdiffusionmodelbad}\footnote{Repository found \href{https://github.com/NVlabs/edm2/tree/main}{here}} was directly used, only adapted to allow different discretizations to be passed into the solver. The PI Static EDM condition also uses the EDM sampler, but with a discretization generated by the PI solver.

We retrieved the entropy schedule from the \href{https://github.com/DejanStancevic/Entropic-Time-Schedulers-for-Generative-Diffusion-Models}{github} repository of the paper by \citet{stancevic_entropic_2025}, matching the entropic schedule to the dataset. We used the interpolation method found in their repository to downsample the 128-step entropy schedule to the correct number of steps taken.

\subsubsection{Metric Calculation}
Each metric was calculated using 50000 generated images with all real images as a reference. We generated 3 samples of 50000 images at different seeds to allow for expressing uncertainty. No post-processing, except for the decoder that decodes the image to RGB space, was applied to the generated samples. The pipeline for calculating metrics was based on the implementation of EDM2 \citep{karras_analyzing_2024, karras2024guidingdiffusionmodelbad}. The pipeline was verified to work by comparing the FID results with those of the EDM2 pipeline, which yielded identical results.

NFE for adaptive solvers was computed by, at every time the reverse SDE was evaluated, incrementing total NFE by the batch size. This ensures that the second-order steps actually count for two NFE, and that elements that are already finished in an adaptive solver, which are then removed from the batch by masking, do not count for NFE. Finally, the total NFE was divided by the total number of samples to obtain the NFE per sample. For the non-adaptive solvers, NFE was set by setting the number of evaluation steps. For second-order solvers, this was determined using $n = \lfloor \frac{\text{NFE} + 1}{2}\rfloor$, which ensures that the number of steps set matches the number of score evaluations done. Specifically, the second-order solvers execute $n-1$ second-order steps, and then one first-order step. This does imply that the second-order solvers cannot be evaluated at an even number of NFE, hence, all NFE chosen are odd.

\subsubsection{Hyperparameters}
We used the EDM schedule \citep{karras_elucidating_2022} as a baseline schedule, which is given by:
\begin{equation}
    \label{eq:edm-schedule}
    t_i = \left(\sigma_{max}^{\frac{1}{\rho}} + \frac{i}{N-1} \left[\sigma_{min}^{\frac{1}{\rho}} - \sigma_{max}^{\frac{1}{\rho}}\right]\right)^\rho
\end{equation}
Where $\sigma_{max}$ indicates the highest noise level, and $\sigma_{min}$ the lowest noise level. We set $\rho=7$, $\sigma_{max}=80$, and $\sigma_{min}=0.002$, as are the hyperparameters used by \citep{karras_elucidating_2022}.

For both stochastic Heun and Euler-Maruyama, ODE steps were taken at the end of the trajectory from $t_{ode}=0.05$. These ODE steps were taken by switching the SDE to ODE when $t<t_{ode}$. For the EDM configuration, the same discretization as for EM was used (the one given by \autoref{eq:edm-schedule}). The sampler used is the one devised by \citet{karras_elucidating_2022}. To introduce stochasticity, the churn mechanism of the solver was used, with $S_{churn}=40$, $S_{tmin}=0.05$, $S_{tmax}=50$, and $S_{noise}=1.003$, which are the optimal settings found by \citet{karras_elucidating_2022} for ImageNet-64. These were also used for FFHQ. Whenever the PI discretization is used, we match the NFE from the run used to generate the discretization with the run using the discretization. The hyperparameters used for the PI solver and for the solver by \citet{jolicoeur-martineau_gotta_2021} can be found in \autoref{tab:image_hyperparams}. Note that for FFHQ and ImageNet-64, mostly the same hyperparameters were used. Only $\tau_{rel}$ and $h_0$ were slightly altered between the datasets. How these hyperparameters were obtained is explained in \autoref{app:hyperparameter-selection}.

\begin{table}[]
\centering
\caption{Hyperparameters used for image generation of the PI solver, and the adaptive solver devised by \citet{jolicoeur-martineau_gotta_2021} (Gotta Go Fast)}
\label{tab:image_hyperparams}
\begin{tabular}{lllllll}\toprule
\multicolumn{1}{c}{\multirow{2}{*}{Parameter}} & \multicolumn{3}{c}{PI solver} & \multicolumn{3}{c}{Gotta Go Fast} \\\cmidrule(lr){2-4}\cmidrule(l){5-7}
\multicolumn{1}{c}{}                             & 49 NFE   & 75 NFE  & 99 NFE   & 49 NFE    & 75 NFE    & 99 NFE   \\\midrule
$\tau_{abs}$                                     & 0.005    & 0.005   & 0.005    & 0.0078    & 0.0078    & 0.0078    \\
$\tau_{rel, \text{ImageNet-64}}$                 & 1.09     & 0.87    & 0.72     & 25.5      & 17        & 13.7      \\
$\tau_{rel, \text{FFHQ}}$                        & 1.15     & 0.92    & 0.72       & 25.4      & 17.15     & 13.7      \\
$\alpha$                                         & 0.9      & 0.9     & 0.9      & 0.7       & 0.7       & 0.7       \\
$\alpha_{min}$                                   & 0.2      & 0.2     & 0.2      & N/A       & N/A       & N/A       \\
$\alpha_{max}$                                   & 5        & 5       & 5        & N/A       & N/A       & N/A       \\
$h_{0, \text{ImageNet-64}}$                      & 45       & 40      & 35       & 25        & 25        & 20        \\
$h_{0, \text{FFHQ}}$                             & 40       & 35      & 35       & 25        & 25        & 20        \\
$t_{ode}$                                        & 0.05     & 0.05    & 0.05     & 0.05      & 0.05      & 0.05      \\
$N$                                              & 5        & 7       & 10       & 5         & 7         & 10        \\
$k_i$                                            & 0.3      & 0.3     & 0.3      & N/A       & N/A       & N/A       \\
$k_p$                                            & 0.1      & 0.1     & 0.1      & N/A       & N/A       & N/A       \\
$r$                                              & N/A      & N/A     & N/A      & 0.1       & 0.1       & 0.1       \\\bottomrule
\end{tabular}
\end{table}

\subsection{Bitstream Diffusion}
\label{appsec:bitstream-diffusion}
\subsubsection{Experiment Details}
The Bitstream diffusion experiments were executed using the exact framework introduced by \citet{batzolis_cobit_2026}, using their repository\footnote{Found at \href{https://github.com/GBATZOLIS/BitstreamDiffusion}{https://github.com/GBATZOLIS/BitstreamDiffusion}}. We did not change anything already existing in their repository, only extended it to be able to use our solver. We implemented an interface between their repository and ours to allow the code in their repository to use our PI solver. To allow the solver to produce high-quality samples, we also added support for self-conditioning to our solver. Note that no other alterations to the solver were necessary for functioning on this dataset. We sample from the reverse SDE that \citet{batzolis_cobit_2026} introduce, which is approximately equal to how EDM samplers add churn. This SDE is given by:
\begin{equation}
    \mathrm{d}\mathbf{x} = (1 + \lambda(t))t\nabla_x\log_{t}(\mathbf{x})\mathrm{d}t+\sqrt{2\lambda(t)t}\mathrm{d}\mathbf{\tilde{w}}
\end{equation}
With $\lambda(t): \mathbb{R} \rightarrow\mathbb{R}_{\geq0}$ a function determining the amount of noise added. When $\lambda(t)=0$, this is equivalent to the probability flow ODE (\ref{eq:ode}). Note that we substitute $t$ for $\sigma(t)$ for consistency, as \citet{batzolis_cobit_2026} define $\sigma(t)=t$. The additional drift and injected variance then approximately match EDMs churn mechanism when:
\begin{equation}
    \lambda(t_i)\approx\frac{\gamma_i t_i}{|h_i|}
\end{equation}
With $\gamma_i$ being the EDM-churn parameter indicating the amount of noise added in a step, defined by \citet{karras_elucidating_2022}:
\begin{equation}
    \label{eq:edm_churn}
    \gamma_i = 
    \begin{cases}
        \min(S_{churn} / N, \sqrt{2}-1) & t_i \in(S_{tmin}, S_{tmax}) \\
        0 & \text{otherwise}
    \end{cases}
\end{equation}
Where $S_{churn}$ is a hyperparameter defining the amount of noise added, and $S_{tmin}, S_{tmax}$ define the minimum and maximum $t$ between which noise is applied.

\subsubsection{Hyperparameters}
For parameterizing the EDM-churn parameter, we set $S_{tmin}=0.08$, $S_{tmax}\approx76.756$, as those are the values \citet{batzolis_cobit_2026} set for their samplers. We also set $S_{churn}$ to the value they use, although they choose it to give a constant $\gamma$ of $0.185$ when $t_i \in (S_{tmin}, S_{tmax})$, regardless of the number of steps $N$. We set $N$ to the number of steps taken, so equal to $\lfloor \frac{\text{NFE}+ 1}{2}\rfloor$ for second-order samplers. The hyperparameters we used for the PI sampler can be found in \autoref{tab:cobit_hyperparams}, with the method of how these were obtained described in \autoref{app:hyperparameter-selection}. Note that the 63 NFE configuration has lower $\tau_{rel}$ than the other higher NFE configurations. This is caused by a larger $h_0$ requiring a smaller $\tau_{rel}$ to counterbalance it, but does improve results for 63 NFE. For generating the PI static schedules, we ran the PI sampler for 1024 samples at the NFE the schedule is used at, using the hyperparameters matching that NFE. 

\begin{table}[]
\centering
\caption{Hyperparameters for text generation using the PI solver.}
\label{tab:cobit_hyperparams}
\begin{tabular}{llll}\toprule
\multicolumn{1}{c}{\multirow{2}{*}{Parameter}} & \multicolumn{3}{c}{PI Solver} \\\cmidrule{2-4}
\multicolumn{1}{c}{}                           & 63 NFE  & 127 NFE  & 255 NFE  \\\midrule
$\tau_a$                                       & 0.005   & 0.005    & 0.005    \\
$\tau_r$                                       & 0.0888  & 0.1017   & 0.0896   \\
$\alpha$                                       & 0.9     & 0.9      & 0.9      \\
$\alpha_{min}$                                 & 0.2     & 0.2      & 0.2      \\
$\alpha_{max}$                                 & 5       & 5        & 5        \\
$h_{start}$                                    & 35      & 20       & 20       \\
$k_i$                                          & 0.3     & 0.3      & 0.3      \\
$k_p$                                          & 0.1     & 0.1      & 0.1      \\\bottomrule
\end{tabular}
\end{table}

\subsection{Ablations}
\subsubsection{PI Solver Ablations}
For all PI solver ablations, the setup and hyperparameters described in \autoref{appsec:image_data} was used, specifically using the FFHQ checkpoint. We evaluated four configurations, the first of these four, configuration A, being the PI solver detailed in \autoref{chap:adaptive-solver}, with the hyperparameters found in \autoref{tab:image_hyperparams}. Configuration B is identical to configuration A, except that it sets $k_p=0$, disabling the proportional component of \eqref{eq:pi-controller}, therefore ignoring the contribution of the previous error. Configuration C is another alteration of configuration A. This time, to mixed absolute tolerance, as given by \eqref{eq:abs_tolerance} was used. This did require alteration of the relative tolerance as well. The used values can be found in \autoref{tab:ablation_hyperparams}. Lastly, configuration D uses the error estimator given by \eqref{eq:error_ilie}, which eliminates the relative tolerance and only leaves $\tau_{abs}$. This did have to be tuned to allow for the target NFE to be reached. The used $\tau_{abs}$ can also be found in \autoref{tab:ablation_hyperparams}.

\begin{table}[]
    \centering
    \caption{PI solver ablations hyperparameters.}
    \label{tab:ablation_hyperparams}
    \begin{tabular}{lllll}\toprule
    Configuration & Variable Name & 49 NFE & 75 NFE & 99 NFE  \\\midrule
    Config C      & $\tau_{rel}$  & 91     & 89     & 67      \\
    Config D      & $\tau_{abs}$  & 8.5    & 5.3    & 4.5     \\\bottomrule
    \end{tabular}
\end{table}

\subsubsection{Schedule Across Modalities}
For the foreign FFHQ schedule, we used the LM1B PI static schedule generated at 255 NFE, using the exact setup described in \autoref{appsec:bitstream-diffusion} to generate the schedule. For generating images, we used the setup described in \autoref{appsec:image_data}. For the foreign schedule on LM1B, we used the FFHQ PI static schedule generated at 99 NFE, using the setup described in \autoref{appsec:image_data} to generate the schedule, and using the setup described in \autoref{appsec:bitstream-diffusion} to generate the samples.

%% file: sections/appendices/samples.tex
\section{Samples}
\label{app:samples}

\subsection{ImageNet-64}
\begin{figure}[H]
    \begin{subfigure}[b]{0.46\textwidth}
        \centering
        \includegraphics[width=0.99\linewidth]{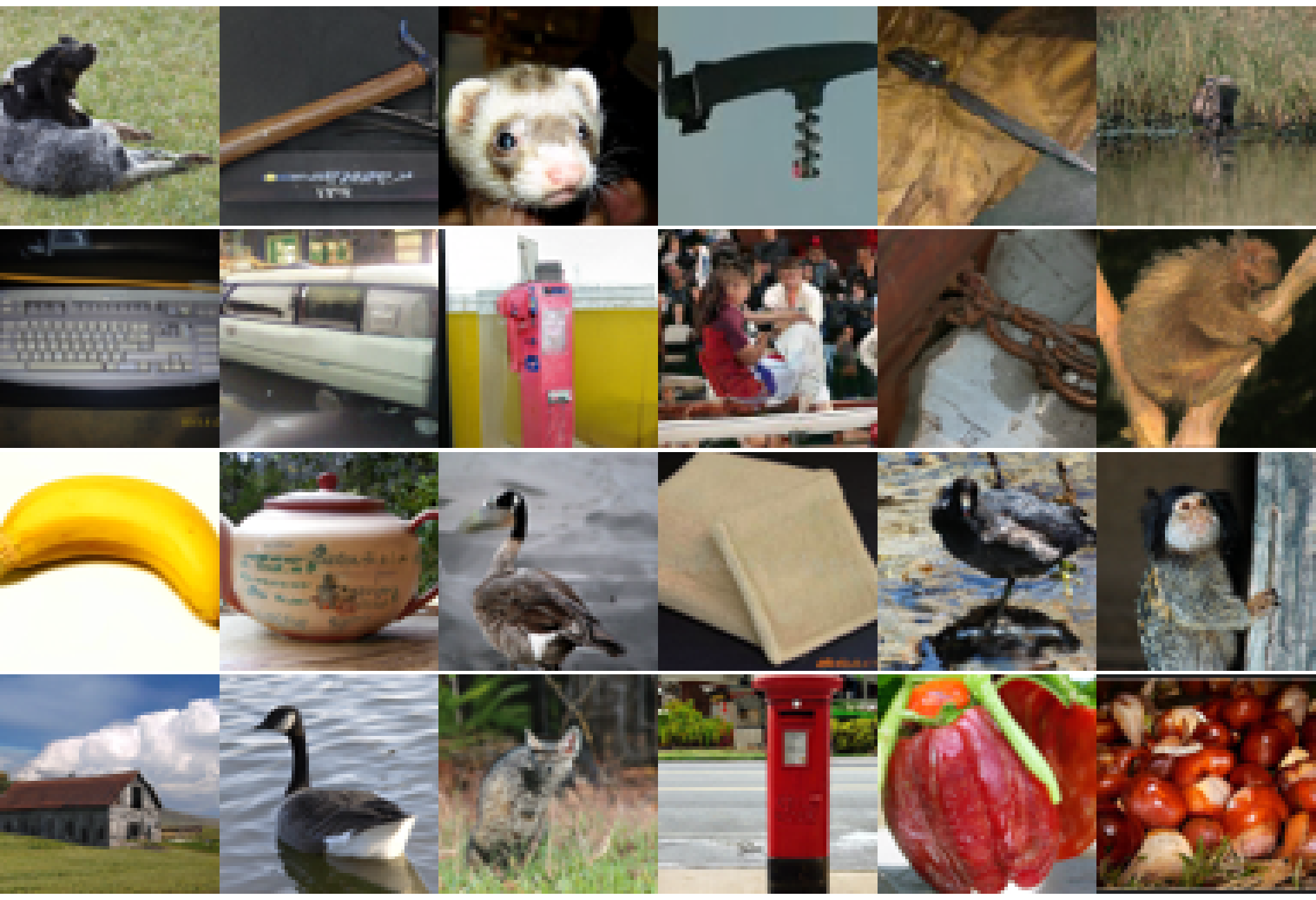}
        \caption{EDM-churn sampler with EDM schedule.}
    \end{subfigure}
    \hfill
    \begin{subfigure}[b]{0.46\textwidth}
        \centering
        \includegraphics[width=0.99\linewidth]{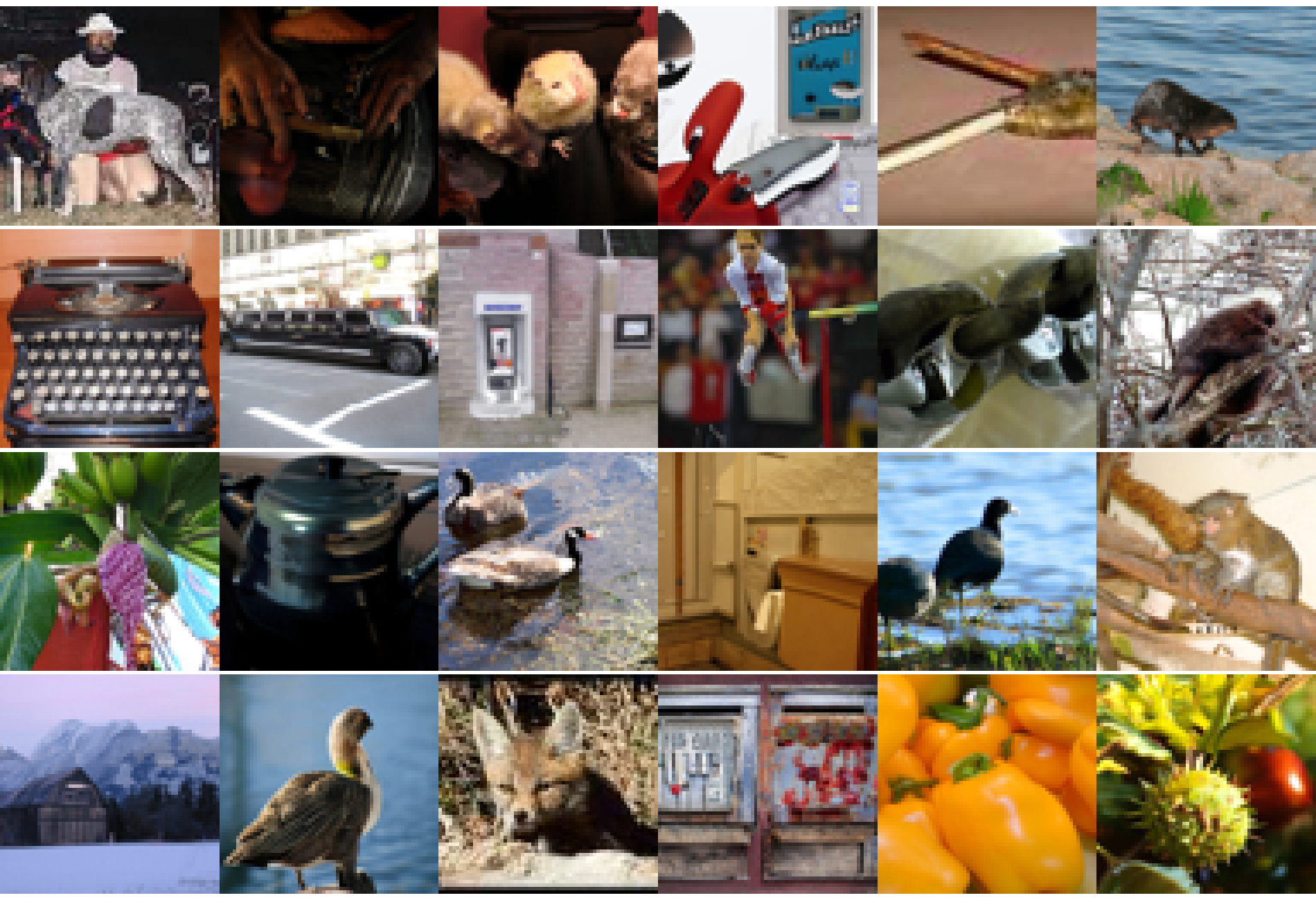}
        \caption{EDM-churn sampler with PI static schedule.}
    \end{subfigure}
    \begin{subfigure}[b]{0.46\textwidth}
        \centering
        \includegraphics[width=0.99\linewidth]{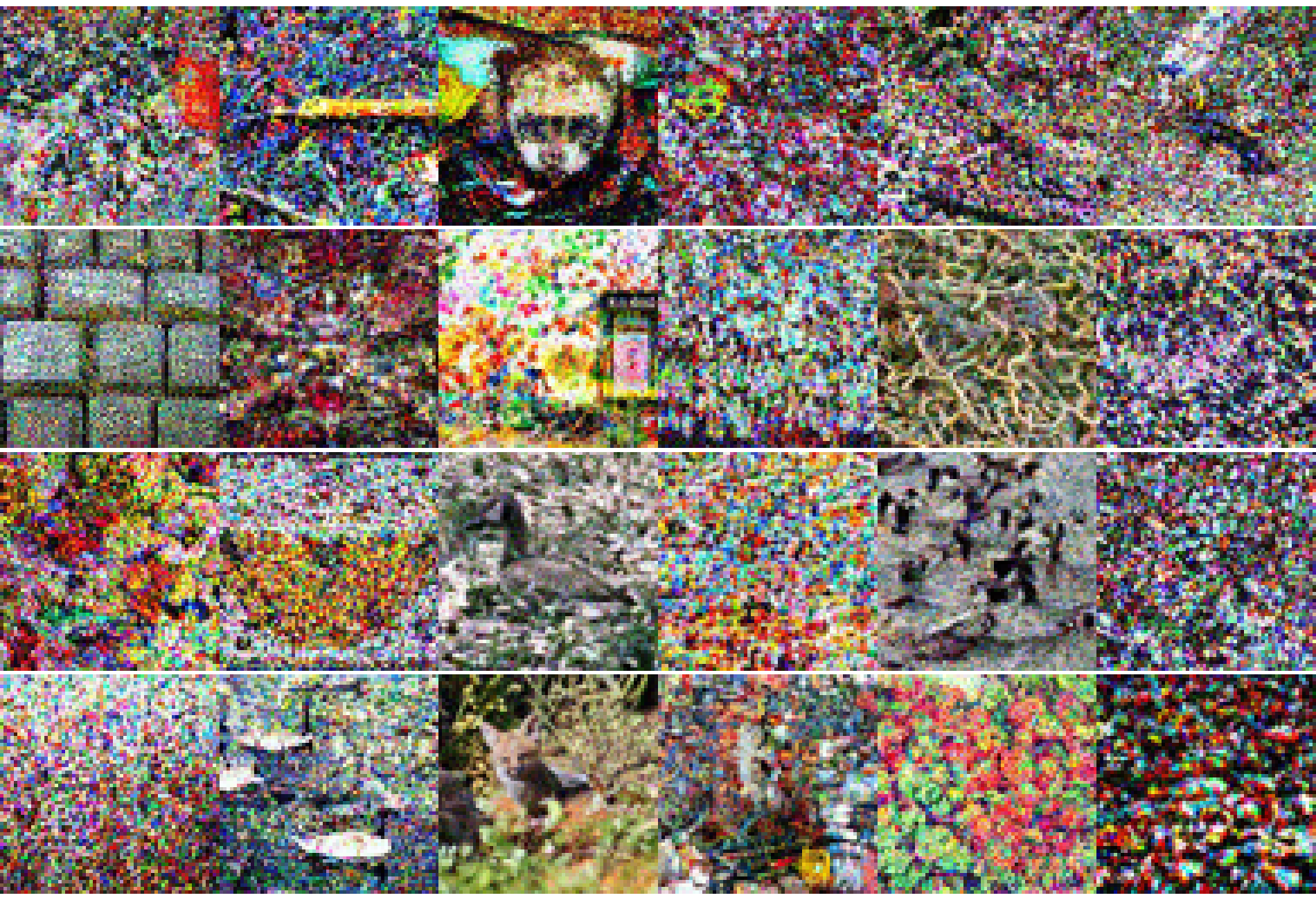}
        \caption{Gotta Go Fast adaptive sampler.}
    \end{subfigure}
    \hfill
    \begin{subfigure}[b]{0.46\textwidth}
        \centering
        \includegraphics[width=0.99\linewidth]{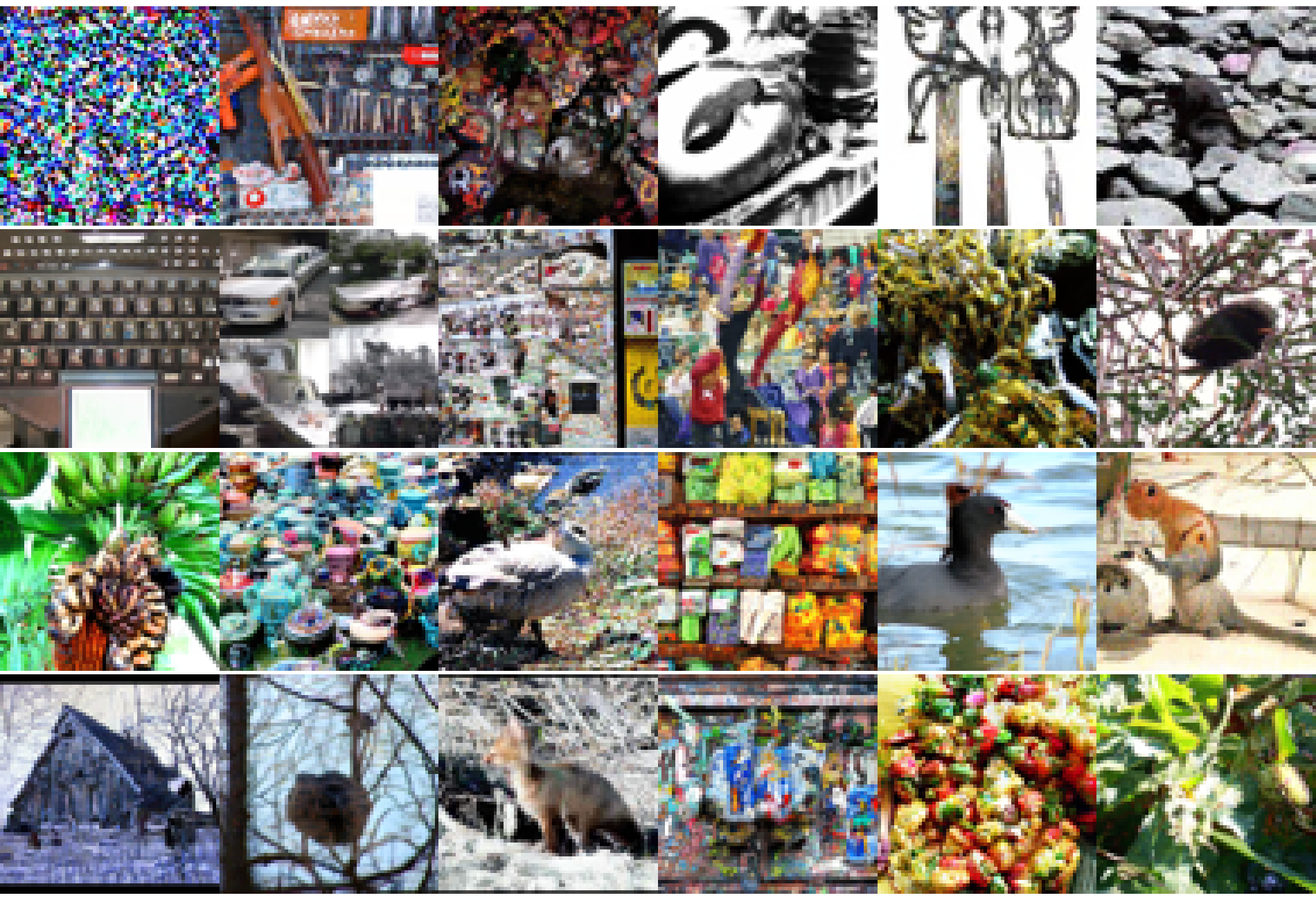}
        \caption{Proportional-Integral (ours) adaptive sampler.}
    \end{subfigure}
        \begin{subfigure}[b]{0.46\textwidth}
        \centering
        \includegraphics[width=0.99\linewidth]{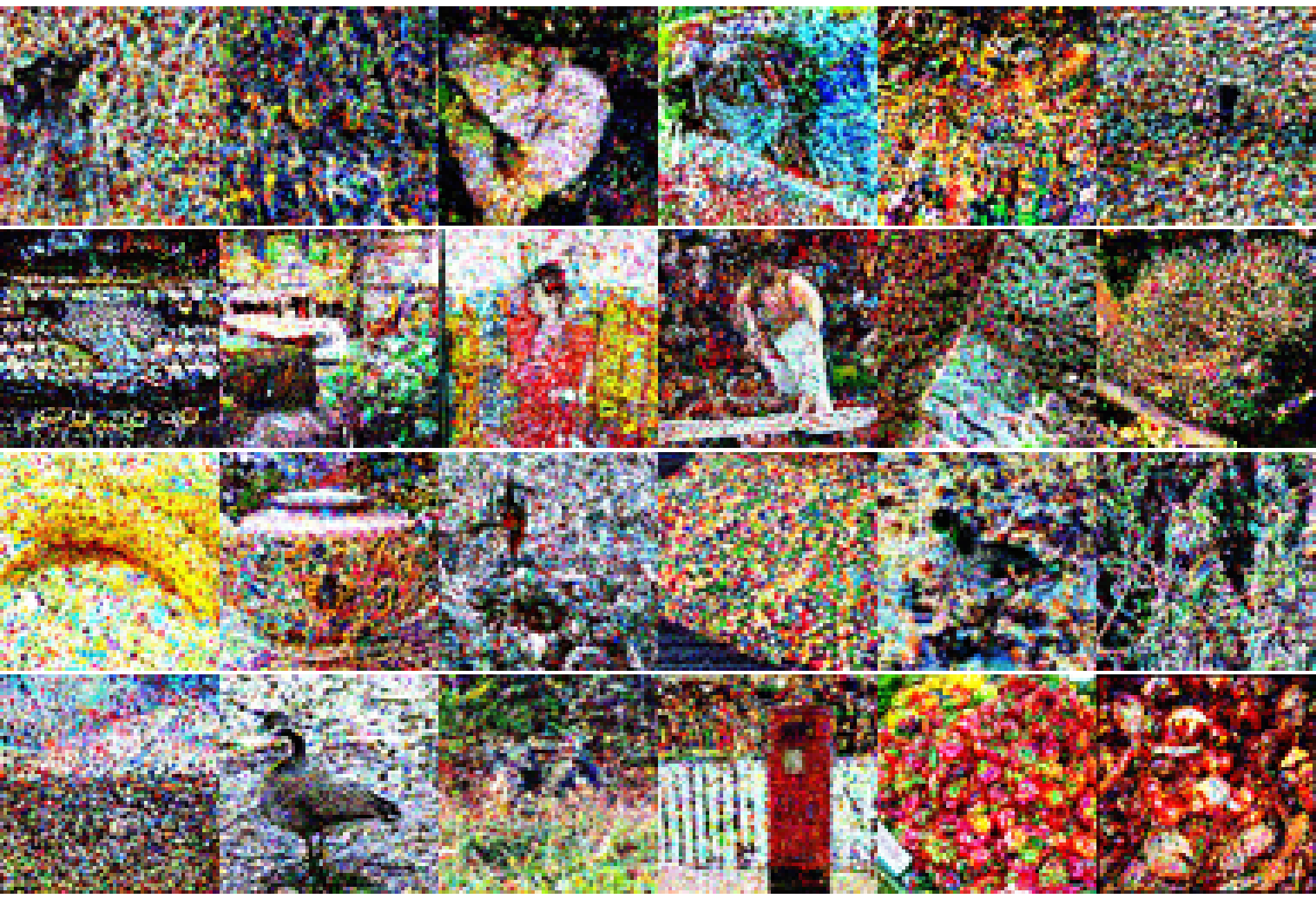}
        \caption{Stochastic Heun with EDM schedule.}
    \end{subfigure}
    \hfill
    \begin{subfigure}[b]{0.46\textwidth}
        \centering
        \includegraphics[width=0.99\linewidth]{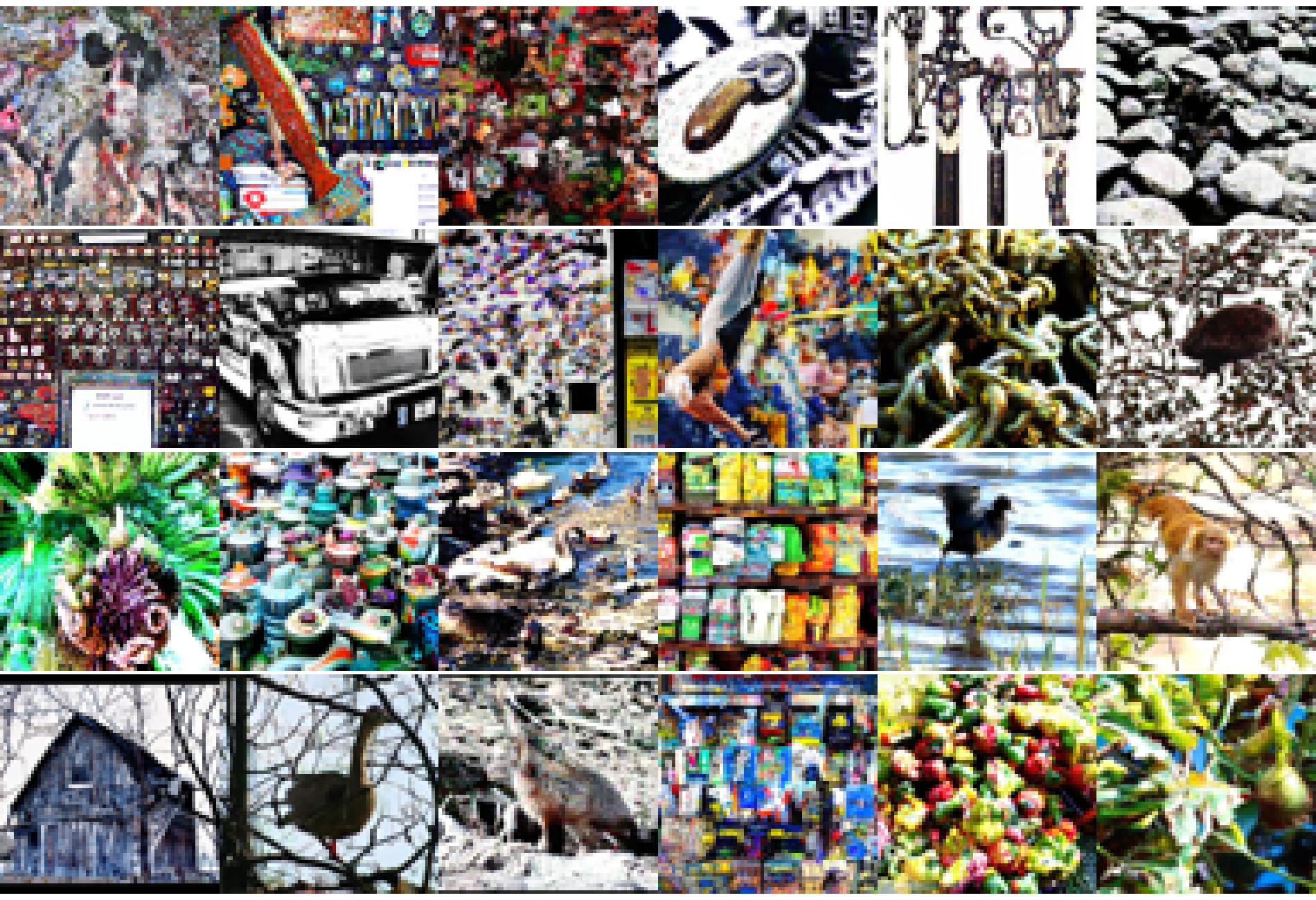}
        \caption{Stochastic Heun with PI static schedule.}
    \end{subfigure}
            \begin{subfigure}[b]{0.46\textwidth}
        \centering
        \includegraphics[width=0.99\linewidth]{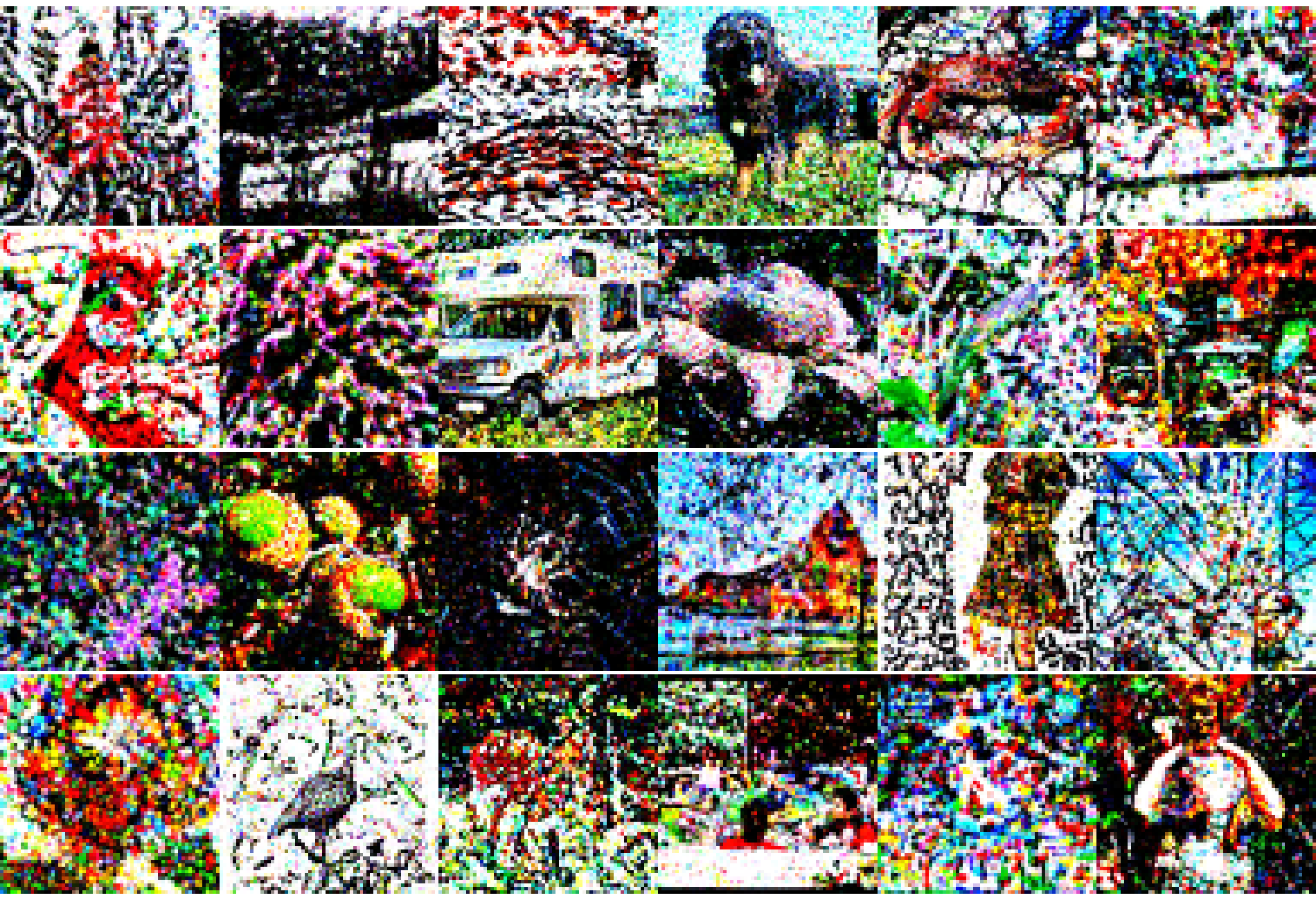}
        \caption{Euler-Maruyama with EDM schedule.}
    \end{subfigure}
    \hfill
    \begin{subfigure}[b]{0.46\textwidth}
        \centering
        \includegraphics[width=0.99\linewidth]{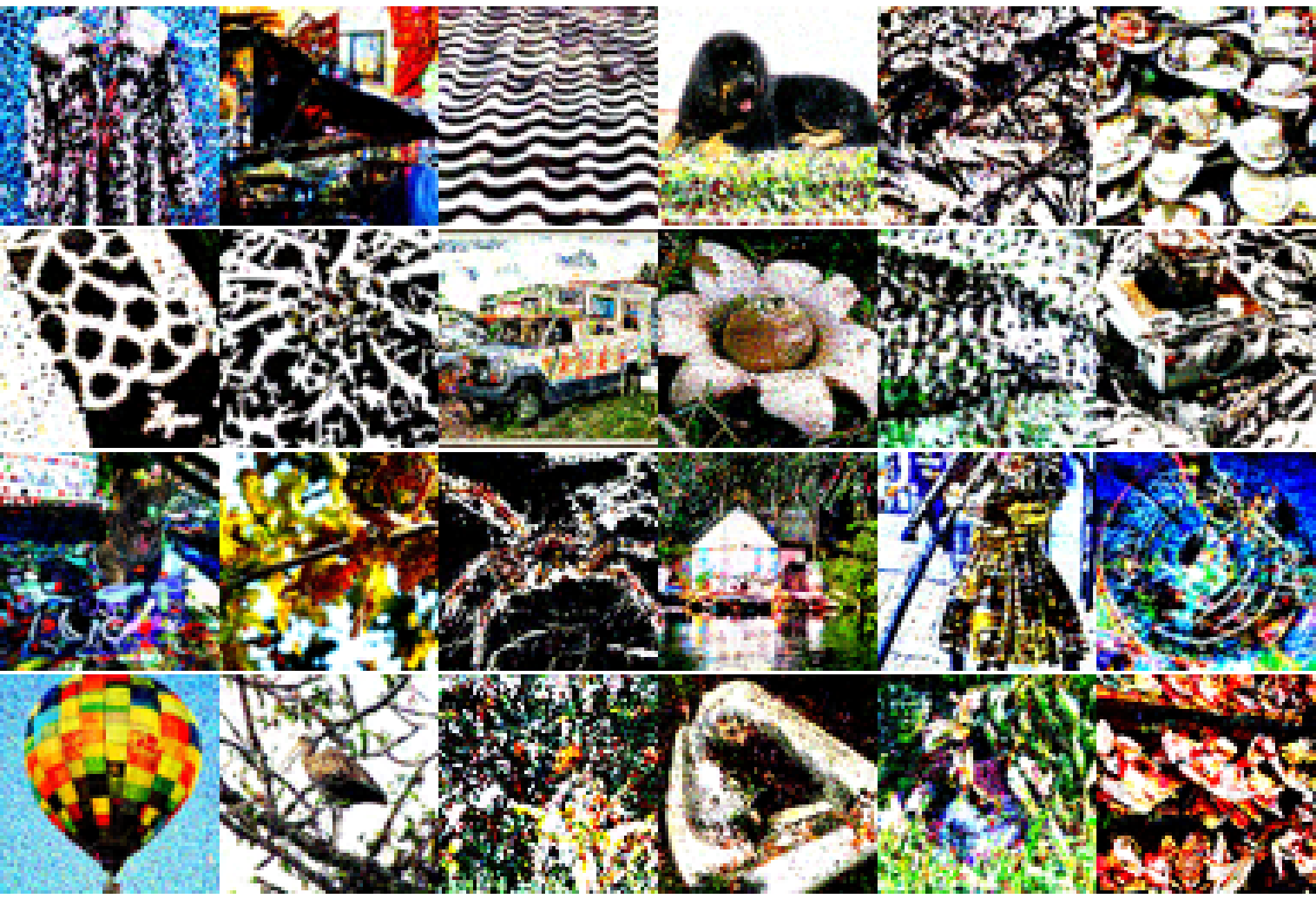}
        \caption{Euler-Maruyama with entropic schedule.}
    \end{subfigure}
    \caption{ImageNet-64 Images generated using different SDE solvers using the EDM2-XXL FID optimized checkpoint at 49 function evaluations}
    \label{fig:50NFE_samples}
\end{figure}

\begin{figure}[H]
    \begin{subfigure}[b]{0.46\textwidth}
        \centering
        \includegraphics[width=0.99\linewidth]{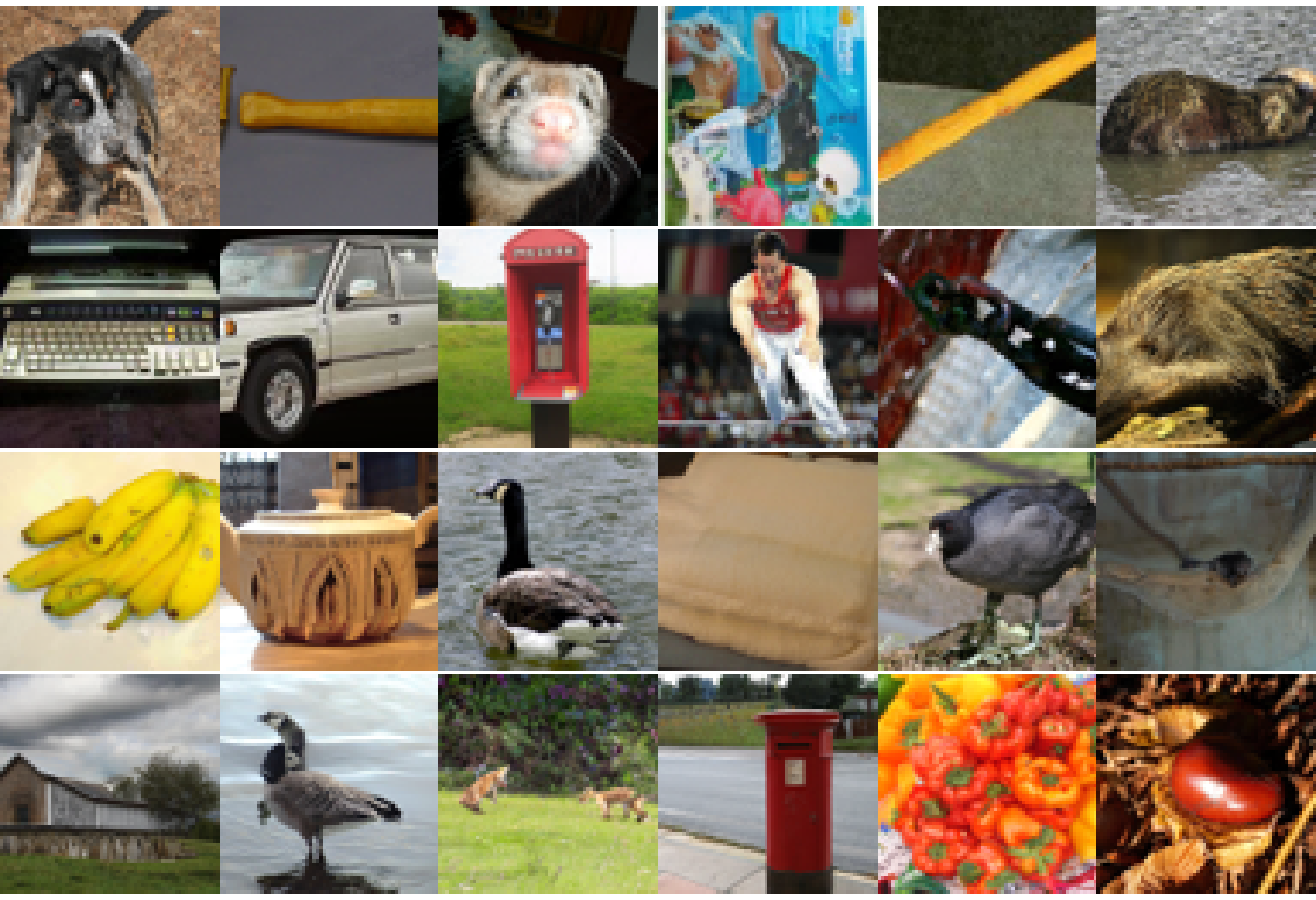}
        \caption{EDM-churn sampler with EDM schedule.}
    \end{subfigure}
    \hfill
    \begin{subfigure}[b]{0.46\textwidth}
        \centering
        \includegraphics[width=0.99\linewidth]{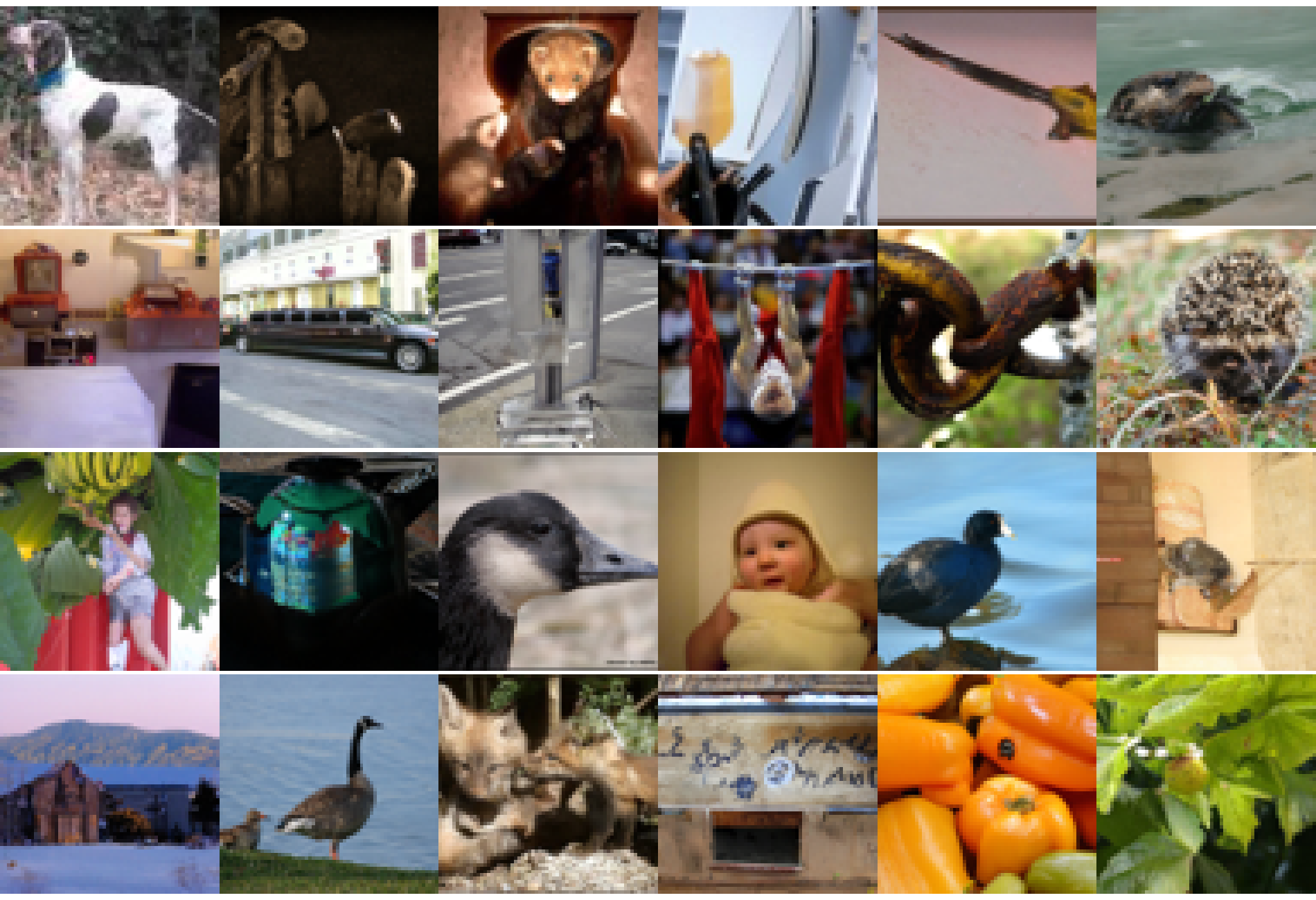}
        \caption{EDM-churn sampler with PI static schedule.}
    \end{subfigure}
    \begin{subfigure}[b]{0.46\textwidth}
        \centering
        \includegraphics[width=0.99\linewidth]{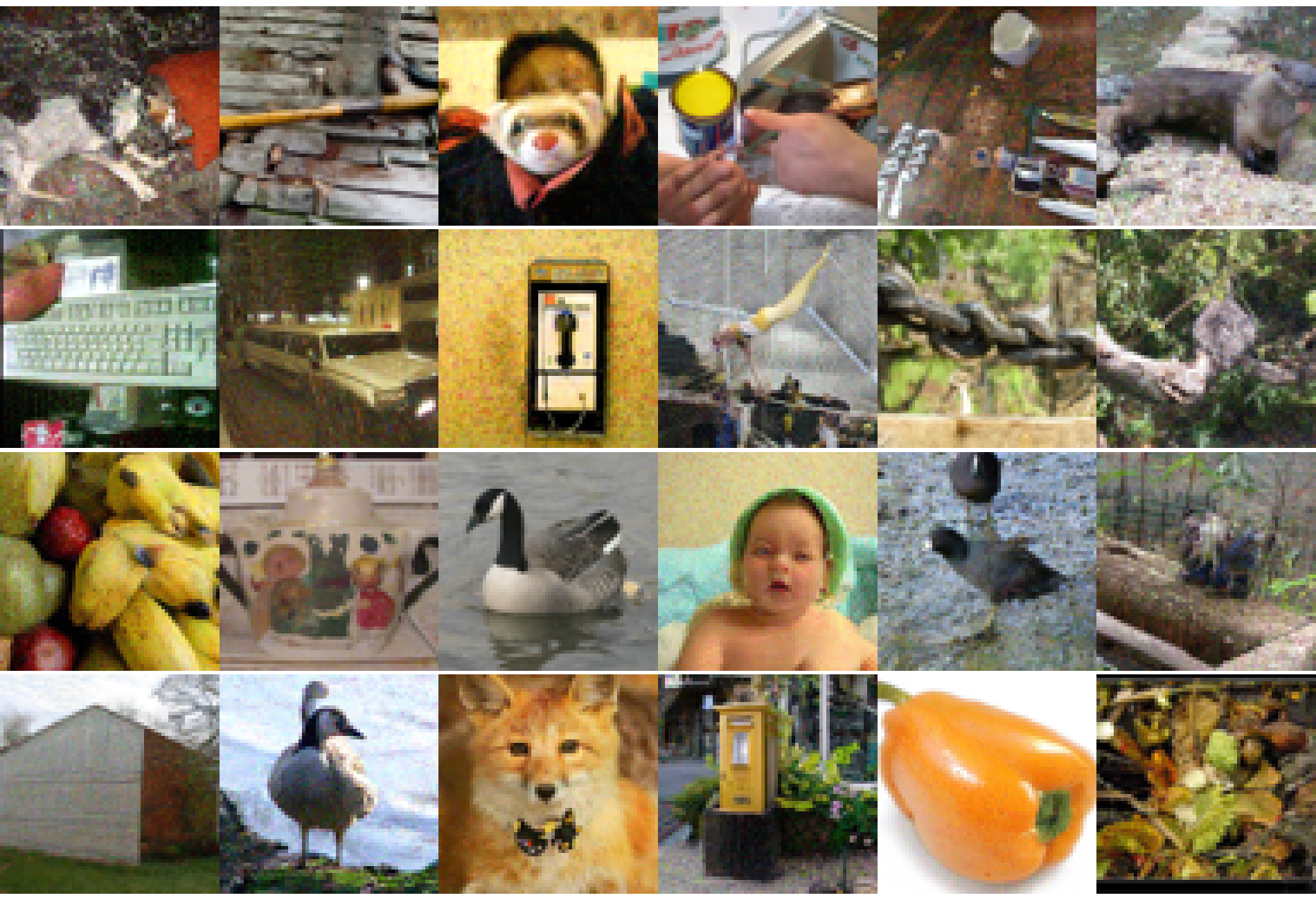}
        \caption{Gotta Go Fast adaptive sampler.}
    \end{subfigure}
    \hfill
    \begin{subfigure}[b]{0.46\textwidth}
        \centering
        \includegraphics[width=0.99\linewidth]{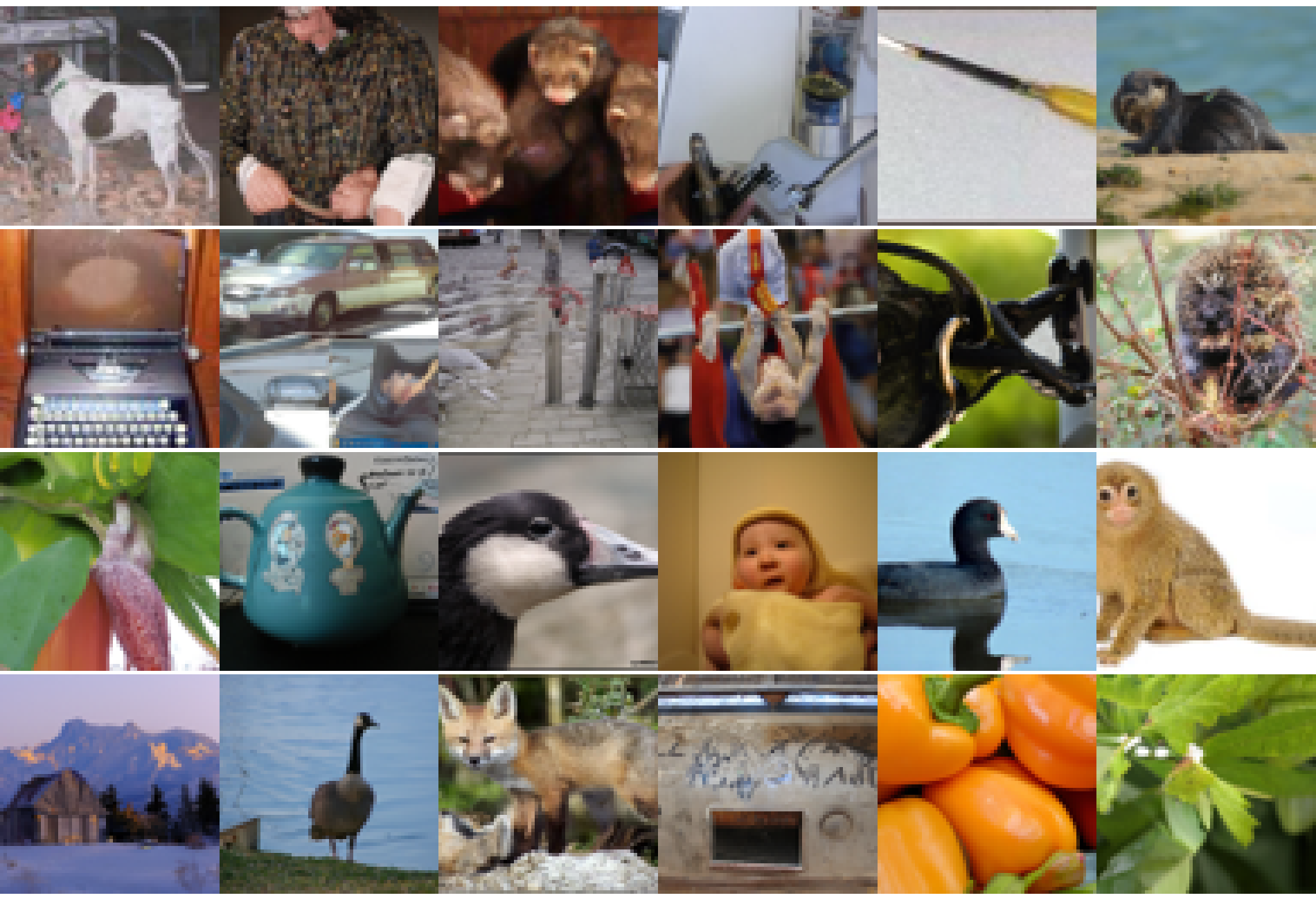}
        \caption{Proportional-Integral (ours) adaptive sampler.}
    \end{subfigure}
            \begin{subfigure}[b]{0.46\textwidth}
        \centering
        \includegraphics[width=0.99\linewidth]{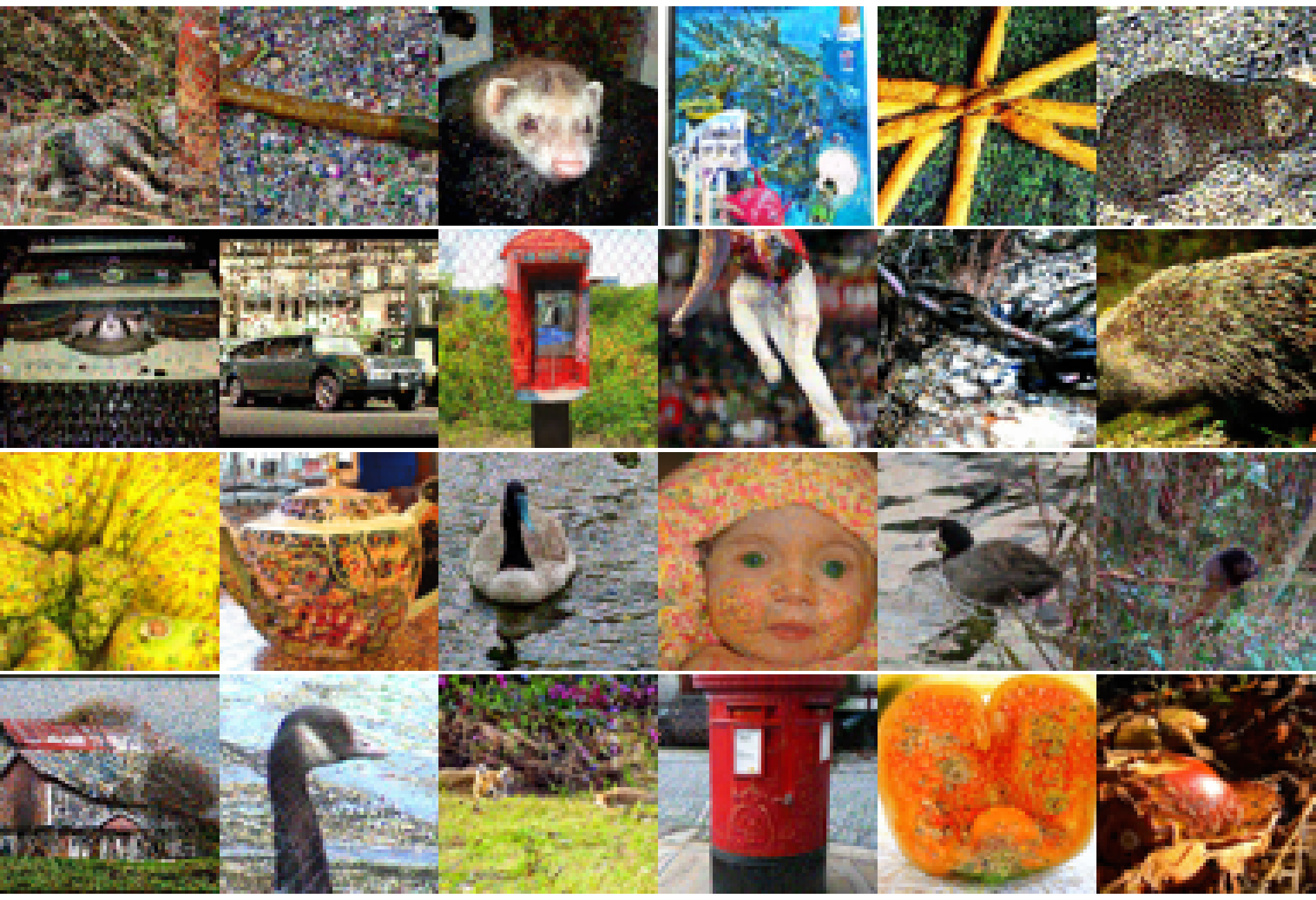}
        \caption{Stochastic Heun with EDM schedule.}
    \end{subfigure}
    \hfill
    \begin{subfigure}[b]{0.46\textwidth}
        \centering
        \includegraphics[width=0.99\linewidth]{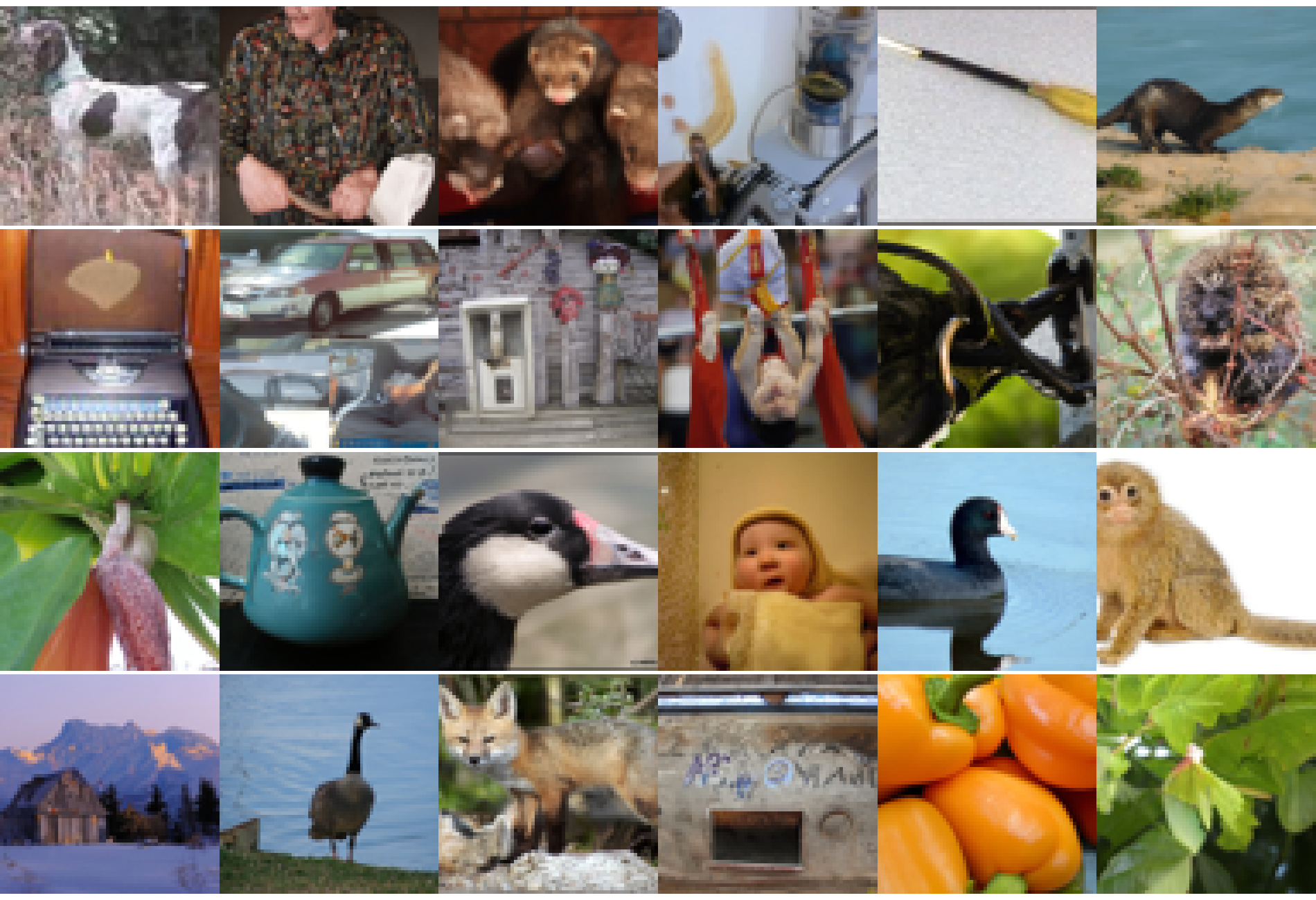}
        \caption{Stochastic Heun with PI static schedule.}
    \end{subfigure}
    \begin{subfigure}[b]{0.46\textwidth}
        \centering
        \includegraphics[width=0.99\linewidth]{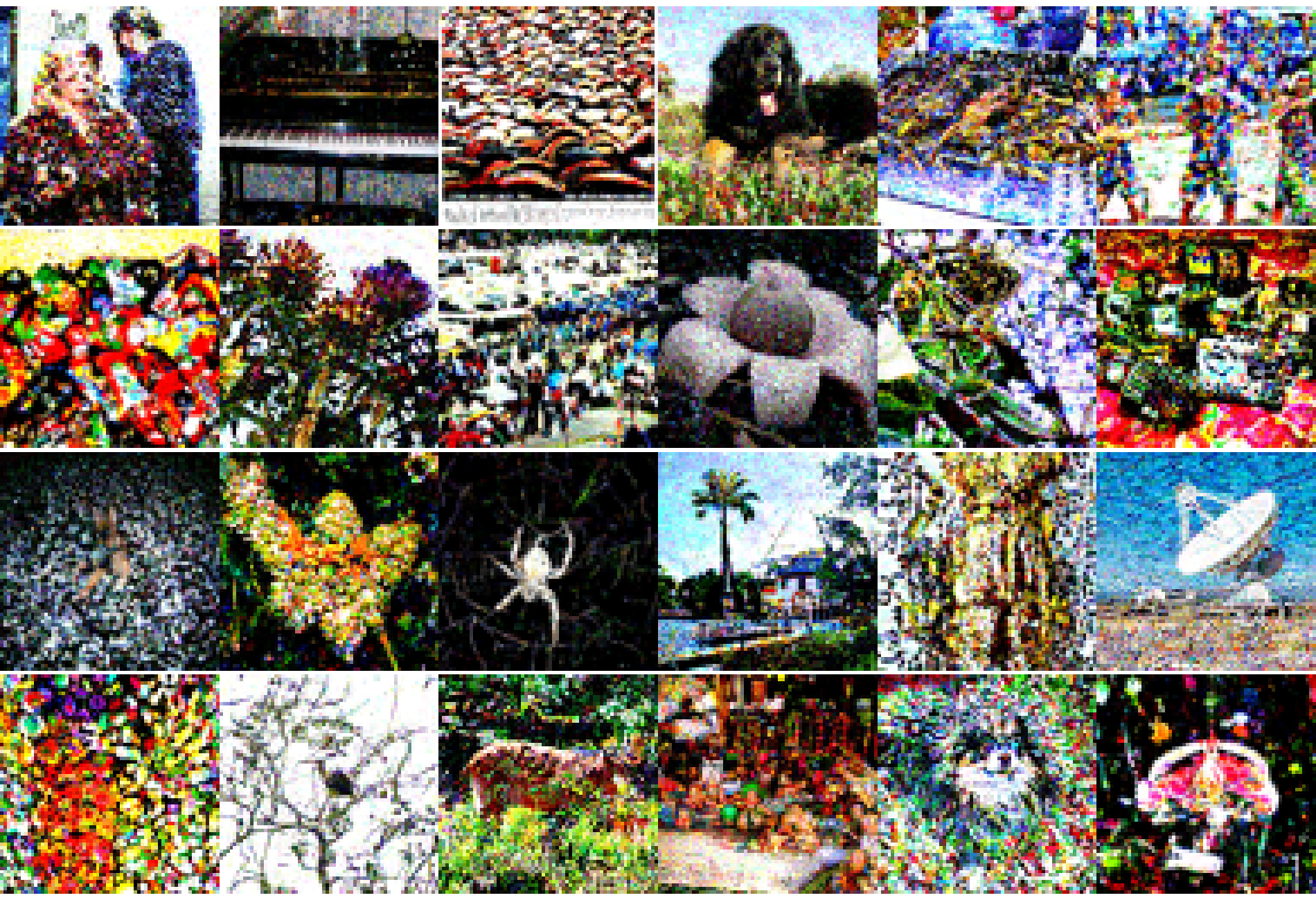}
        \caption{Euler-Maruyama with EDM schedule.}
    \end{subfigure}
    \hfill
    \begin{subfigure}[b]{0.46\textwidth}
        \centering
        \includegraphics[width=0.99\linewidth]{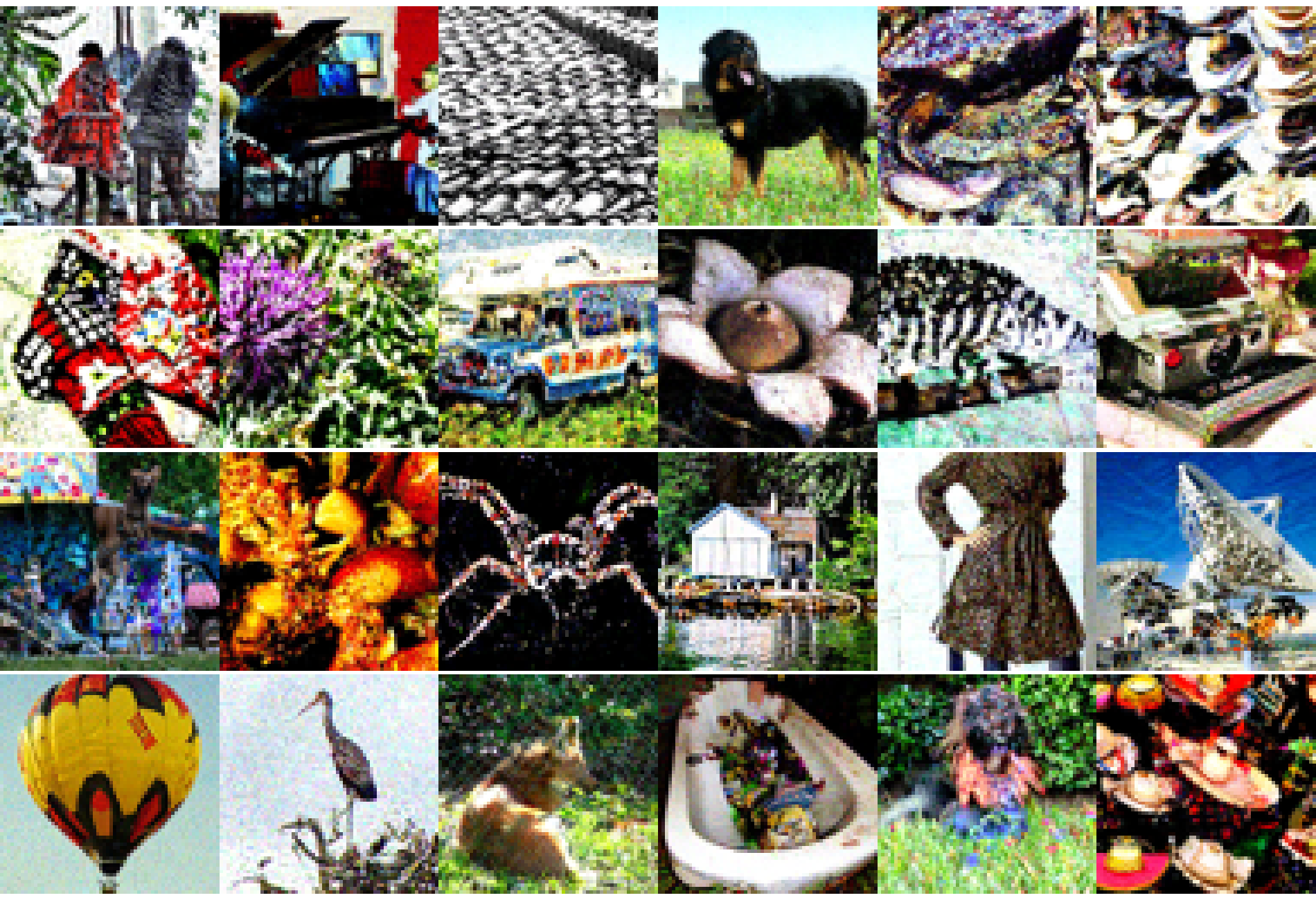}
        \caption{Euler-Maruyama with entropic schedule.}
    \end{subfigure}
    \caption{ImageNet-64 Images generated using different SDE solvers using the EDM2-XXL FID optimized checkpoint at 75 function evaluations}
    \label{fig:75NFE_samples}
\end{figure}

\begin{figure}[H]
    \begin{subfigure}[b]{0.46\textwidth}
        \centering
        \includegraphics[width=0.99\linewidth]{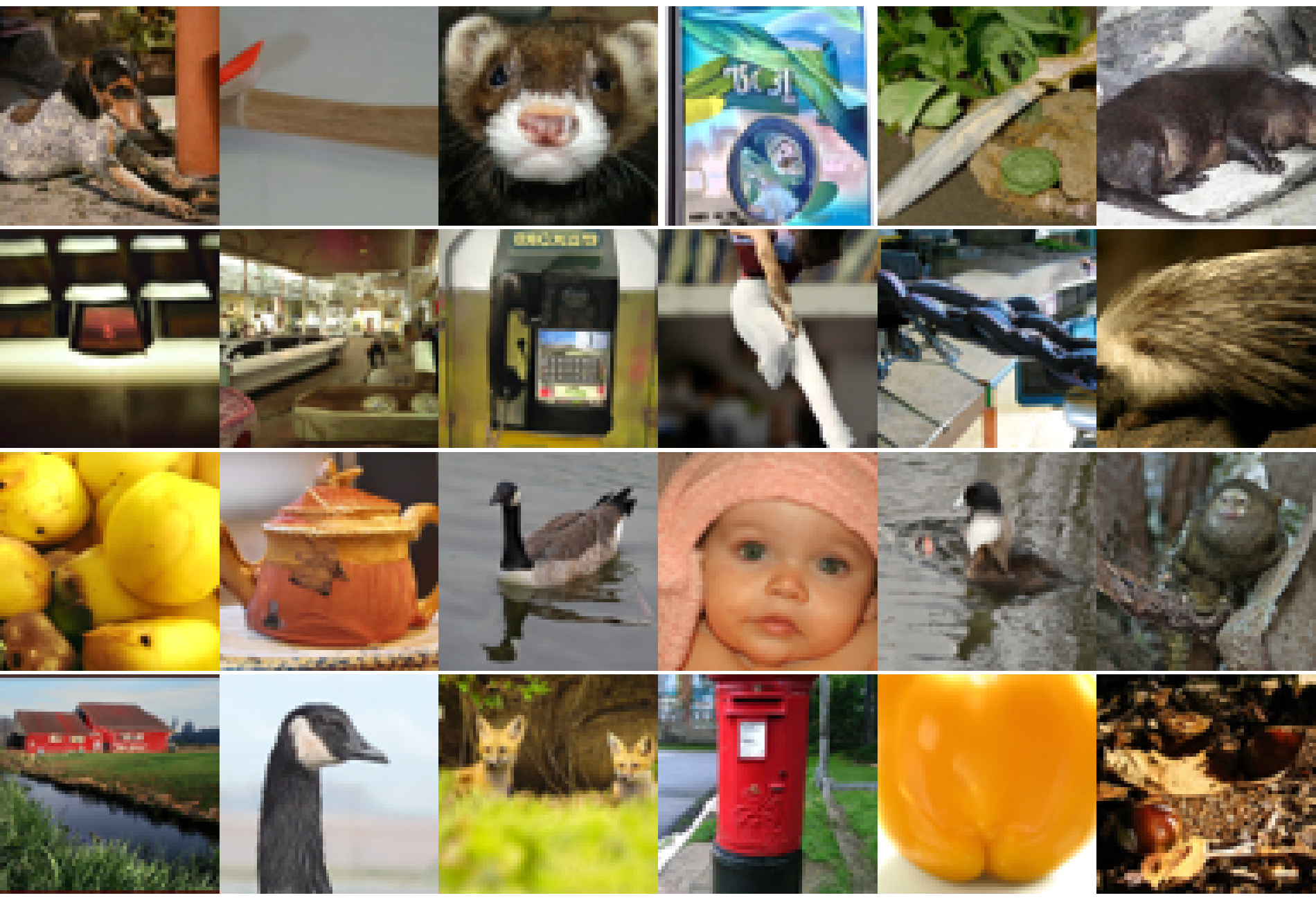}
        \caption{EDM-churn sampler with EDM schedule.}
    \end{subfigure}
    \hfill
    \begin{subfigure}[b]{0.46\textwidth}
        \centering
        \includegraphics[width=0.99\linewidth]{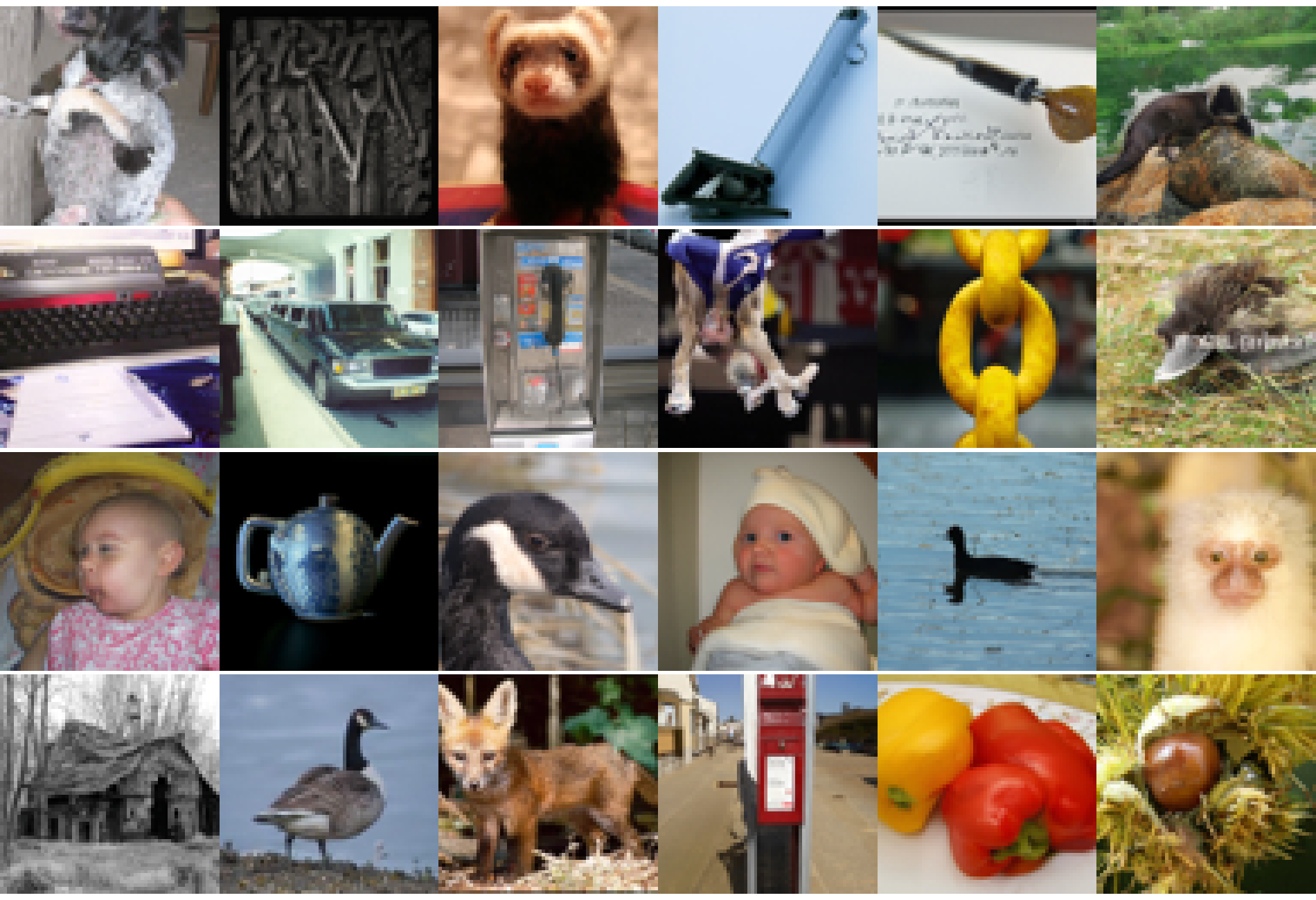}
        \caption{EDM-churn sampler with PI static schedule.}
    \end{subfigure}
    \begin{subfigure}[b]{0.46\textwidth}
        \centering
        \includegraphics[width=0.99\linewidth]{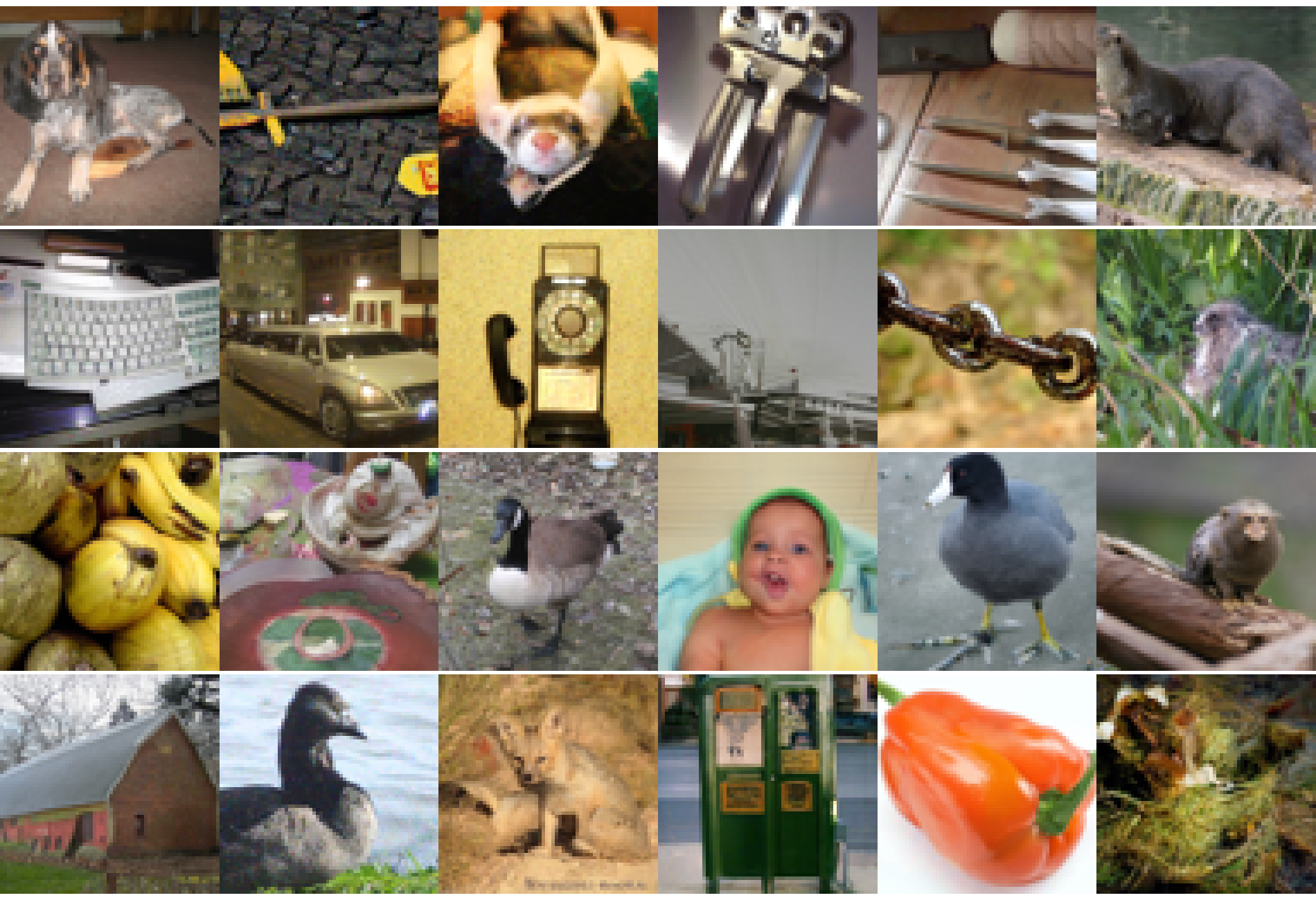}
        \caption{Gotta Go Fast adaptive sampler.}
    \end{subfigure}
    \hfill
    \begin{subfigure}[b]{0.46\textwidth}
        \centering
        \includegraphics[width=0.99\linewidth]{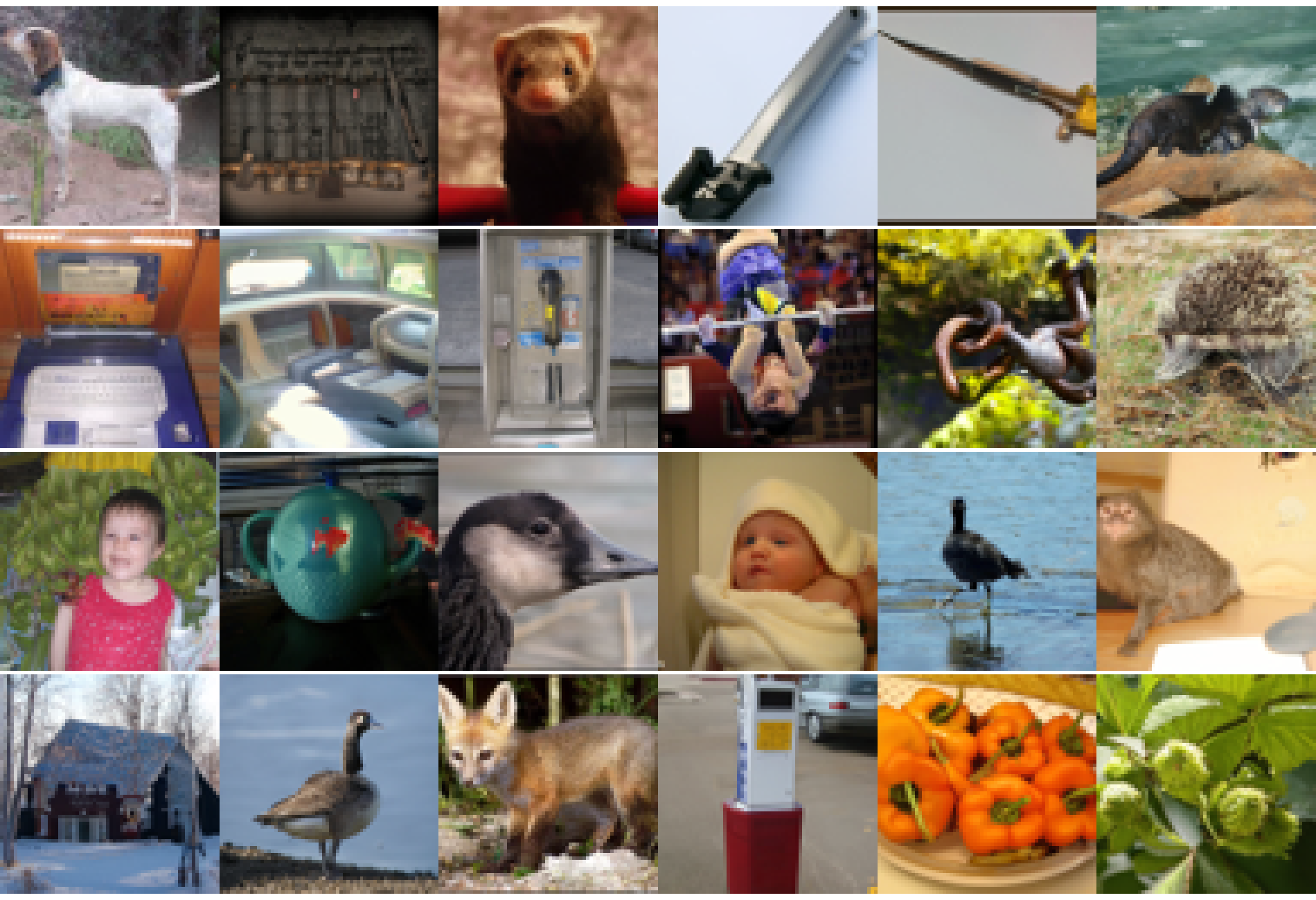}
        \caption{Proportional-Integral (ours) adaptive sampler.}
    \end{subfigure}
            \begin{subfigure}[b]{0.46\textwidth}
        \centering
        \includegraphics[width=0.99\linewidth]{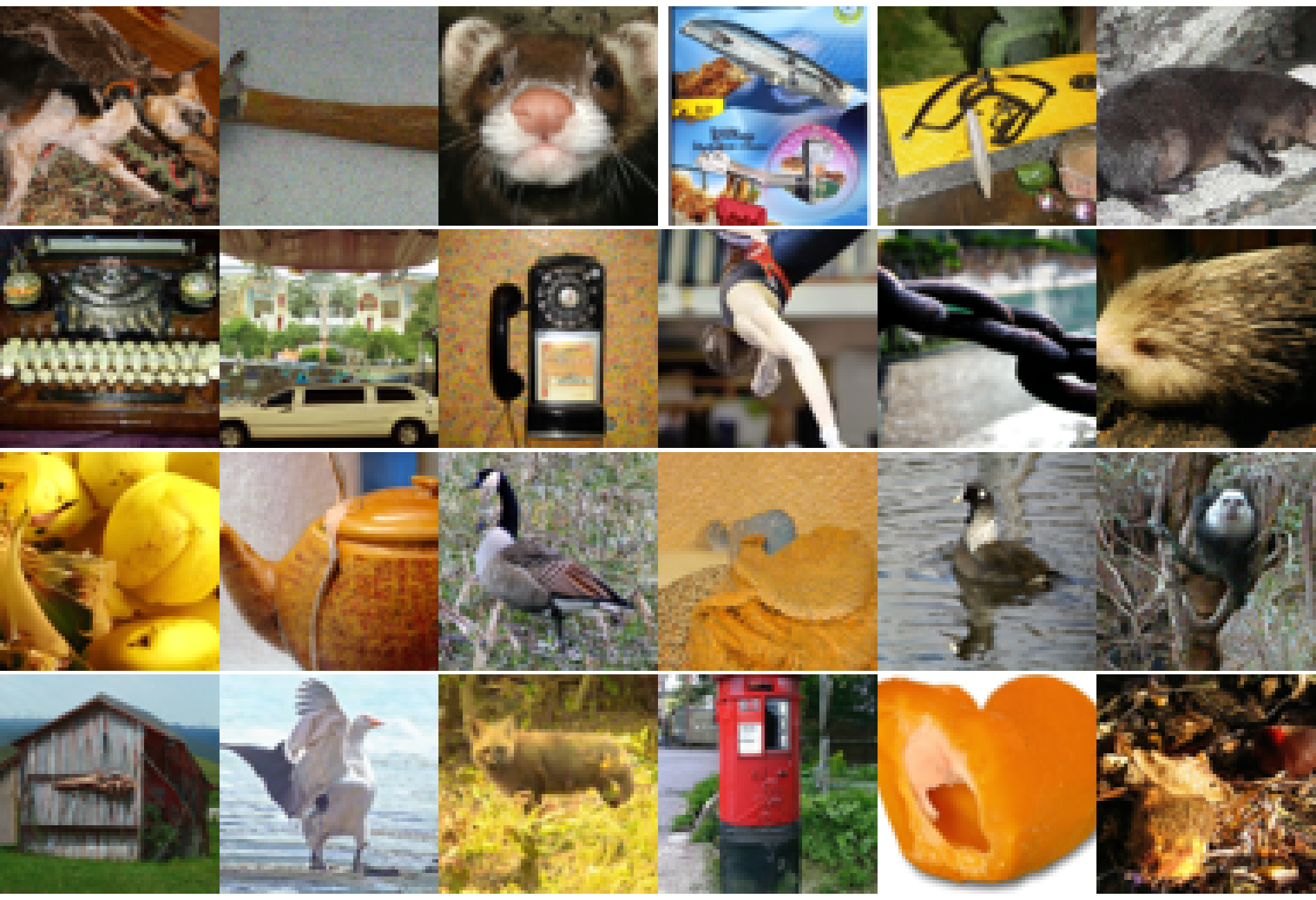}
        \caption{Stochastic Heun with EDM schedule.}
    \end{subfigure}
    \hfill
    \begin{subfigure}[b]{0.46\textwidth}
        \centering
        \includegraphics[width=0.99\linewidth]{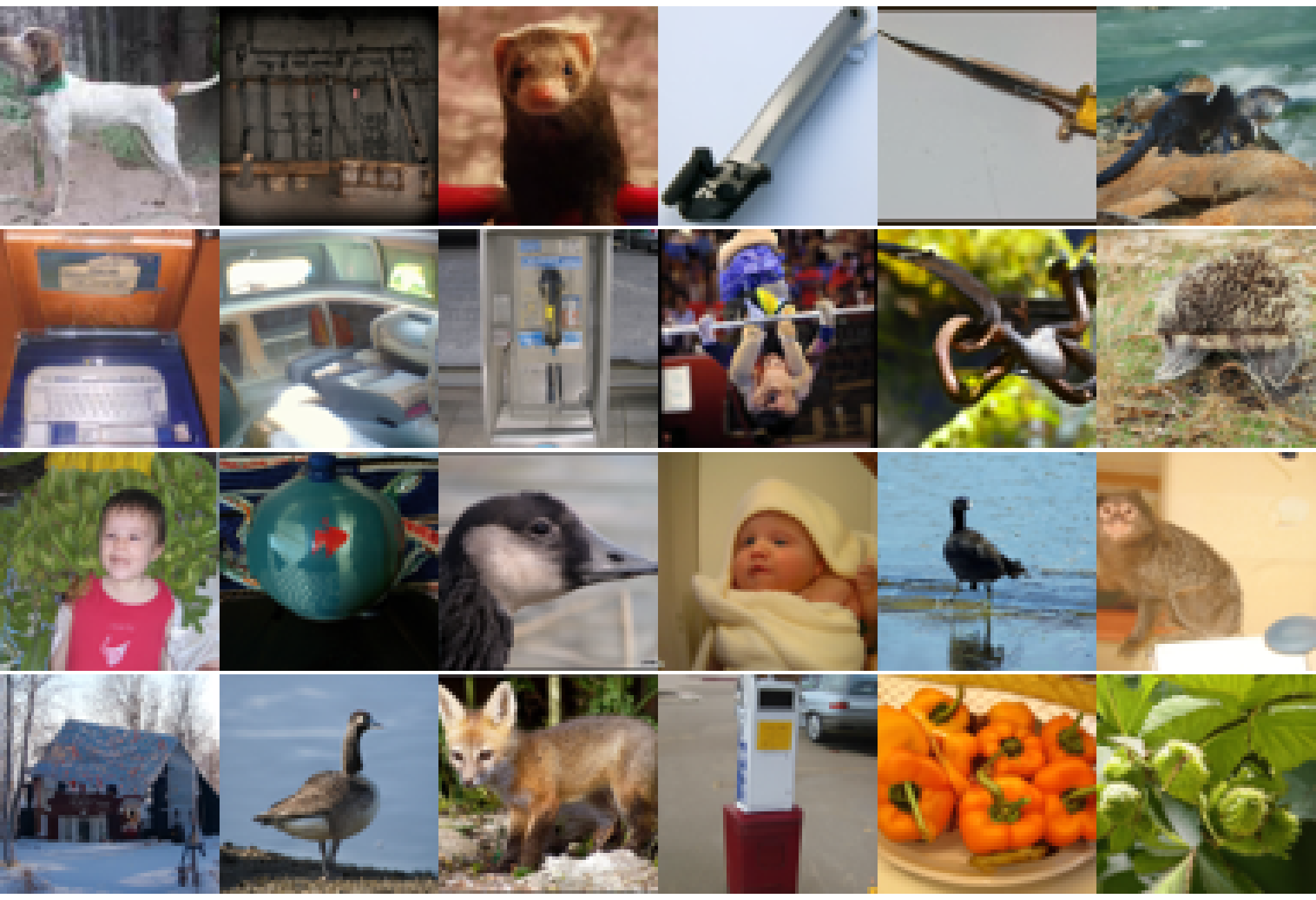}
        \caption{Stochastic Heun with PI static schedule.}
    \end{subfigure}
        \begin{subfigure}[b]{0.46\textwidth}
        \centering
        \includegraphics[width=0.99\linewidth]{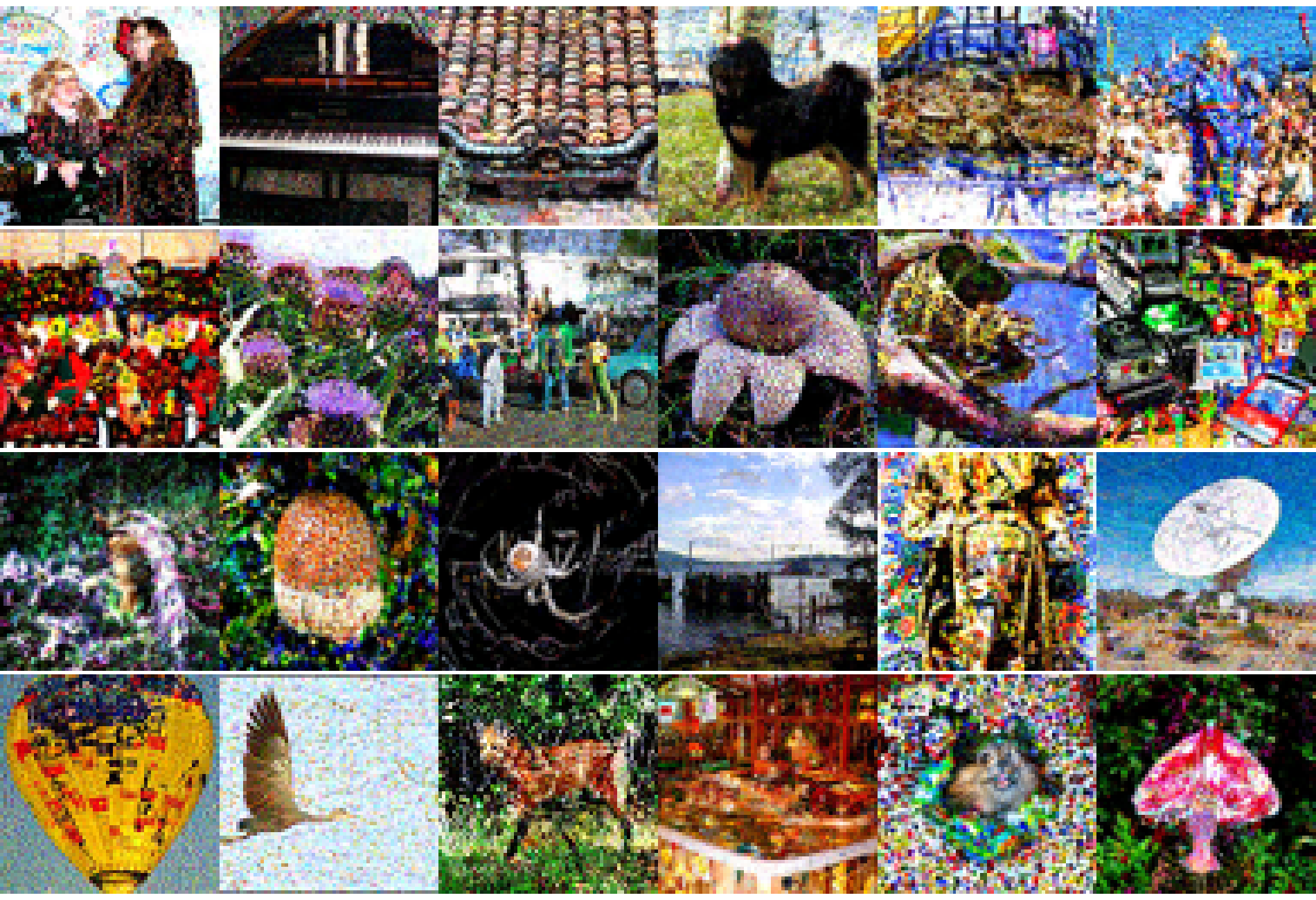}
        \caption{Euler-Maruyama with EDM schedule.}
    \end{subfigure}
    \hfill
    \begin{subfigure}[b]{0.46\textwidth}
        \centering
        \includegraphics[width=0.99\linewidth]{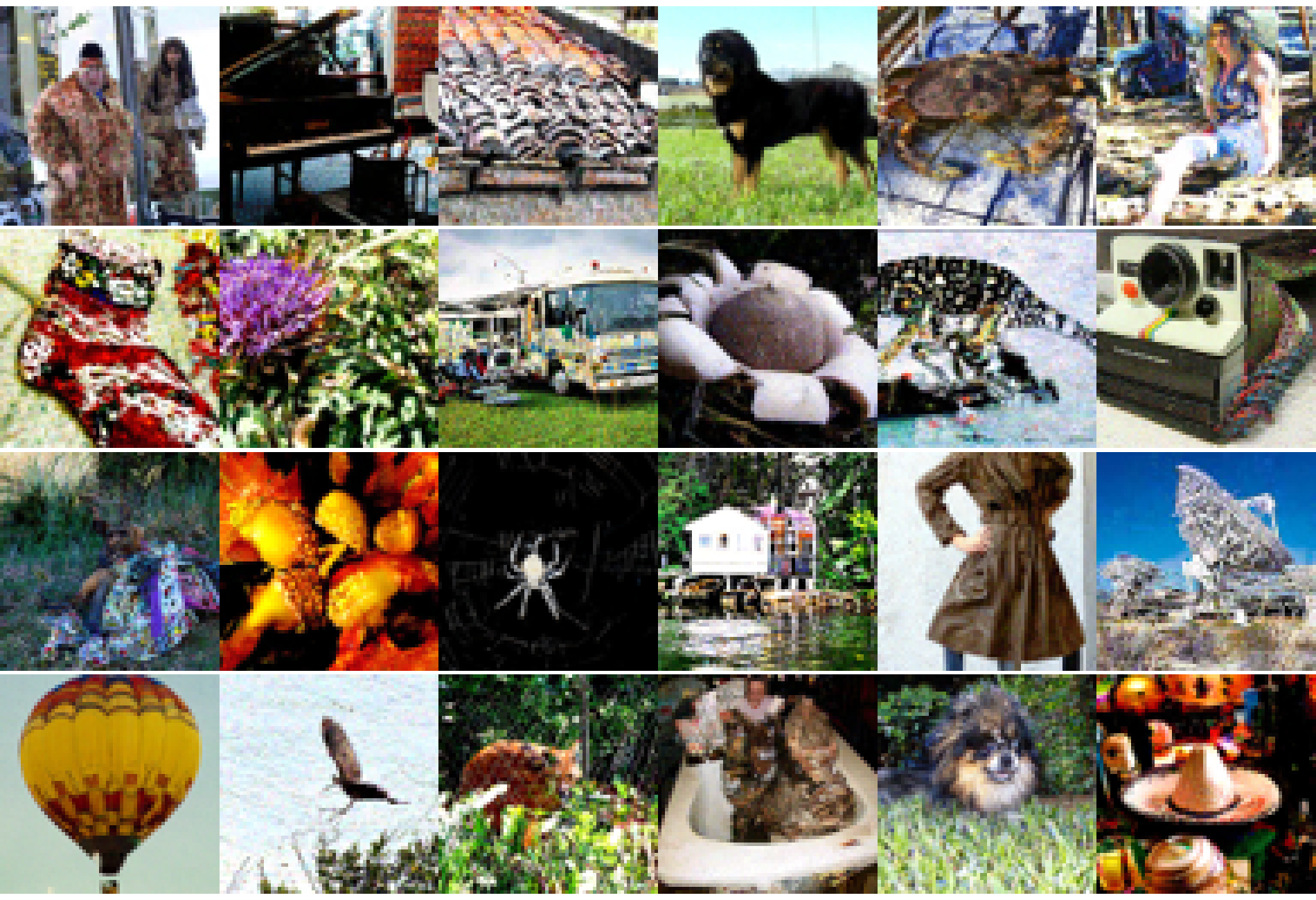}
        \caption{Euler-Maruyama with entropic schedule.}
    \end{subfigure}
    \caption{ImageNet-64 Images generated using different SDE solvers using the EDM2-XXL FID optimized checkpoint at 99 function evaluations}
    \label{fig:99NFE_samples}
\end{figure}

\newpage
\subsection{FFHQ}
\begin{figure}[H]
    \begin{subfigure}[b]{0.46\textwidth}
        \centering
        \includegraphics[width=0.99\linewidth]{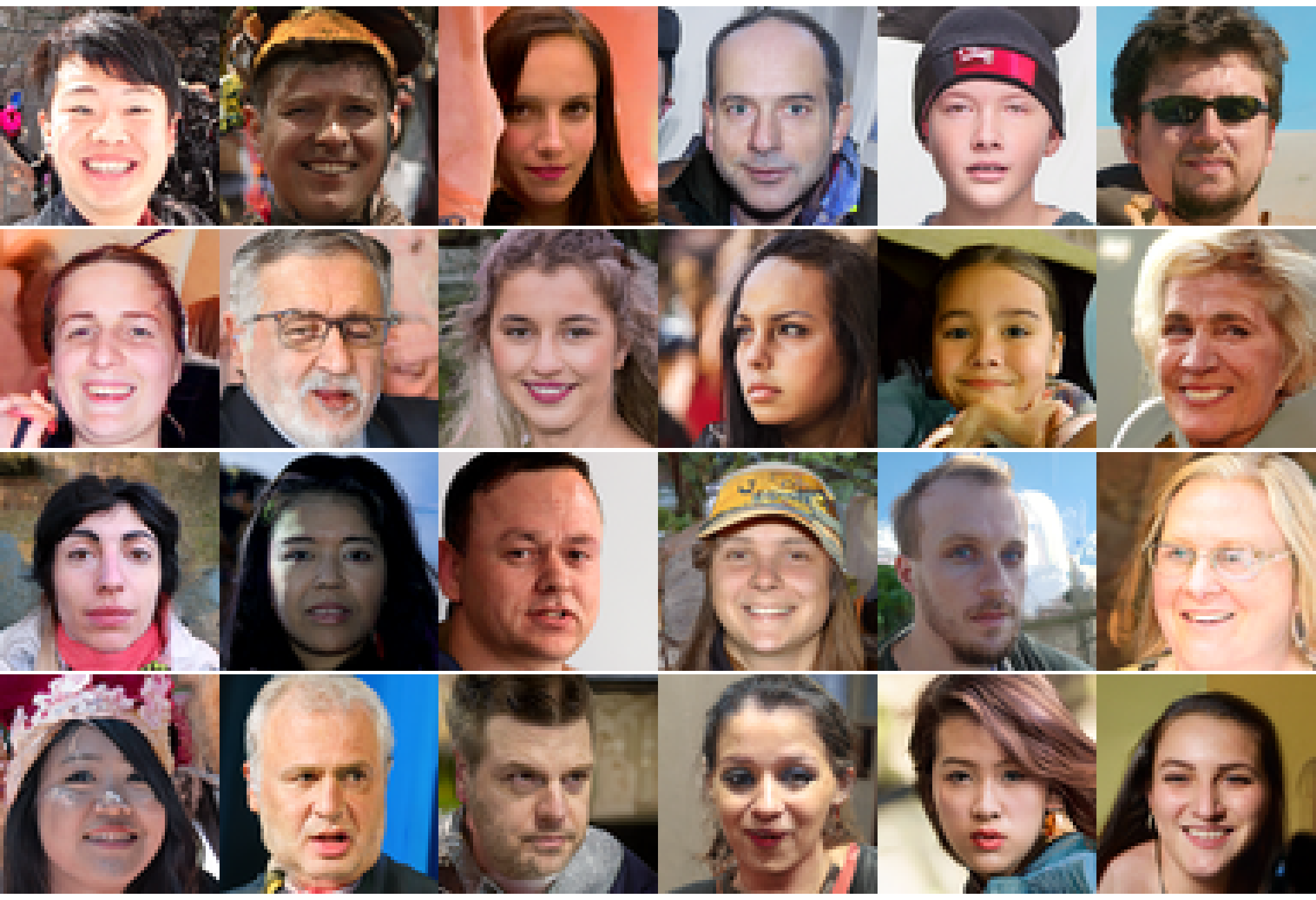}
        \caption{EDM-churn sampler with EDM schedule.}
    \end{subfigure}
    \hfill
    \begin{subfigure}[b]{0.46\textwidth}
        \centering
        \includegraphics[width=0.99\linewidth]{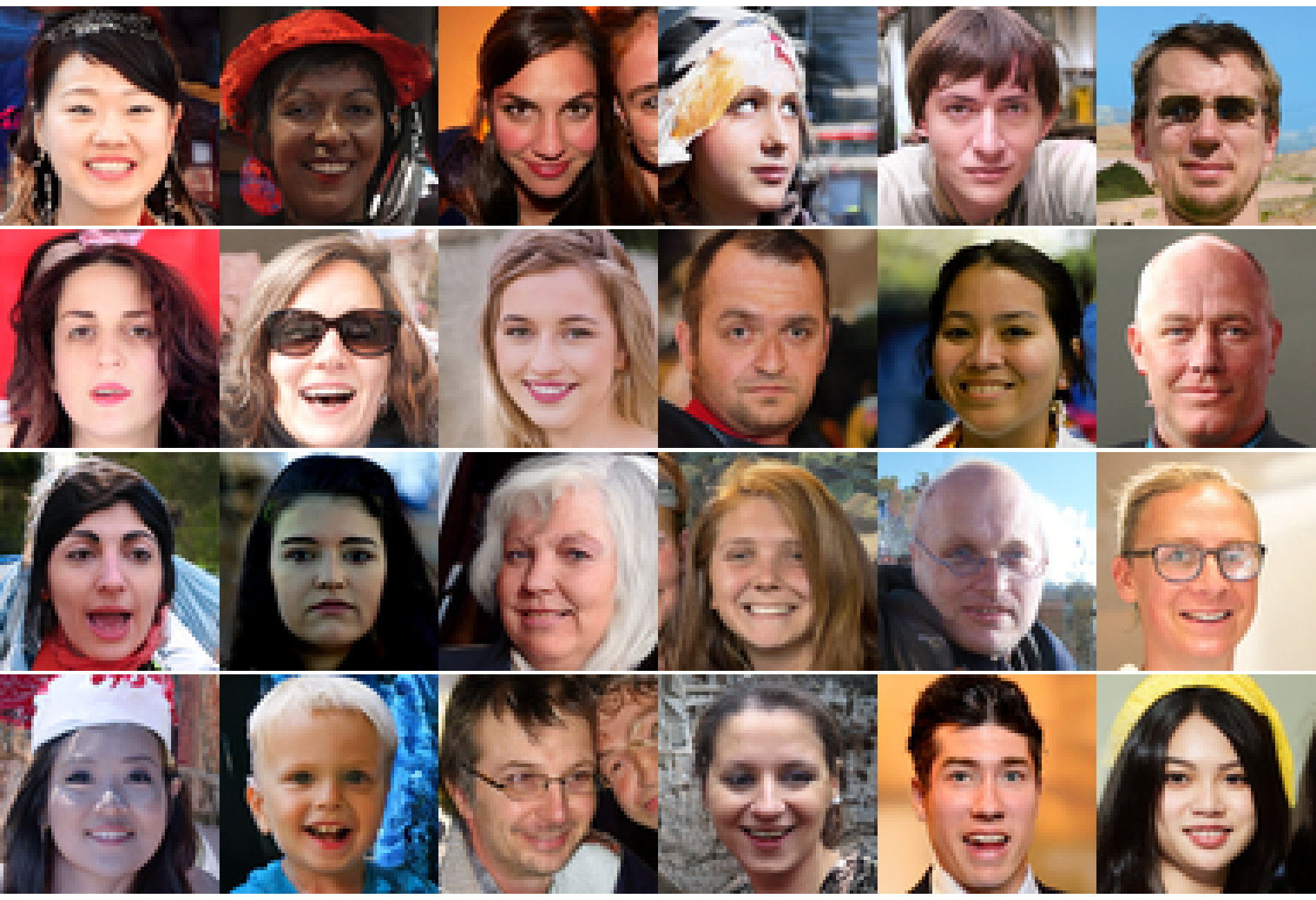}
        \caption{EDM-churn sampler with PI static schedule.}
    \end{subfigure}
    \begin{subfigure}[b]{0.46\textwidth}
        \centering
        \includegraphics[width=0.99\linewidth]{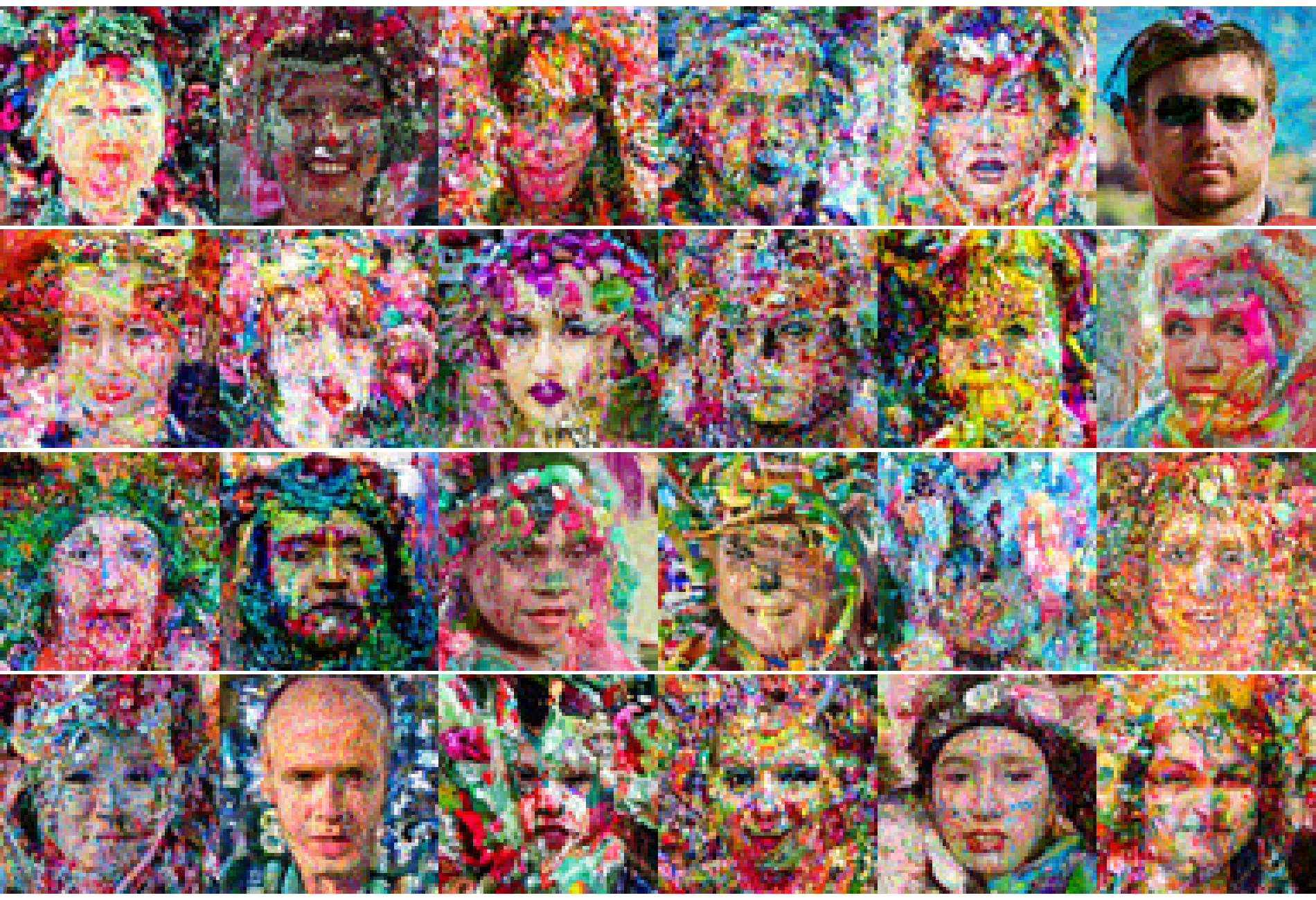}
        \caption{Gotta Go Fast adaptive sampler.}
    \end{subfigure}
    \hfill
    \begin{subfigure}[b]{0.46\textwidth}
        \centering
        \includegraphics[width=0.99\linewidth]{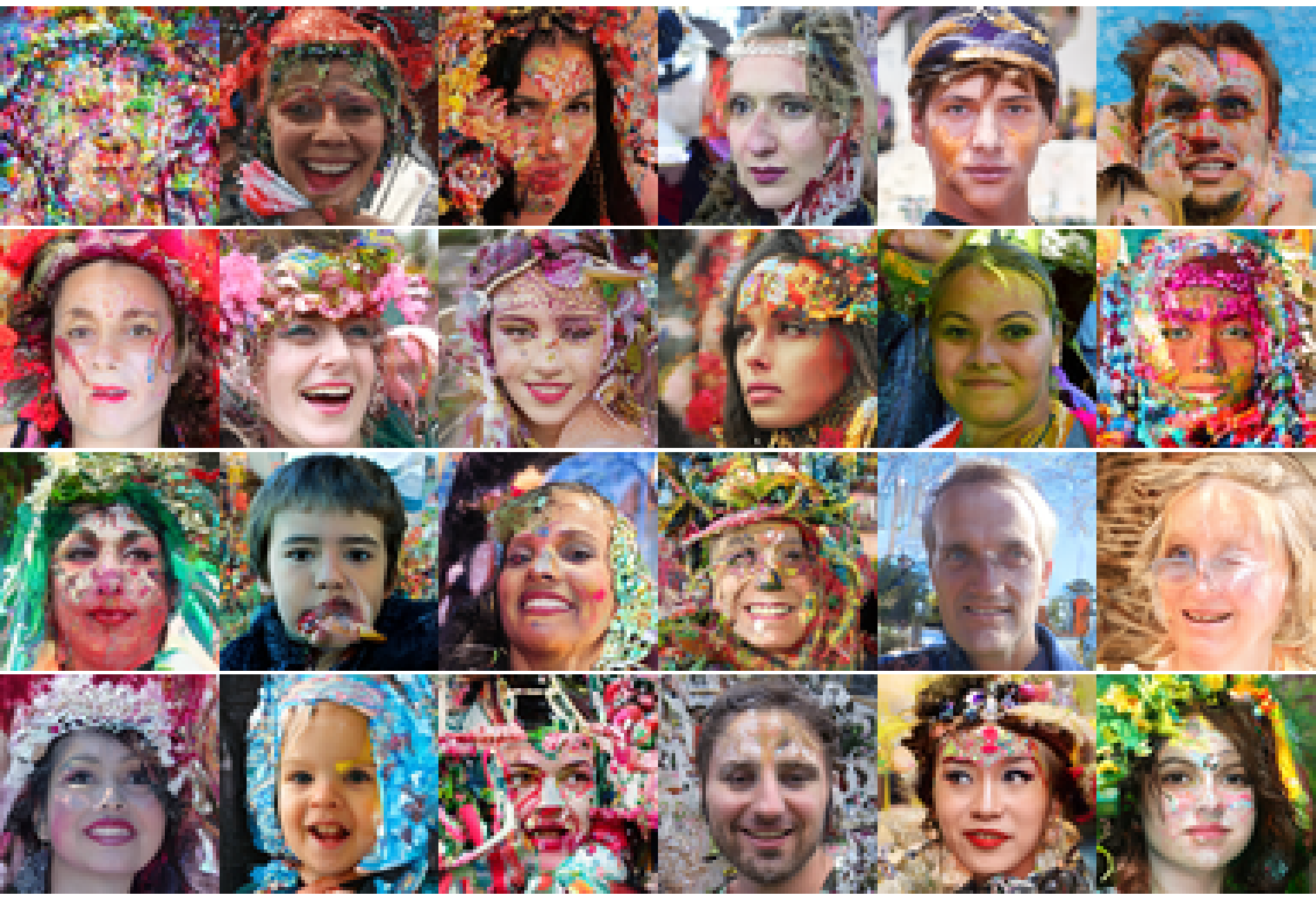}
        \caption{Proportional-Integral (ours) adaptive sampler.}
    \end{subfigure}
            \begin{subfigure}[b]{0.46\textwidth}
        \centering
        \includegraphics[width=0.99\linewidth]{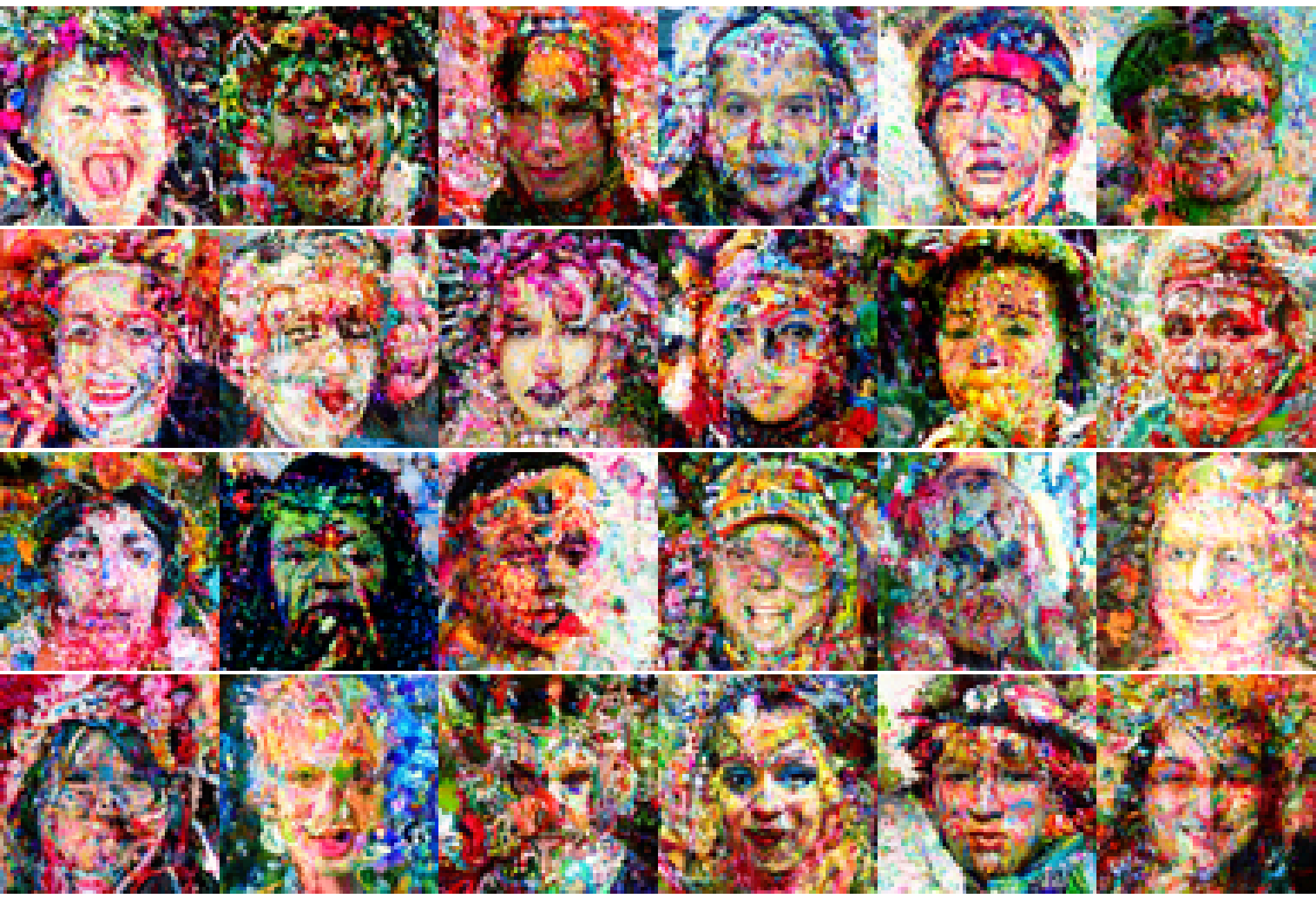}
        \caption{Stochastic Heun with EDM schedule.}
    \end{subfigure}
    \hfill
    \begin{subfigure}[b]{0.46\textwidth}
        \centering
        \includegraphics[width=0.99\linewidth]{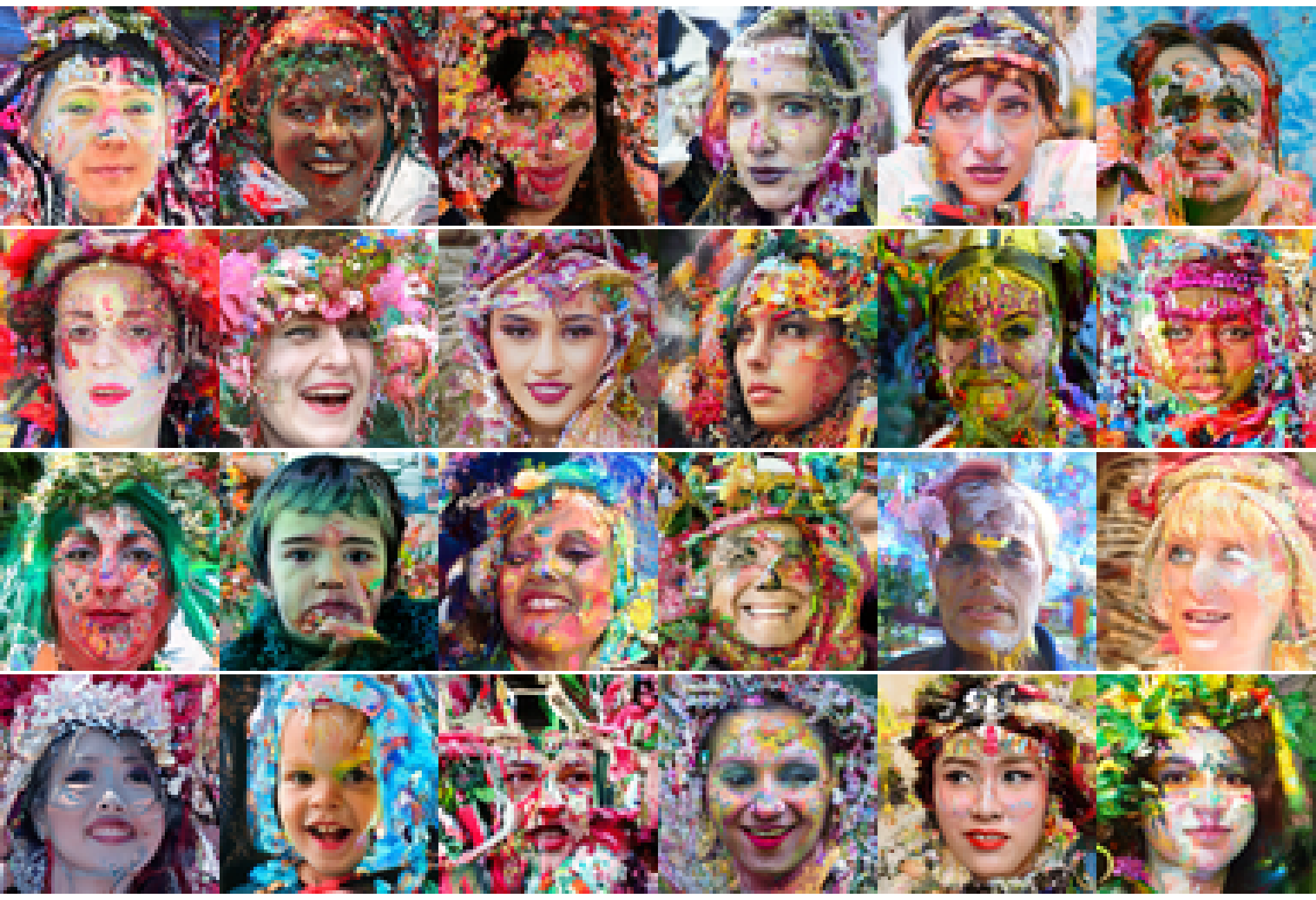}
        \caption{Stochastic Heun with PI static schedule.}
    \end{subfigure}
            \begin{subfigure}[b]{0.46\textwidth}
        \centering
        \includegraphics[width=0.99\linewidth]{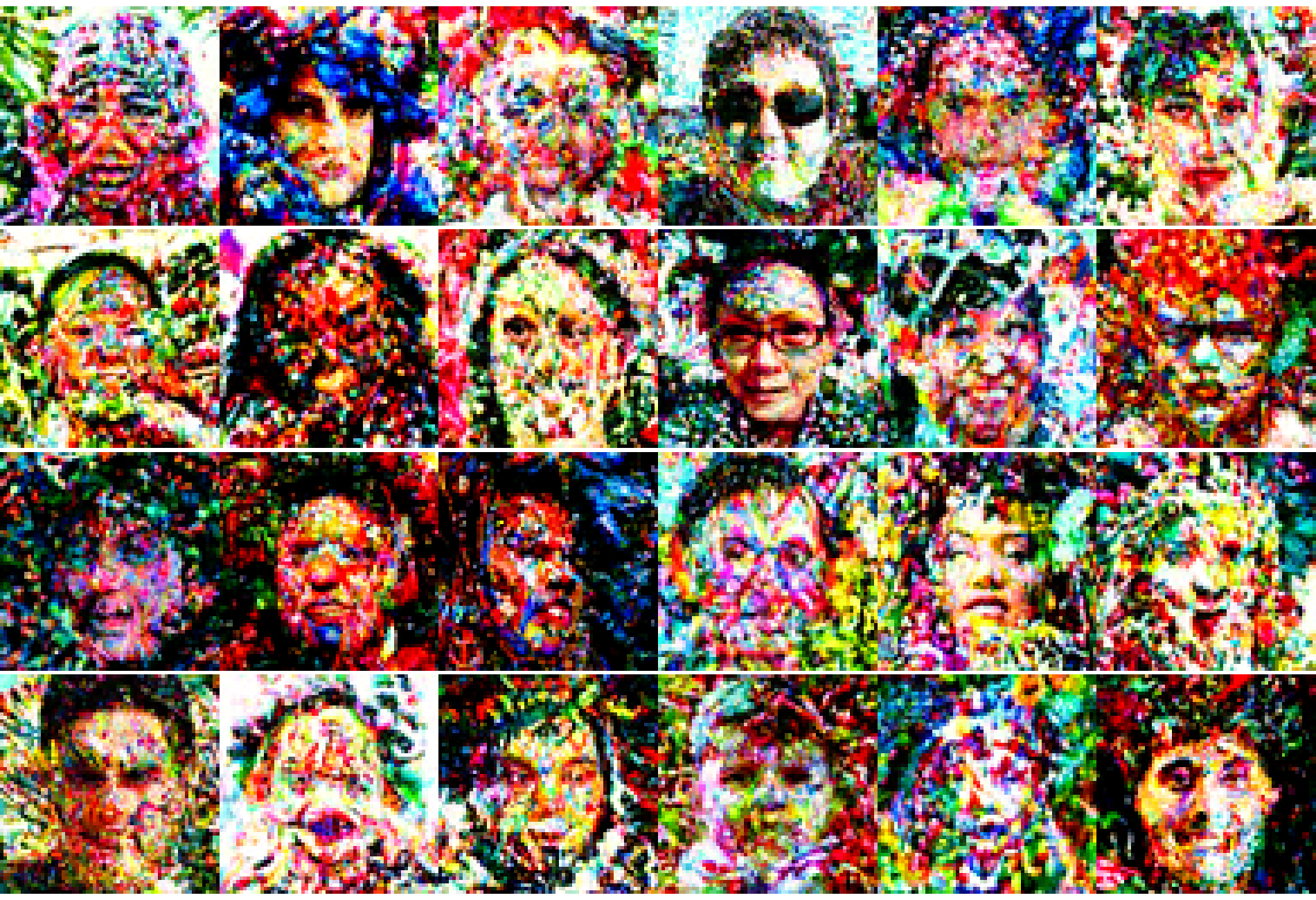}
        \caption{Euler-Maruyama with EDM schedule.}
    \end{subfigure}
    \hfill
    \begin{subfigure}[b]{0.46\textwidth}
        \centering
        \includegraphics[width=0.99\linewidth]{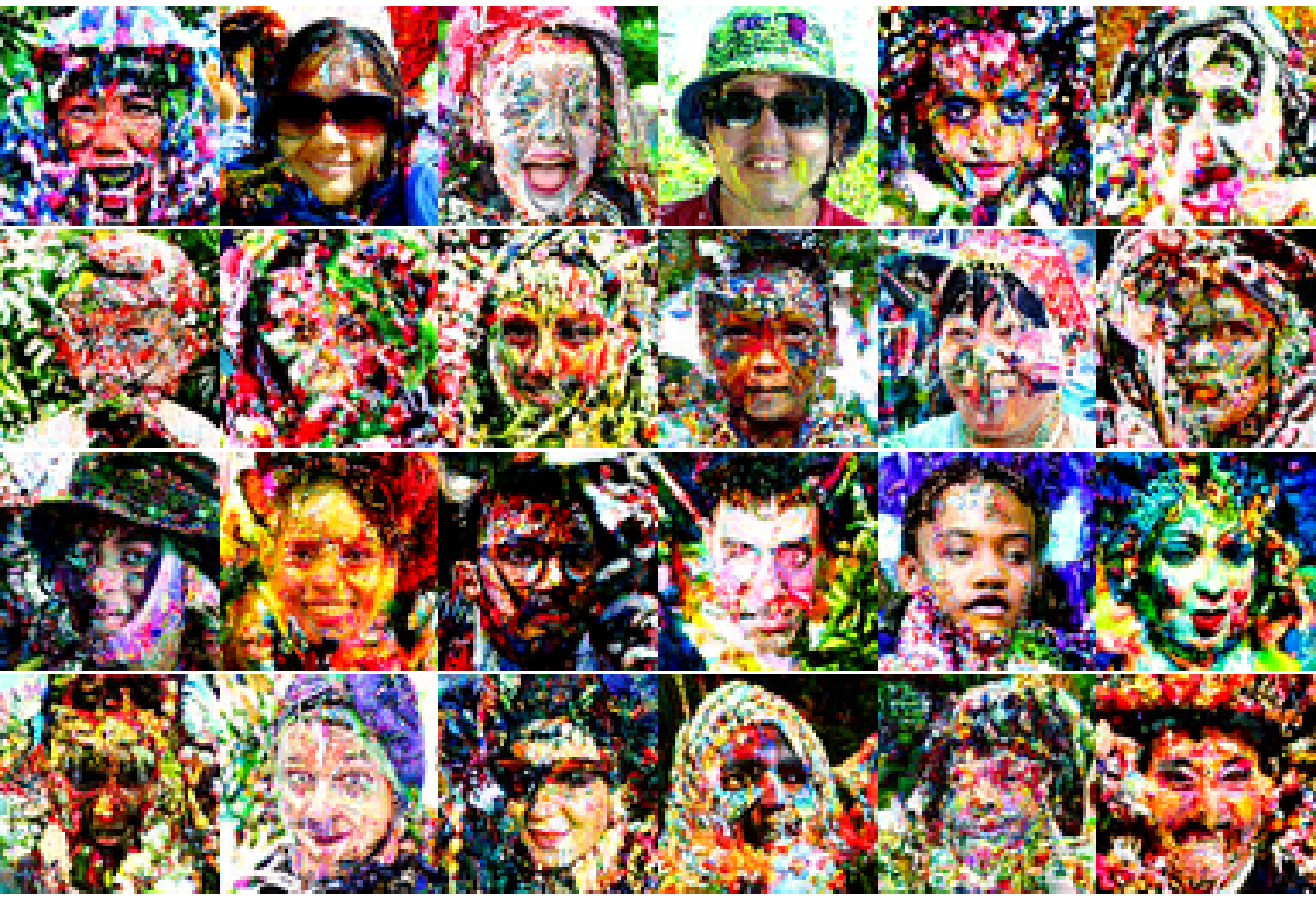}
        \caption{Euler-Maruyama with entropic schedule.}
    \end{subfigure}
    \caption{FFHQ Images generated using different SDE solvers using the EDM-FFHQ-64x64 unconditional VP checkpoint at 49 function evaluations}
    \label{fig:ffhq_49NFE_samples}
\end{figure}

\begin{figure}[H]
    \begin{subfigure}[b]{0.46\textwidth}
        \centering
        \includegraphics[width=0.99\linewidth]{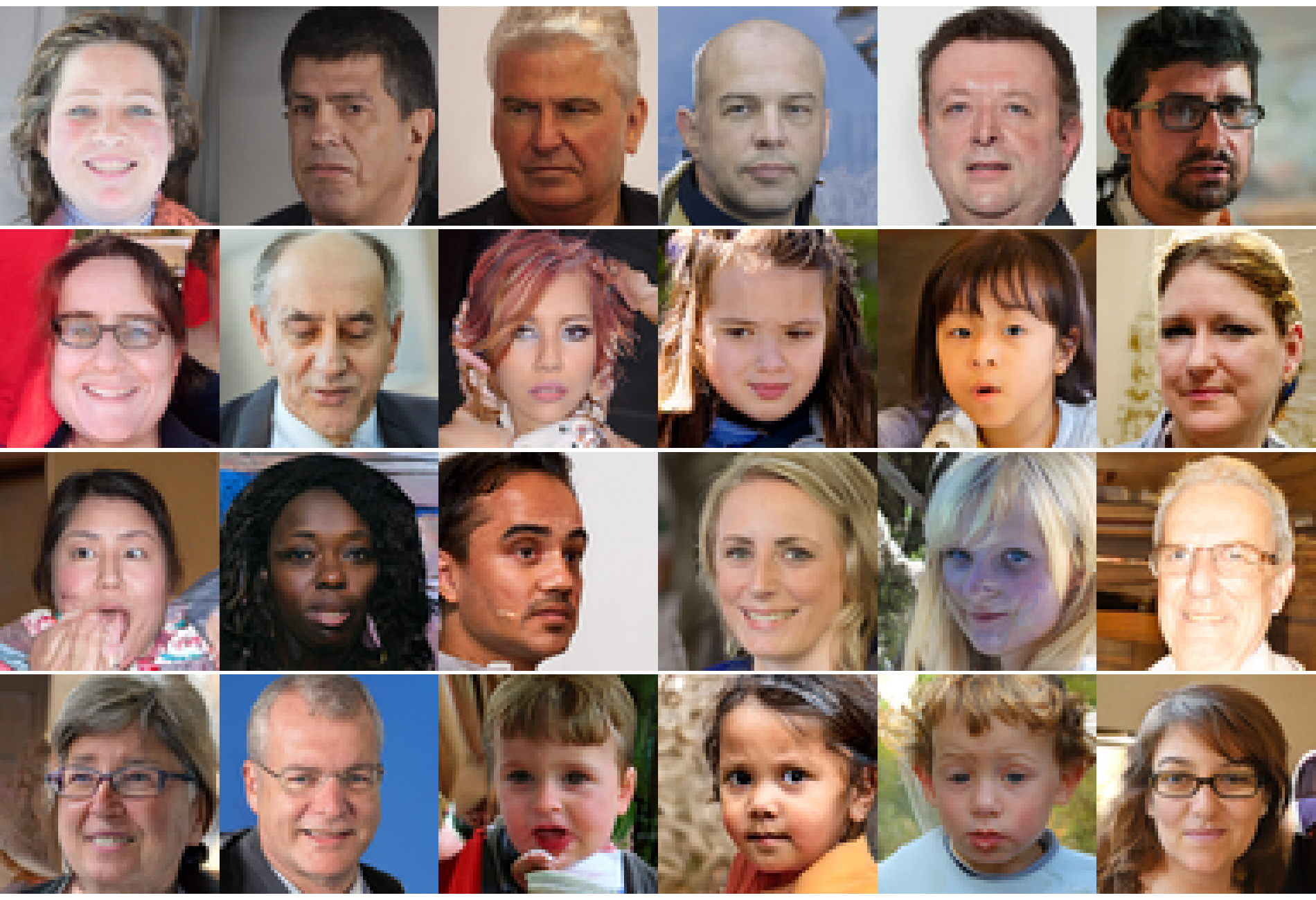}
        \caption{EDM-churn sampler with EDM schedule.}
    \end{subfigure}
    \hfill
    \begin{subfigure}[b]{0.46\textwidth}
        \centering
        \includegraphics[width=0.99\linewidth]{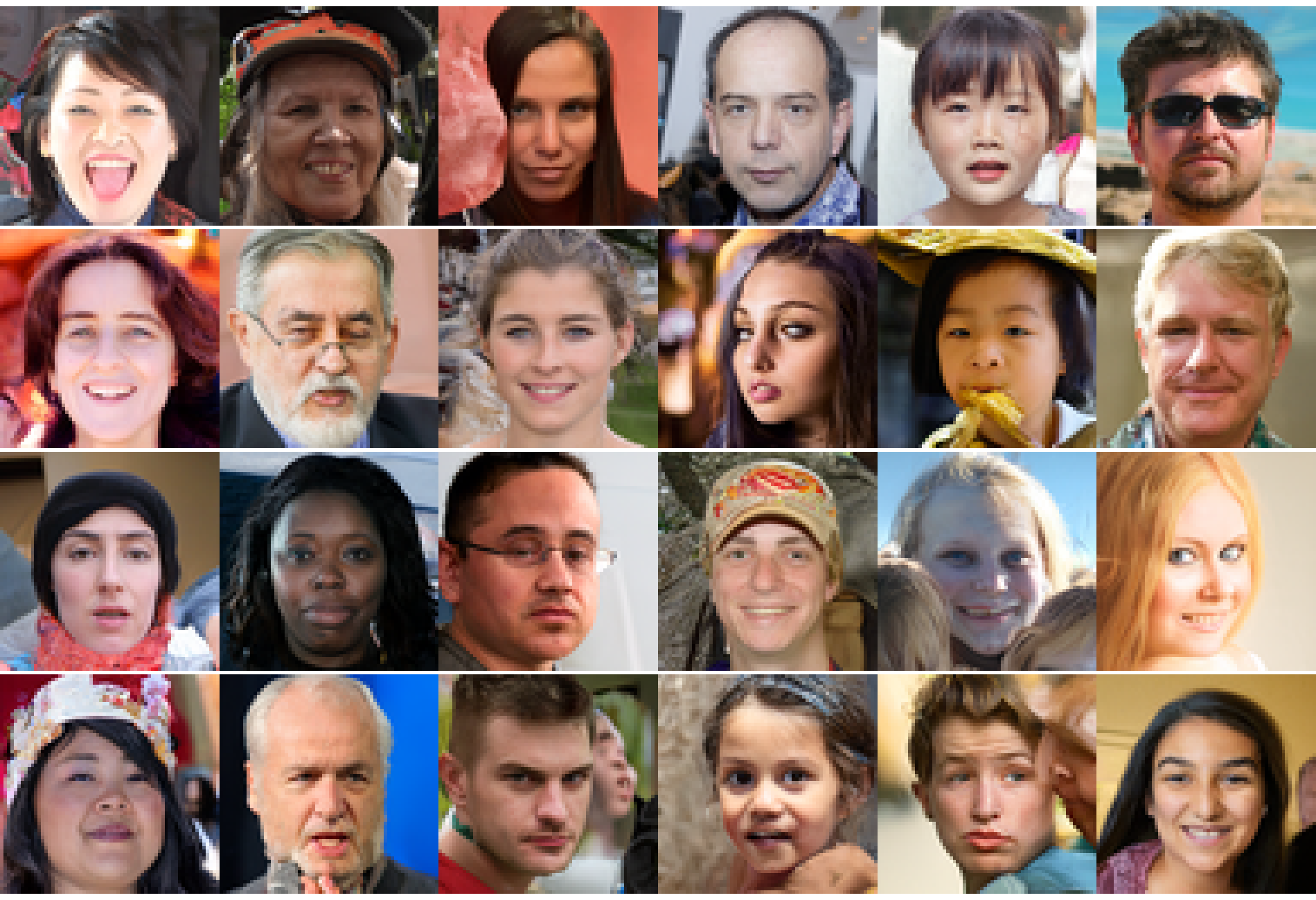}
        \caption{EDM-churn sampler with PI static schedule.}
    \end{subfigure}
    \begin{subfigure}[b]{0.46\textwidth}
        \centering
        \includegraphics[width=0.99\linewidth]{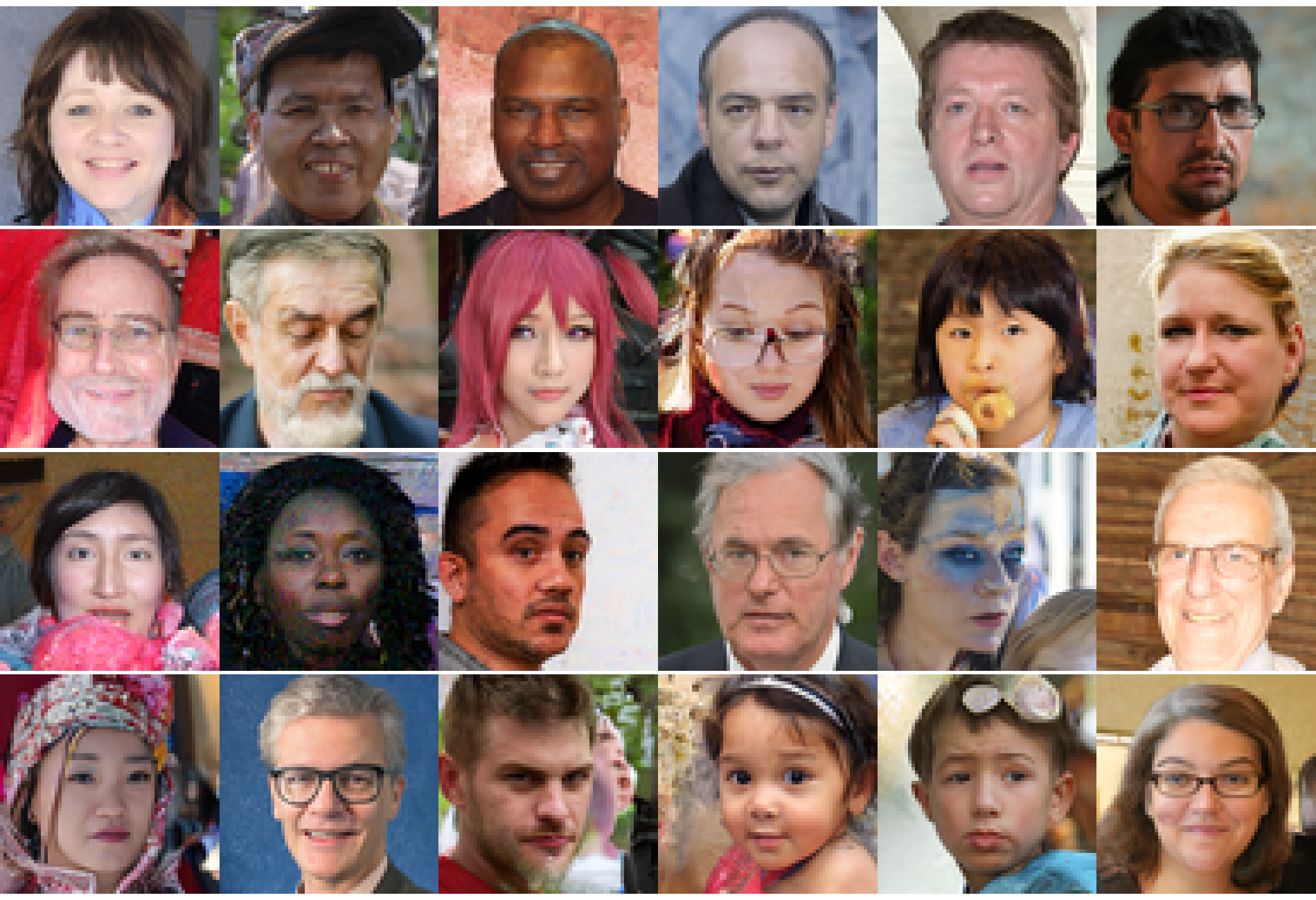}
        \caption{Gotta Go Fast adaptive sampler.}
    \end{subfigure}
    \hfill
    \begin{subfigure}[b]{0.46\textwidth}
        \centering
        \includegraphics[width=0.99\linewidth]{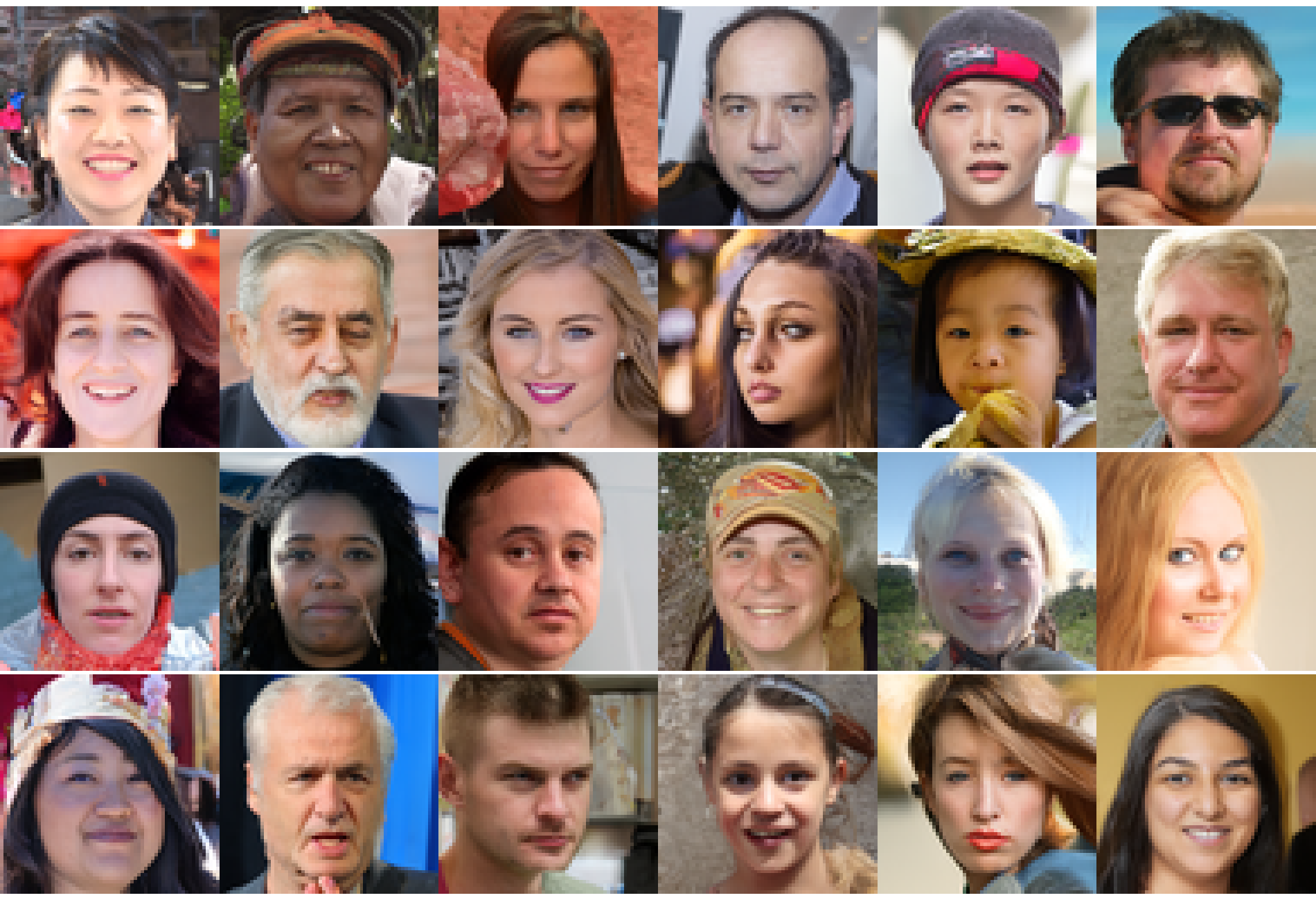}
        \caption{Proportional-Integral (ours) adaptive sampler.}
    \end{subfigure}
            \begin{subfigure}[b]{0.46\textwidth}
        \centering
        \includegraphics[width=0.99\linewidth]{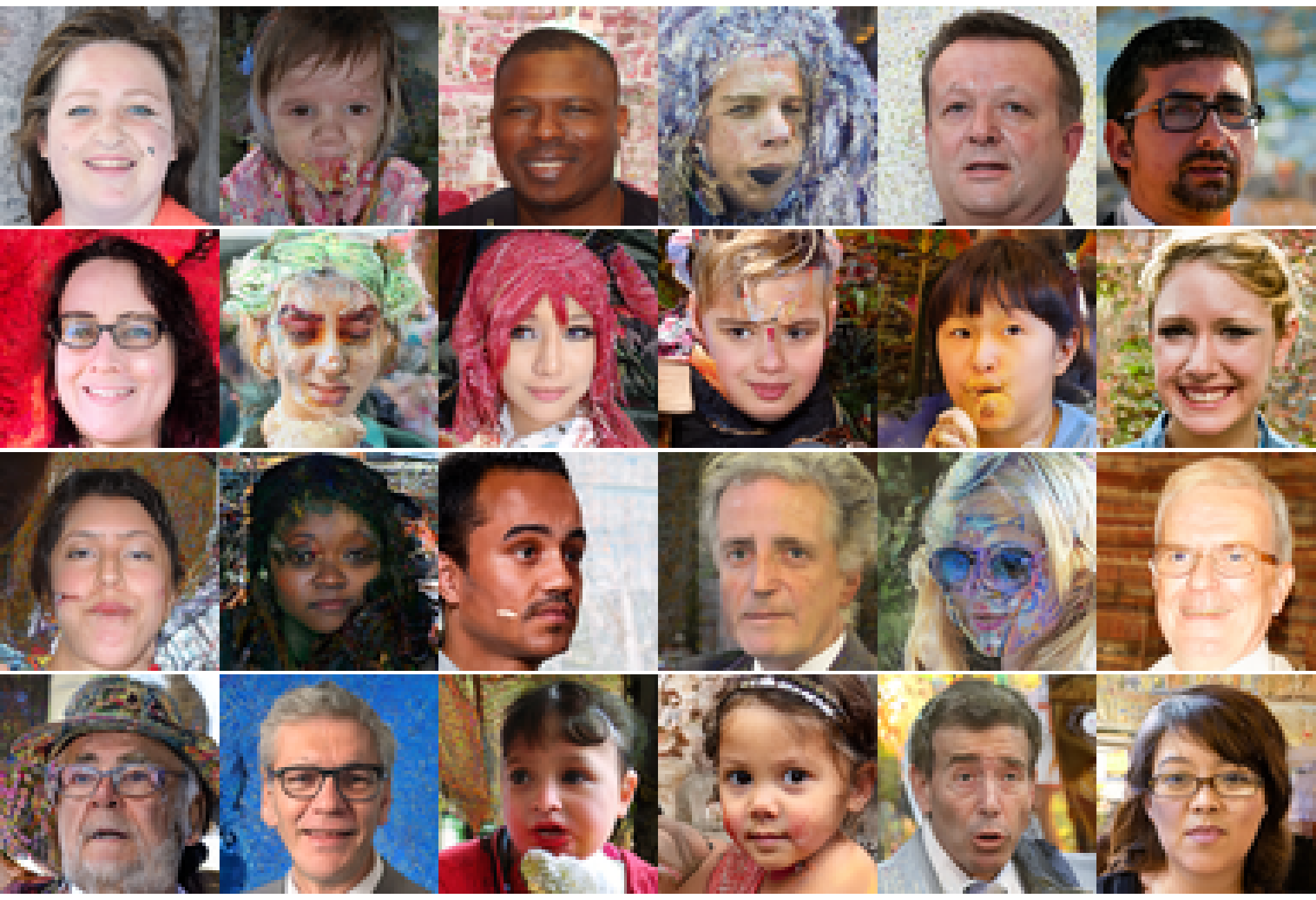}
        \caption{Stochastic Heun with EDM schedule.}
    \end{subfigure}
    \hfill
    \begin{subfigure}[b]{0.46\textwidth}
        \centering
        \includegraphics[width=0.99\linewidth]{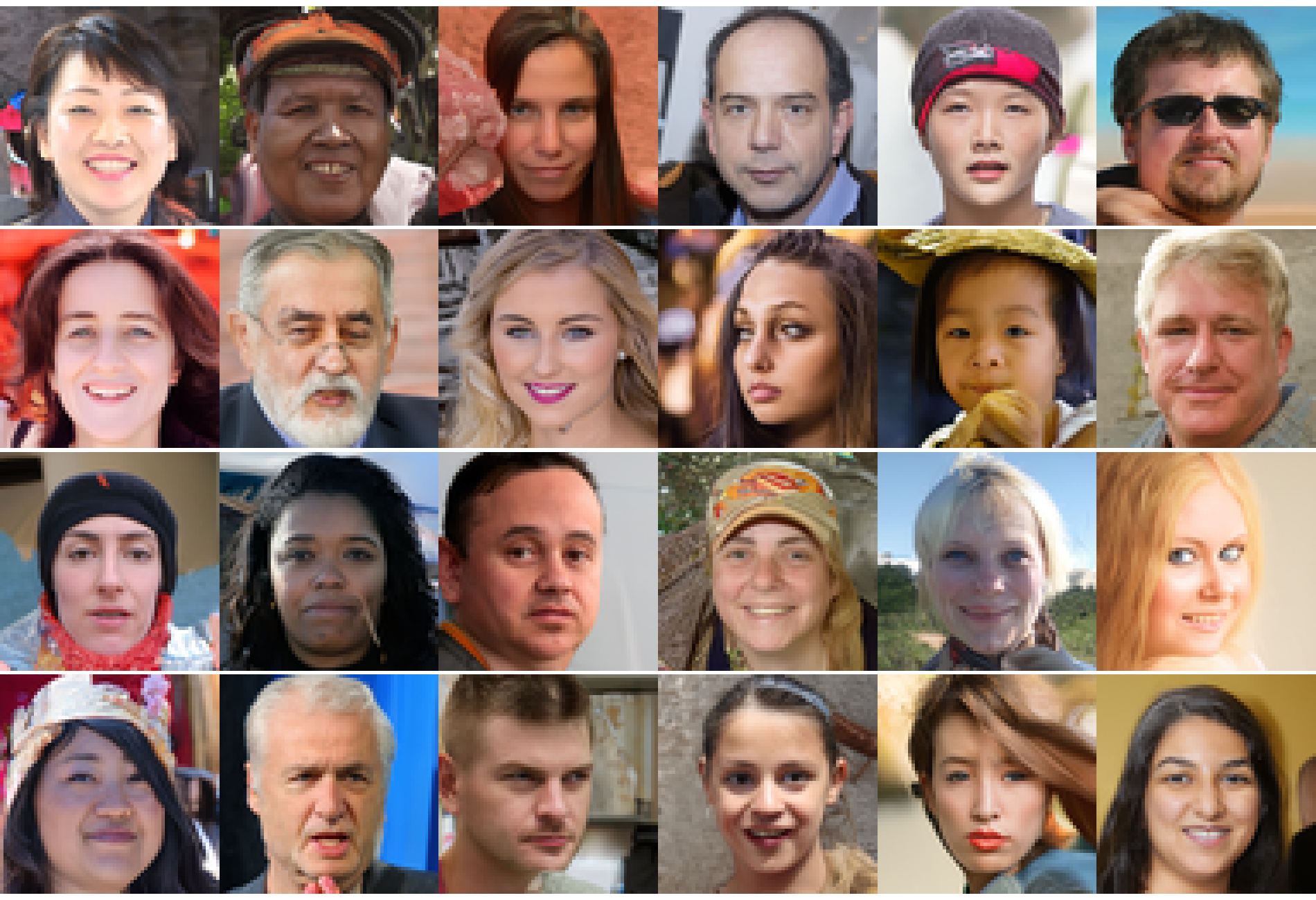}
        \caption{Stochastic Heun with PI static schedule.}
    \end{subfigure}
                \begin{subfigure}[b]{0.46\textwidth}
        \centering
        \includegraphics[width=0.99\linewidth]{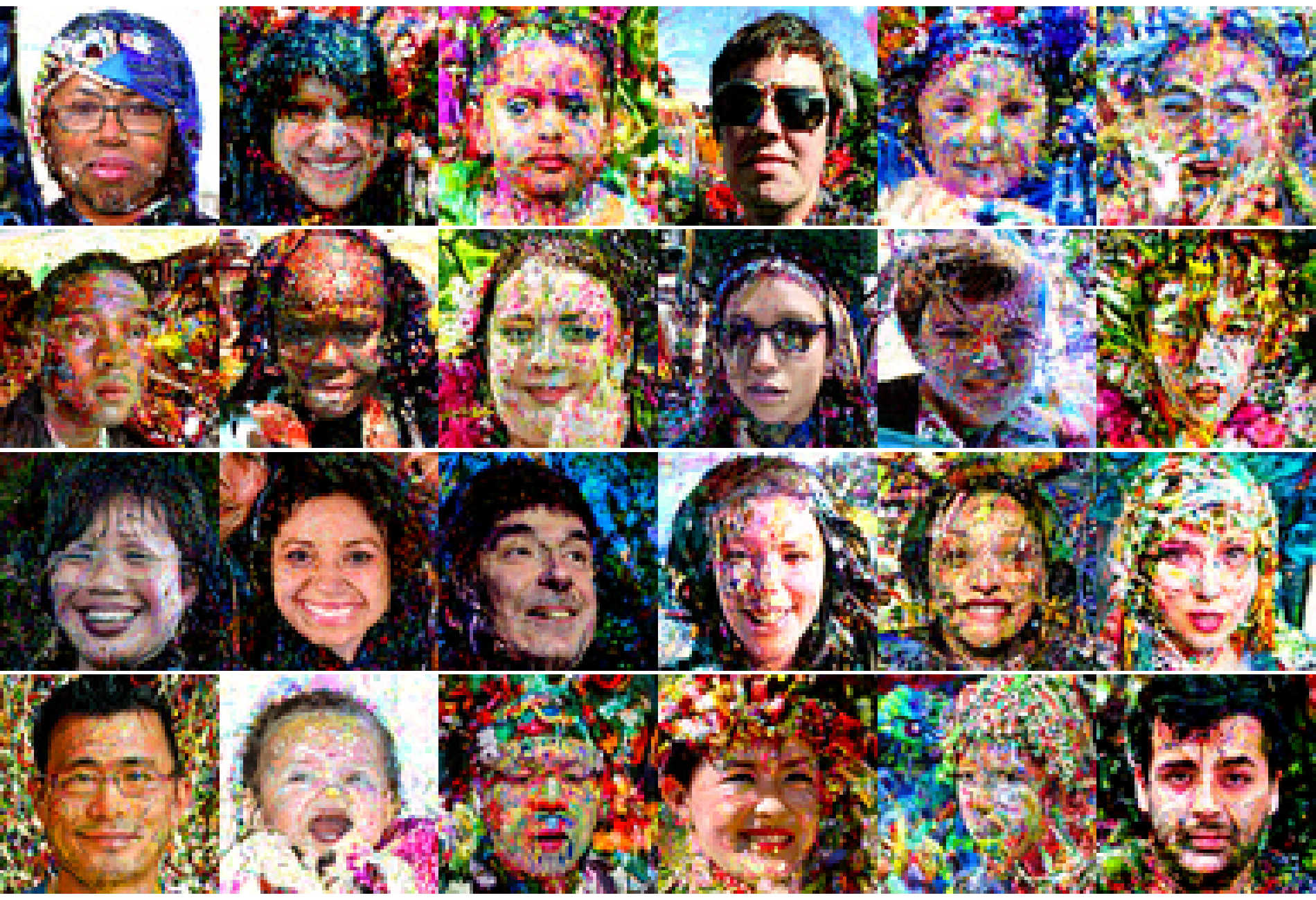}
        \caption{Euler-Maruyama with EDM schedule.}
    \end{subfigure}
    \hfill
    \begin{subfigure}[b]{0.46\textwidth}
        \centering
        \includegraphics[width=0.99\linewidth]{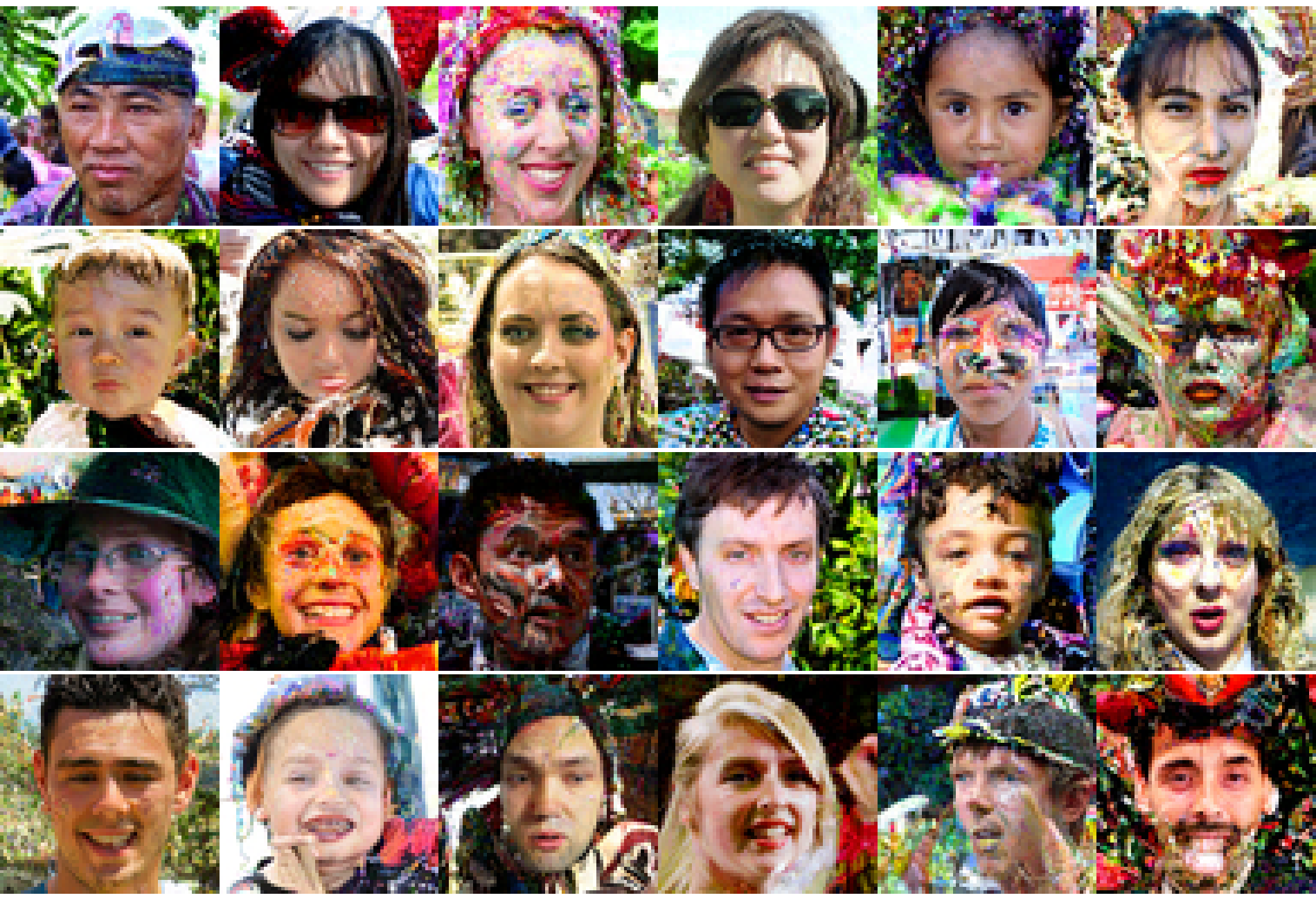}
        \caption{Euler-Maruyama with entropic schedule.}
    \end{subfigure}
    \caption{FFHQ Images generated using different SDE solvers using the EDM-FFHQ-64x64 unconditional VP checkpoint at 75 function evaluations}
    \label{fig:ffhq_75NFE_samples}
\end{figure}

\begin{figure}[H]
    \begin{subfigure}[b]{0.46\textwidth}
        \centering
        \includegraphics[width=0.99\linewidth]{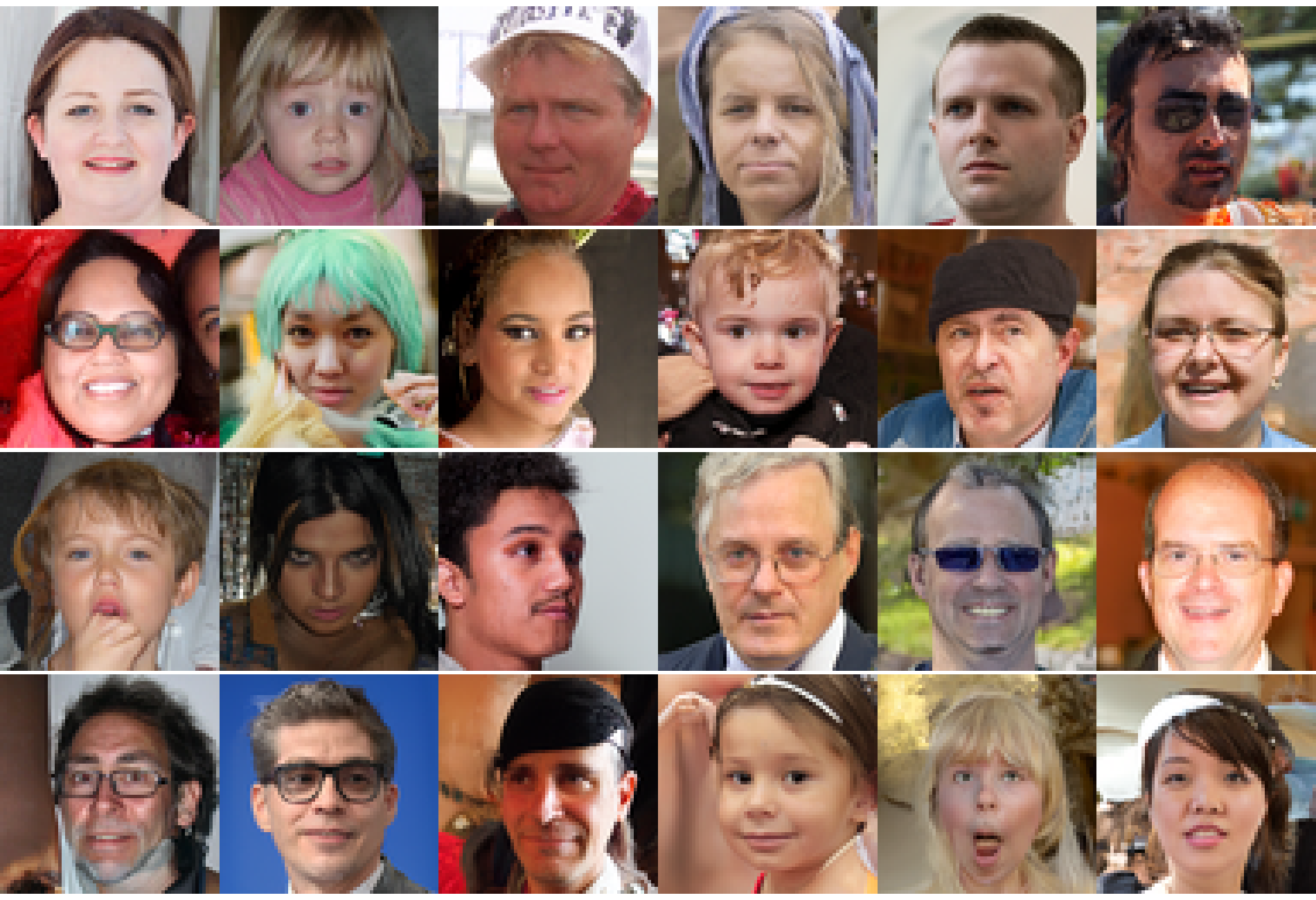}
        \caption{EDM-churn sampler with EDM schedule.}
    \end{subfigure}
    \hfill
    \begin{subfigure}[b]{0.46\textwidth}
        \centering
        \includegraphics[width=0.99\linewidth]{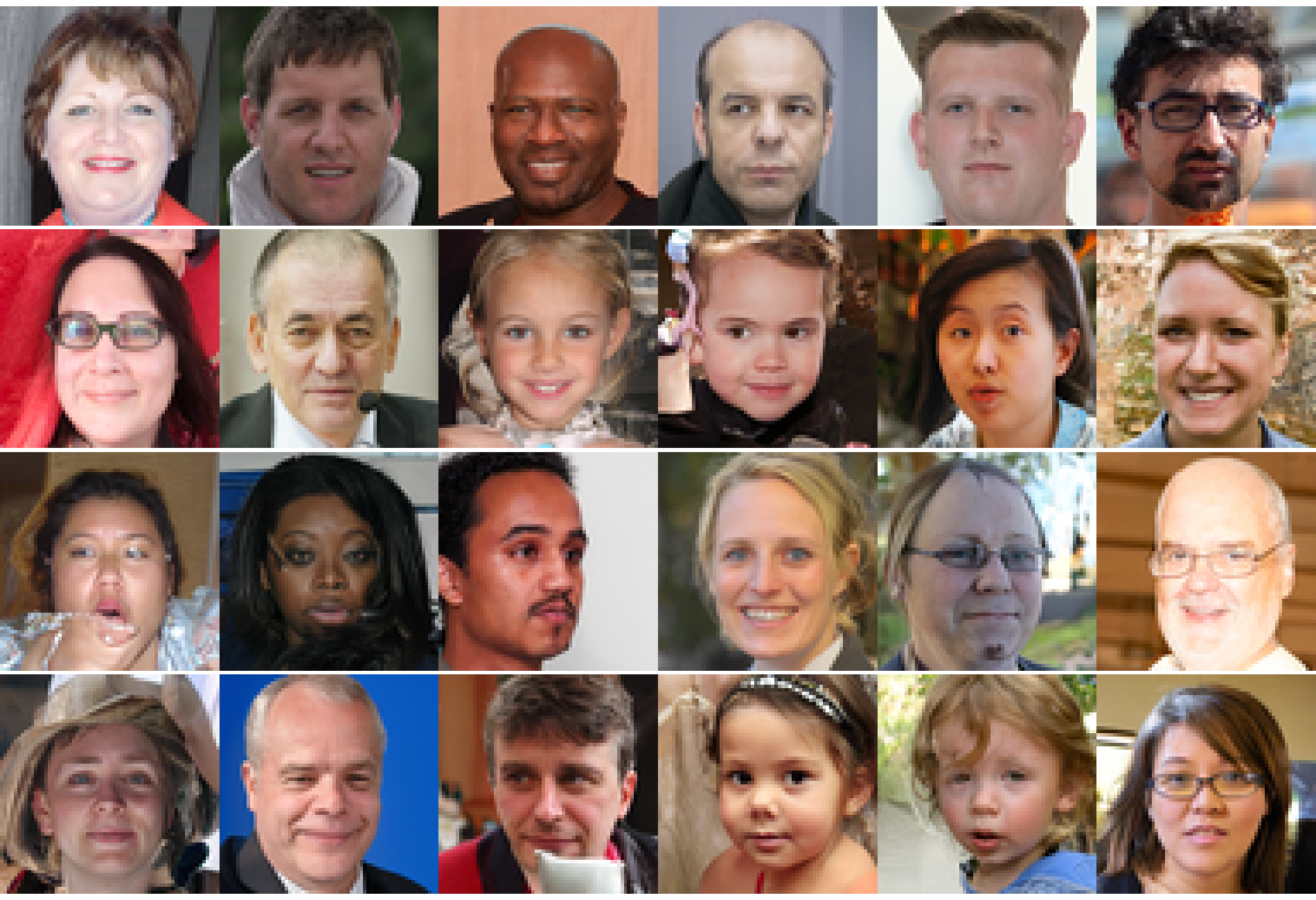}
        \caption{EDM-churn sampler with PI static schedule.}
    \end{subfigure}
    \begin{subfigure}[b]{0.46\textwidth}
        \centering
        \includegraphics[width=0.99\linewidth]{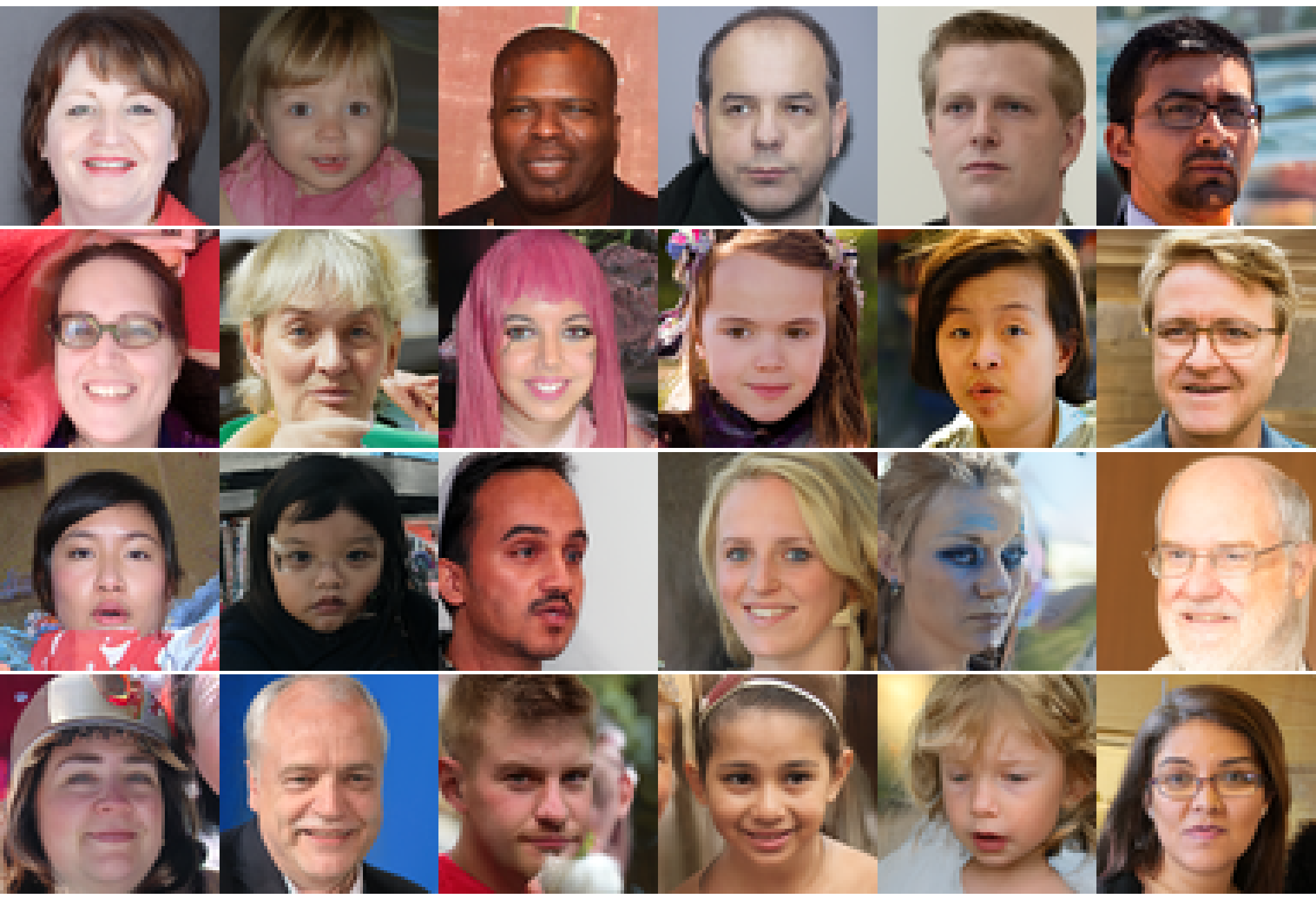}
        \caption{Gotta Go Fast adaptive sampler.}
    \end{subfigure}
    \hfill
    \begin{subfigure}[b]{0.46\textwidth}
        \centering
        \includegraphics[width=0.99\linewidth]{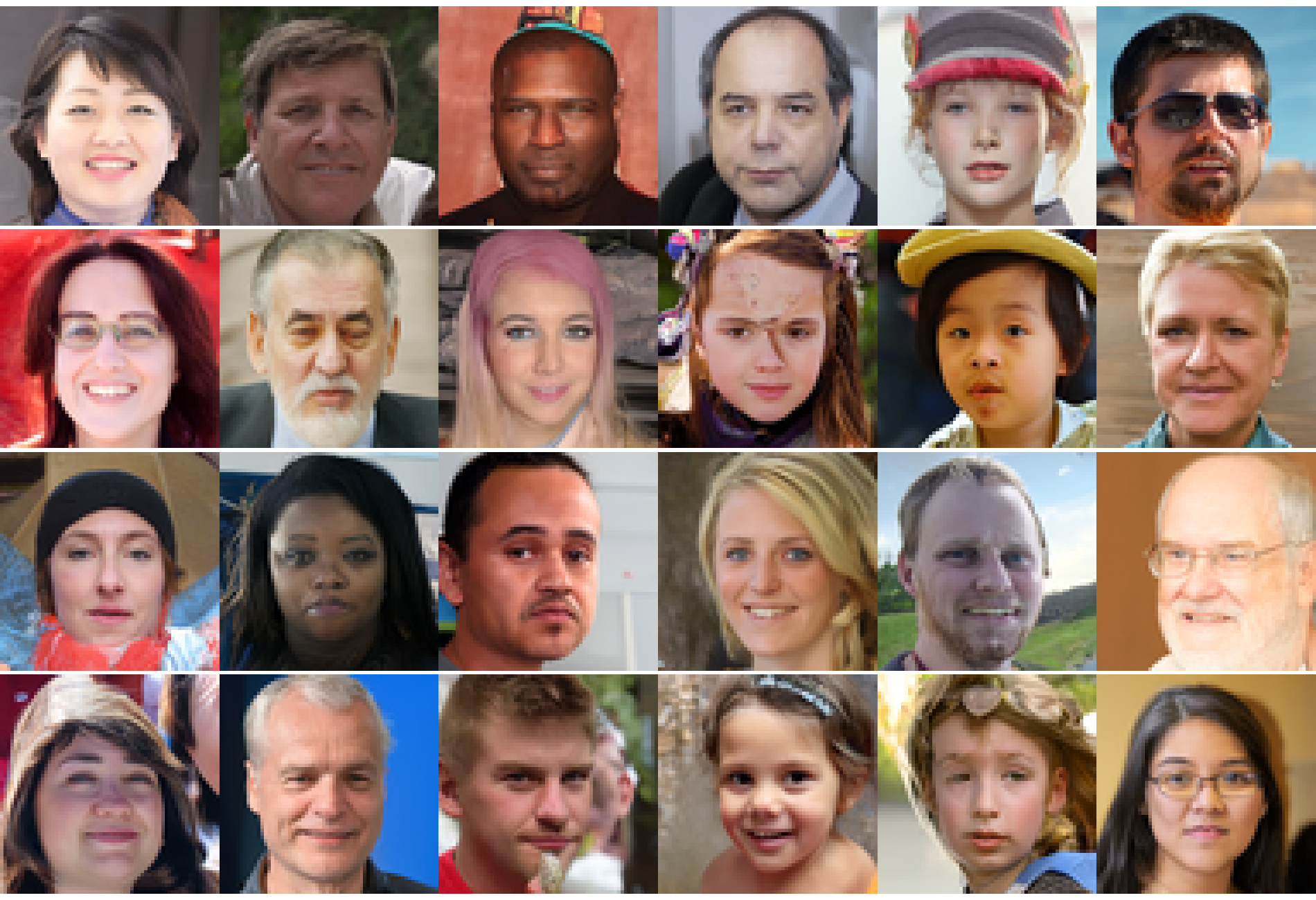}
        \caption{Proportional-Integral (ours) adaptive sampler.}
    \end{subfigure}
            \begin{subfigure}[b]{0.46\textwidth}
        \centering
        \includegraphics[width=0.99\linewidth]{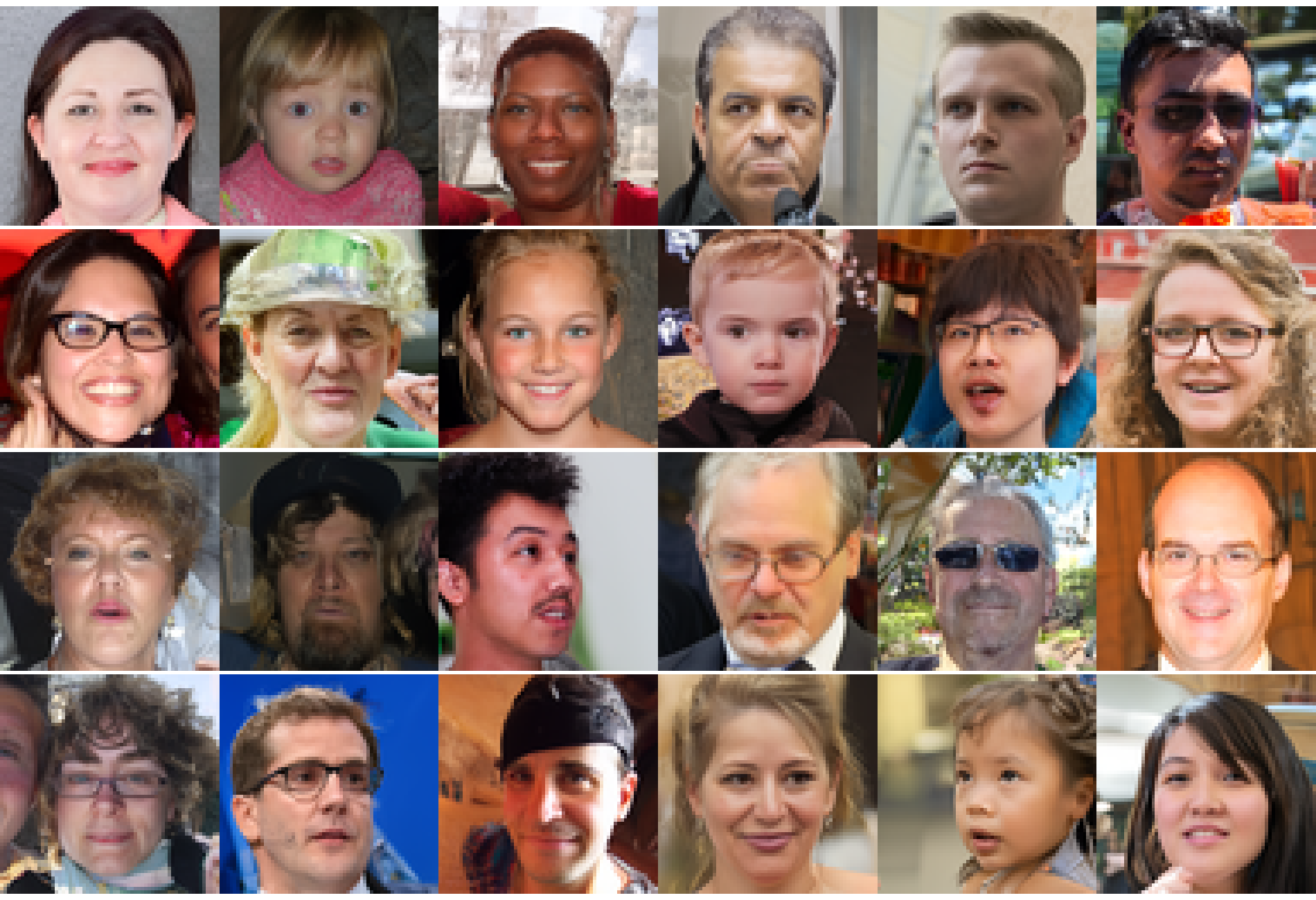}
        \caption{Stochastic Heun with EDM schedule.}
    \end{subfigure}
    \hfill
    \begin{subfigure}[b]{0.46\textwidth}
        \centering
        \includegraphics[width=0.99\linewidth]{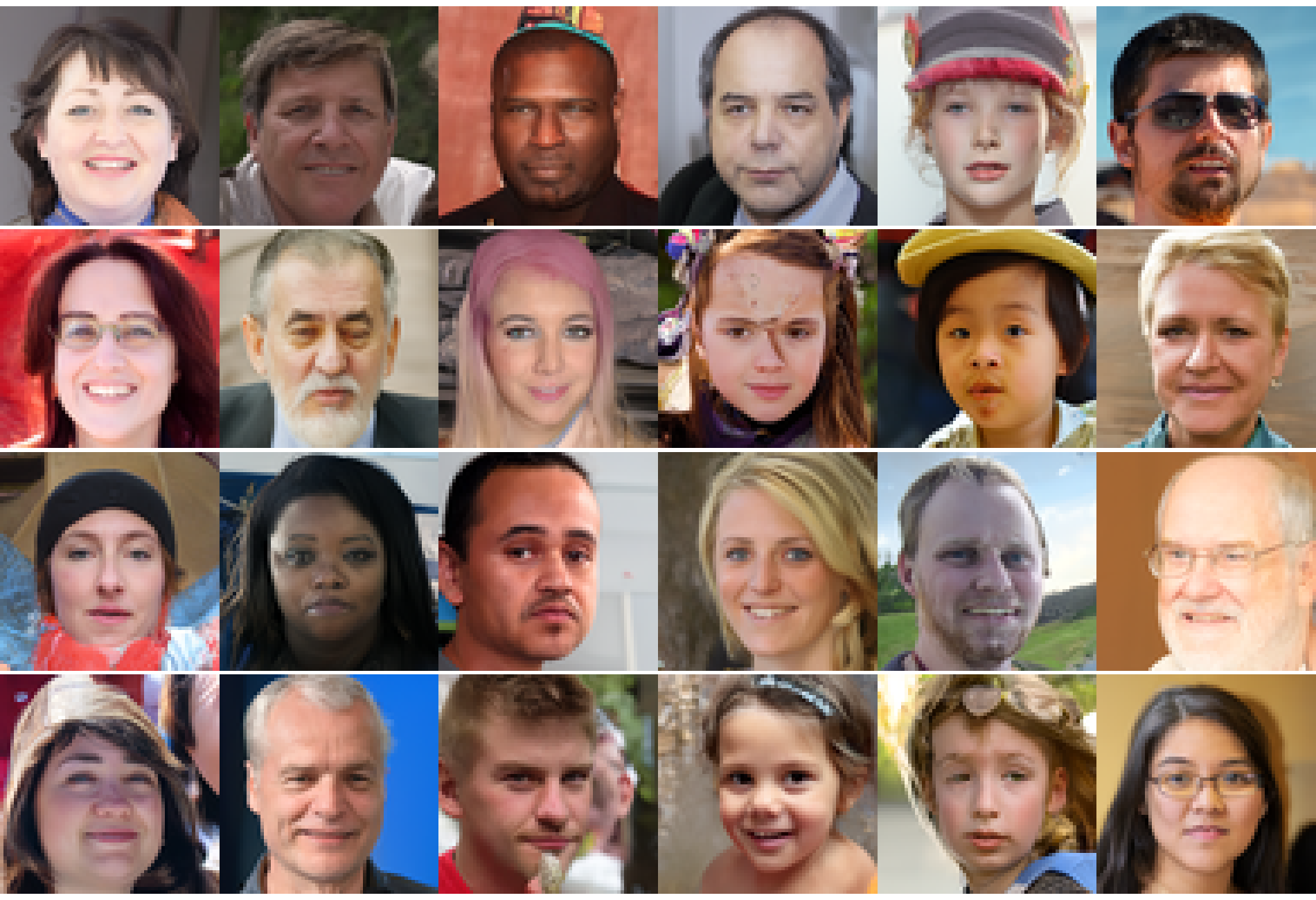}
        \caption{Stochastic Heun with PI static schedule.}
    \end{subfigure}
                \begin{subfigure}[b]{0.46\textwidth}
        \centering
        \includegraphics[width=0.99\linewidth]{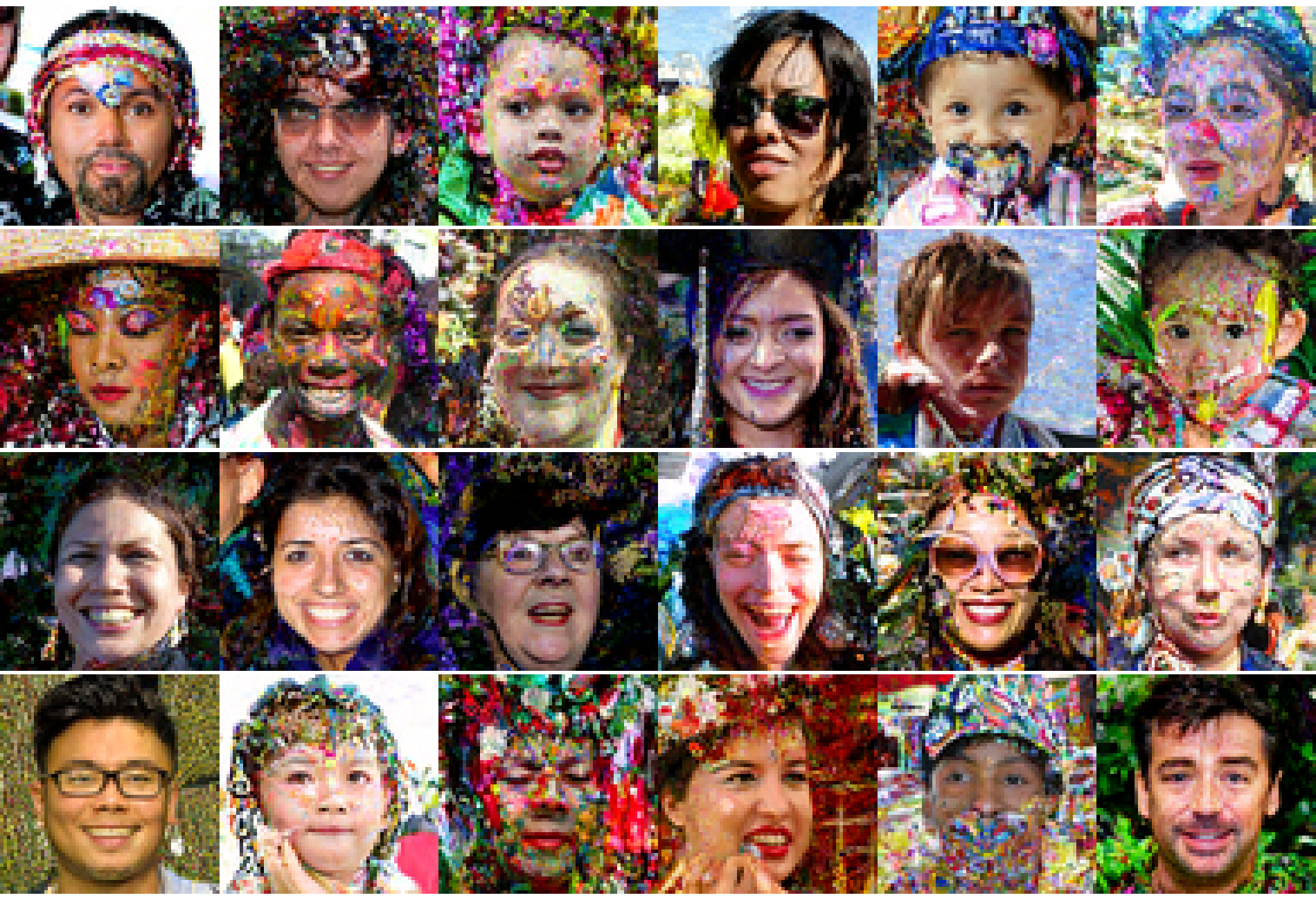}
        \caption{Euler-Maruyama with EDM schedule.}
    \end{subfigure}
    \hfill
    \begin{subfigure}[b]{0.46\textwidth}
        \centering
        \includegraphics[width=0.99\linewidth]{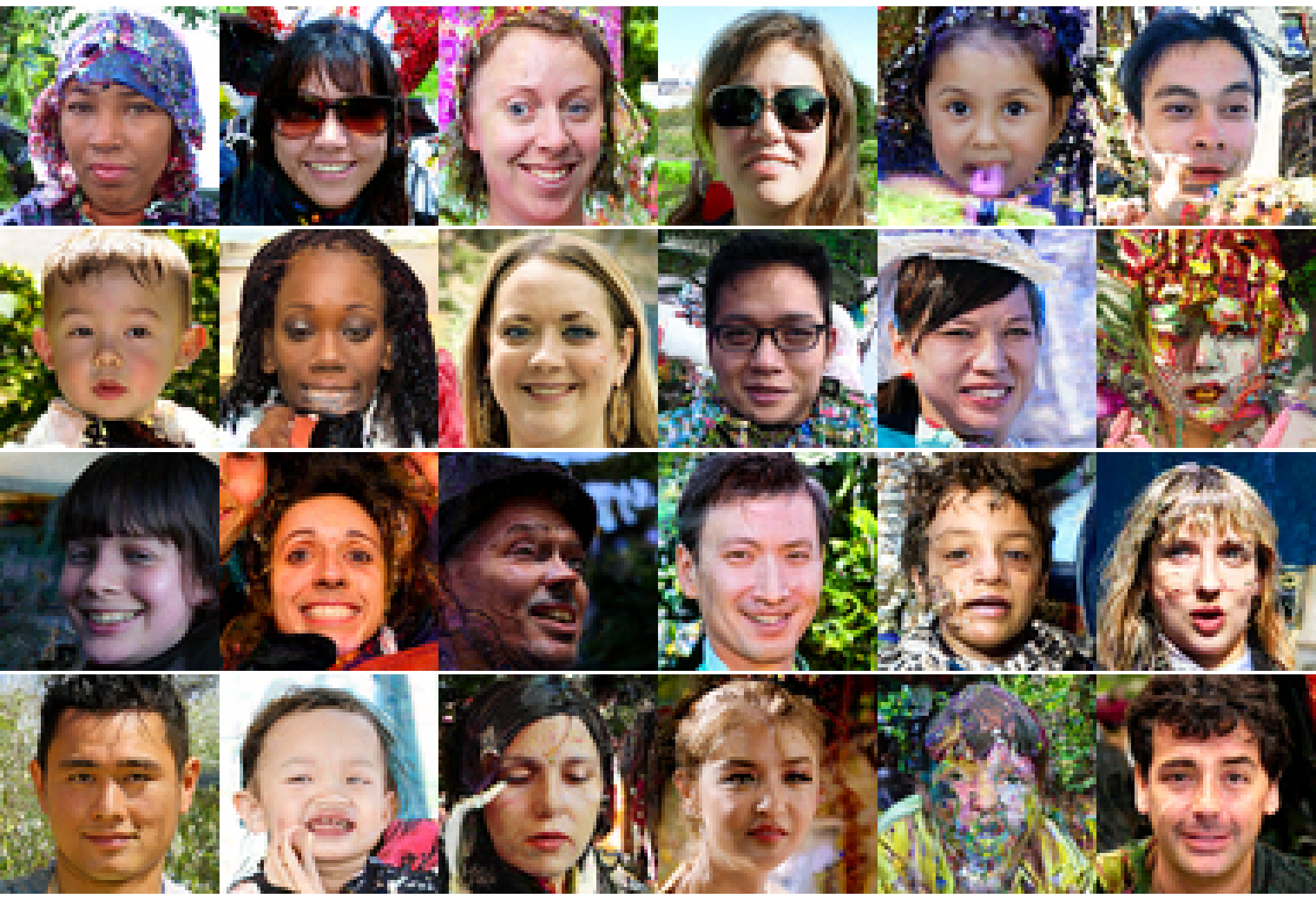}
        \caption{Euler-Maruyama with entropic schedule.}
    \end{subfigure}
    \caption{FFHQ Images generated using different SDE solvers using the EDM-FFHQ-64x64 unconditional VP checkpoint at 99 function evaluations}
    \label{fig:ffhq_99_NFE_samples}
\end{figure}